\pdfoutput=1

\documentclass[11pt]{article}

\usepackage[preprint]{acl}

\usepackage{times}
\usepackage{latexsym}

\usepackage[T1]{fontenc}

\usepackage[utf8]{inputenc}

\usepackage{microtype}

\usepackage{inconsolata}

\usepackage{graphicx}
\usepackage{subcaption} 
\usepackage{booktabs}
\usepackage{multirow}
\usepackage{amsmath}
\usepackage{amsfonts}

\usepackage[capitalise]{cleveref}

\usepackage{todonotes}
\usepackage{soul}
\usepackage[normalem]{ulem}
\usepackage{bm}
\usepackage{dsfont}
\usepackage{mathtools}
\usepackage{pifont}
\newcommand{\ours}{CoPL}

\newcommand{\textBF}[1]{%
    \pdfliteral direct {2 Tr 0.3 w} 
     #1%
    \pdfliteral direct {0 Tr 0 w}%
} 

\newcommand{\set}[1]{\mathcal{#1}}

\newcommand{\embedding}[1]{\textBF{#1}}

\title{CoPL: Collaborative Preference Learning for Personalizing LLMs}

\author{
 \textbf{Youngbin Choi\textsuperscript{1}},
 \textbf{Seunghyuk Cho\textsuperscript{1}},
 \textbf{Minjong Lee \textsuperscript{2}},
 \textbf{MoonJeong Park\textsuperscript{1}}, \\
 \textbf{Yesong Ko\textsuperscript{2}},
 \textbf{Jungseul Ok\textsuperscript{1,2}},
 \textbf{Dongwoo Kim\textsuperscript{1,2,*}}
\\
 \textsuperscript{1}Graduate School of Artificial Intelligence, POSTECH,
\\
 \textsuperscript{2}Department of Computer Science and Engineering, POSTECH, 
\\
 \small{
    \{choi.youngbin, shhj1998, minjong.lee, mjeongp, yesong.ko, jungseul, dongwoo.kim\}@postech.ac.kr
 }
}

\begin{document}
\maketitle
\begingroup
\renewcommand\thefootnote{}\footnotetext{$^{*}$\textbf{Correspondence to:} Dongwoo Kim  \href{mailto:dongwoo.kim@postech.ac.kr}{<dongwoo.kim@postech.ac.kr>}}
\endgroup


\begin{abstract}
    Personalizing large language models (LLMs) is important for aligning outputs with diverse user preferences, yet existing methods struggle with flexibility and generalization. We propose CoPL (Collaborative Preference Learning), a graph-based collaborative filtering framework that models user-response relationships to enhance preference estimation, particularly in sparse annotation settings. By integrating a mixture of LoRA experts, CoPL efficiently fine-tunes LLMs while dynamically balancing shared and user-specific preferences. Additionally, an optimization-free adaptation strategy enables generalization to unseen users without fine-tuning. Experiments on TL;DR, UltraFeedback-P, and PersonalLLM datasets demonstrate that CoPL outperforms existing personalized reward models, effectively capturing both common and controversial preferences, making it a scalable solution for personalized LLM alignment. The code is available at \href{https://github.com/ml-postech/CoPL}{\texttt{https://github.com/ml-postech/CoPL}}.

\end{abstract}

\section{Introduction}

Large language models (LLMs) have rapidly expanded across diverse applications, from customer service and tutoring to creative content generation~\cite{shi2024chops, molina2024leveraging, venkatraman2024collabstory}. As increasing numbers of users with varied backgrounds interact with LLMs, accounting for diverse preferences has become essential. Most reward models rely on the Bradley-Terry-Luce (BTL) framework~\citep{bradley1952rank}, which learns preferences from pairwise comparisons provided by human annotators. However, earlier studies largely assumed a single, uniform preference and neglected the diversity of user preferences~\citep{siththaranjan2024DPL, li2024personalized, li20251}. This limitation has led to growing interest in personalized reward models~\citep{sorensen2024roadmap, liu2025survey, guan2025survey}. 

\begin{figure}[t!]
    \centering
    
    \begin{subfigure}{0.48\columnwidth}
        \centering
        \includegraphics[width=\linewidth]{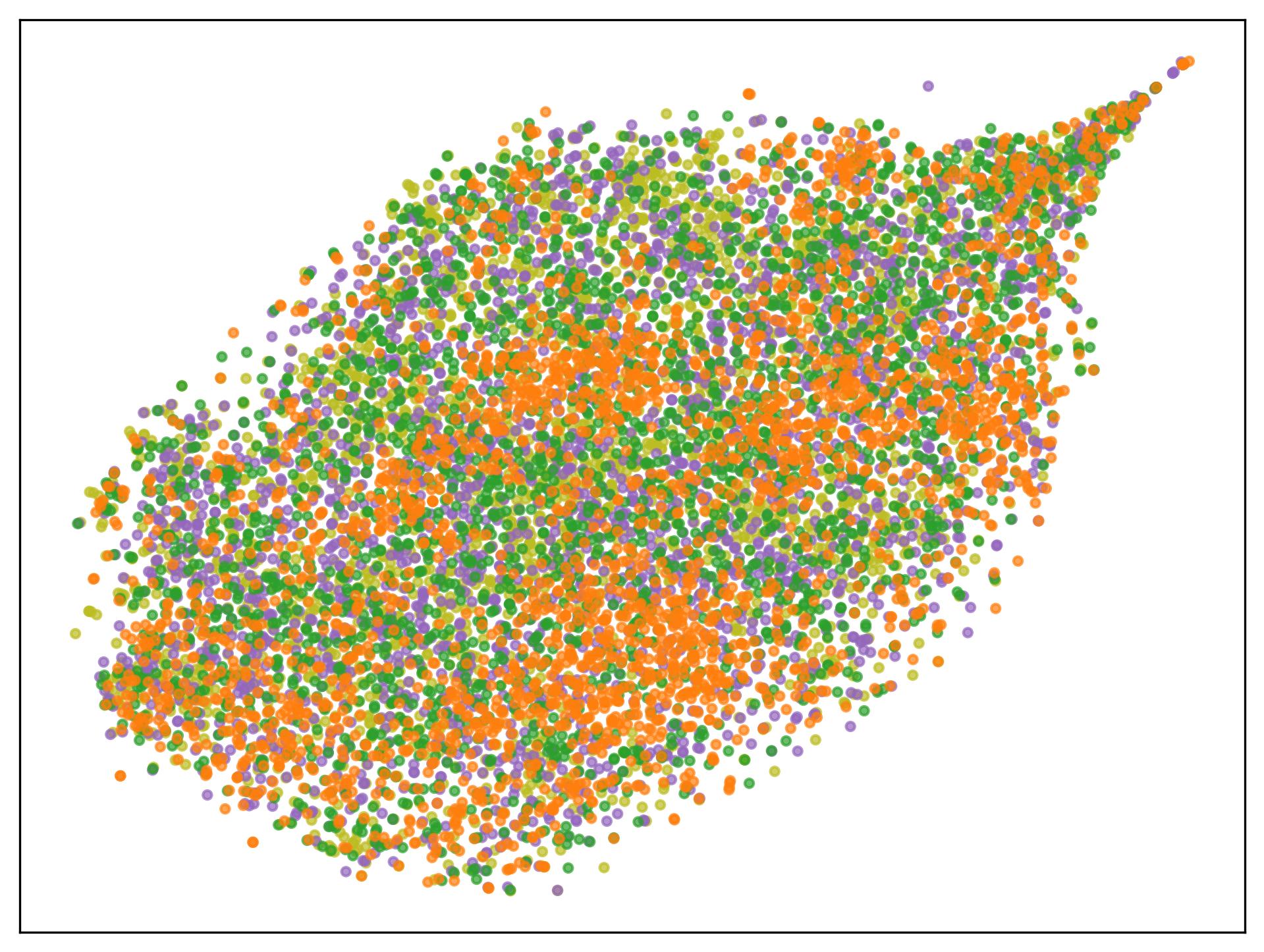}
        \caption{VPL}
    \end{subfigure}
    \begin{subfigure}{0.48\columnwidth}
        \centering
        \includegraphics[width=\linewidth]{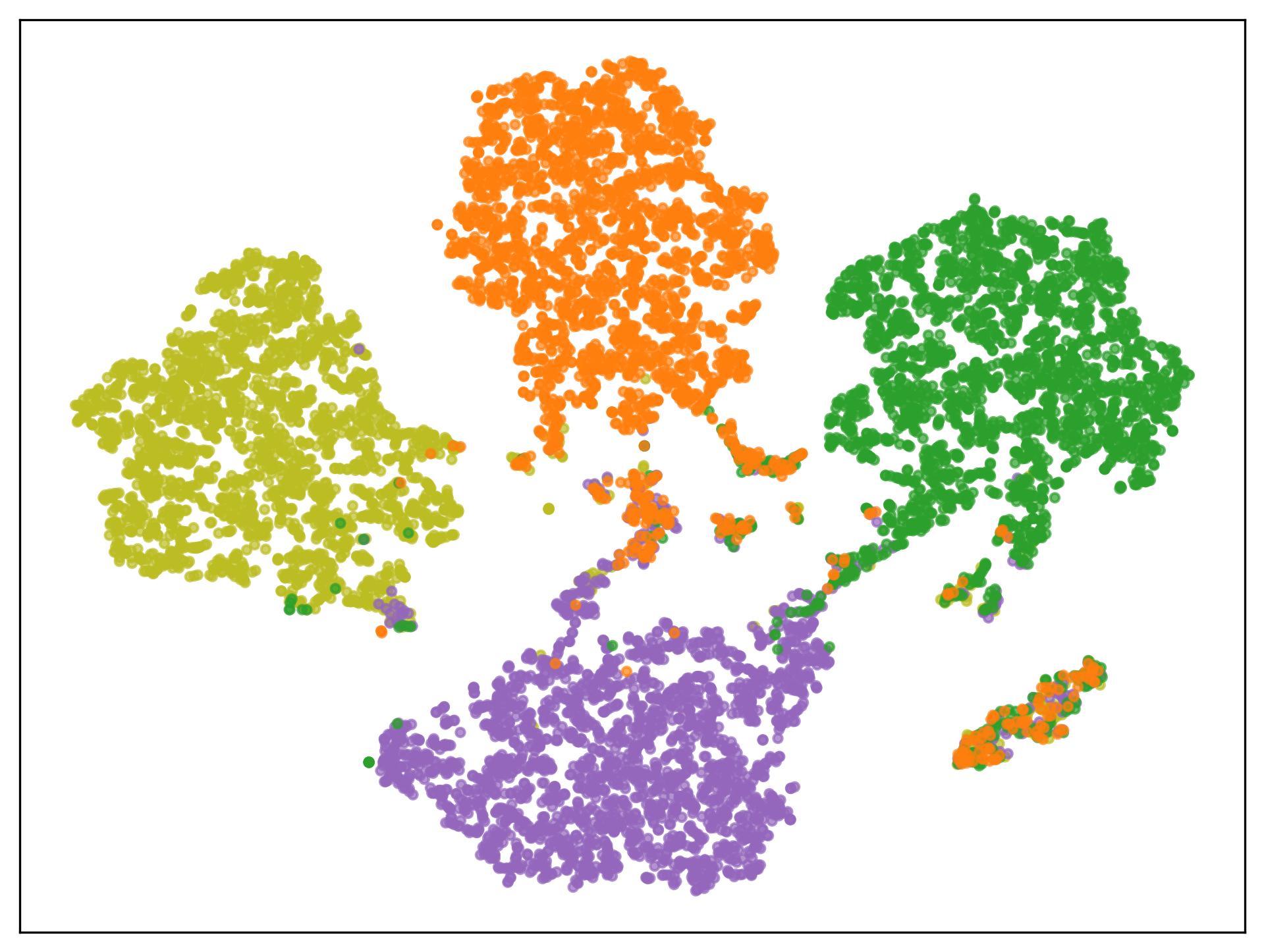}
        \caption{\ours{}}
    \end{subfigure}
    \caption{T-SNE visualization of seen user embeddings in UF-P-4 (AVG) with \texttt{gemma-2b-it}. Points are colored by their preference group.  Our method clusters users in the same group more effectively. T-SNE visualizations of other baselines are provided in \cref{fig:embedding_space}.}
    \label{fig:simple_embedding_space}
\end{figure}

There are two different approaches to utilizing the BTL framework for personalized reward models. The first approach has explored combining multiple reward models, each trained for a specific preference and later aggregated~\citep{jang2023personalized, oh2024activepreferencebasedlearningmultidimensional}. However, this approach relies on pre-trained models for different preference types, reducing flexibility. Another line of work introduces user-specific latent variables into a single BTL framework, learning personalized representations from user annotations~\citep{chen2024palpluralisticalignmentframework, poddar2024personalizing, li2024personalized, barreto2025capturing}. While this method captures individual preferences, the latent variable model does not explicitly account for relationships between users sharing similar responses. As a result, it struggles to generalize in sparse annotation settings.

To address these limitations, we propose Collaborative Preference Learning (\ours{}), which constructs a user-response bipartite preference graph from pairwise annotations and uses a graph-based collaborative filtering (GCF) framework for personalized reward modeling. 
Unlike approaches that model each user separately, GCF on the graph structure allows preference signals to propagate across users, enabling to exploit multi-hop relationships among users and responses~\citep{wang2019neural, he2020lightgcn}. \ours{} can capture diverse preferences of users even in sparse annotation settings.



When annotations are sparse, latent-variable methods face significant challenges, as the scarcity of supervisory signals makes it difficult for randomly initialized user representation encoders to converge toward semantically meaningful representations. As a result, users with similar underlying preferences can sometimes be mapped to distant points in the latent space if their annotated response pair sets do not overlap. In such cases, sparse supervision may cause semantically similar users to appear unrelated in the learned embedding space. For instance, consider three users: user 1 annotates the pairs ${(a,b), (c,d)}$, user 2 annotates ${(c,d), (e,f)}$, and user 3 annotates ${(e,f), (g,h)}$ with the same preference. Although user 1 and user 3 exhibit similar preferences, the lack of overlapping annotations provides no direct signal for aligning their representations. \ours{} addresses this issue by constructing a user–response bipartite graph and propagating preference signals through multi-hop message passing. This mechanism enables the alignment of users with disjoint annotation sets, such as user 1 and user 3, thereby providing better data efficiency and generalization. \cref{fig:simple_embedding_space} illustrates that, under sparse annotation, \ours{} produces embedding spaces in which users with identical preferences are more coherently aligned.

Based on the user embedding, we develop an LLM-based reward model that can predict the preference score of a user given input text. We adopt the mixture of LoRA experts (MoLE)~\citep{chenoctavius, chen2024llava, 10.1145/3626772.3657722} that allows parameter-efficient fine-tuning while routing different users to different paths based on the learned embedding. Specifically, we develop a user preference-aware gating function that dynamically selects the experts in the forward pass, making the LLM predict a personalized preference. 

While the reward model can predict preferences for users included in the training set, the model cannot handle newly participated \emph{unseen} users whose embeddings are unknown. To estimate the preferences of unseen users, we propose an optimization-free adaptation method. Given a few annotations from an unseen user, we exploit the existing graph to find users with similar preferences and aggregate their embeddings to represent the unseen user.

Experimental results demonstrate that CoPL consistently outperforms existing personalized reward models in both seen and unseen users. Especially, \ours{} generalizes to unseen users, maintaining high accuracy with only a few provided annotations. Embedding visualizations show that \ours{} clusters users with similar preferences more closely than competing baselines. Further ablation studies confirm that both GCF and MoLE contribute significantly to performance.

\section{Related Work}
Alignment has emerged as a crucial strategy for mitigating undesirable outcomes~\citep{dai2023safe, Yang_2024_CVPR}. Previous research has often focused on the average preference of annotators~\citep{achiam2023gpt}, ignoring the diverse preferences. To address preference diversity, recent works~\citep{jang2023personalized, oh2024activepreferencebasedlearningmultidimensional, yang2024rewards} view this problem as a soft clustering problem, where user-specific preferences are treated as mixtures of predefined preference types. Although this approach effectively handles diverse preferences, it relies on specifying several preference types in advance.

Another line of work introduces a user latent variable in the BTL framework~\citep{poddar2024personalizing, li2024personalized, chen2024palpluralisticalignmentframework}.
The main challenge lies in obtaining user representations. One approach is to treat each user embedding as learnable parameters~\citep{li2024personalized, chen2024palpluralisticalignmentframework}, and the other strategy is to train an encoder that infers embeddings from the set of annotated pairs provided by each user~\citep{poddar2024personalizing}. 

We also discuss preference learning with sparse interactions, closely related to our approach, in \cref{appendix.related_works}.

\section{Problem Formulation}



We aim to develop a reward model that can capture diverse user preferences from a limited set of preference annotations. Instead of directly defining a user’s preference, we collect pairwise comparisons indicating which item a user prefers. Let $\set{U} = \{1, \cdots, U\}$ be a set of users and $\set{X}$ be a space of LLM's responses.
To estimate the preferences of users, we first curate a \emph{survey set} $\set{S} = \{(q_i, a_i, b_i)\}_{i=1}^{R}$ consisting of predefined questions $q_i$ and two different responses $a_i, b_i \in \set{X}$ from LLMs.
For each user $u$, we first randomly sample $N_u$ number of survey items and then collect the preferences over the response pairs, resulting in \emph{preference dataset} $\set{D}_u$. 
We use $(a \succ b) \in \set{D}_u$ to denote that user $u$ prefers response $a$ over the response $b$.
Given these pairwise preferences, we aim to learn a numerical reward function
\begin{align}
    \label{eqn:reward_fn}
    f(u, r): \mathcal{U} \times \mathcal{X} \rightarrow \mathbb{R},
\end{align}
where $f(u, r)$ represents a scalar \emph{preference score} of response $r$ for user $u$. The model is trained to satisfy
$$f(u, a) > f(u, b)$$ 
for all $u$ and preference pairs $a \succ b$ observed in the data.

Following previous works~\citep{li2024personalized, poddar2024personalizing}, we consider the Bradly-Terry-Luce (BTL) choice model~\citep{bradley1952rank} with maximum likelihood estimation to train the reward function. The likelihood of user \(u\) prefers item \(a\) over \(b\) can be defined using the BTL model as
\[
p(a \succ b \mid u) 
= \frac{\exp\bigl(f(u, a)\bigr)}
       {\exp\bigl(f(u, a)\bigr) + \exp\bigl(f(u, b)\bigr)}.
\]
Conversely, if \(b\) was chosen over $a$, i.e., $a \prec b$, the likelihood is
$$p(b \succ a \mid u) = 1 - p(a \succ b \mid u).$$
Through the maximum likelihood estimation with preference data for all users, one can learn the reward function $f$ to make the reward function align with user preference. In the case of the universal preference model, user $u$ is ignored in \cref{eqn:reward_fn}~\citep{chen2024improving, achiam2023gpt, dai2023safe, bai2022training}. In practice, the user $u$ is replaced by a user embedding~\citep{poddar2024personalizing, li2024personalized, chen2024palpluralisticalignmentframework}.

\section{Method}

\begin{figure*}[t!]
    \centering
    \includegraphics[width=\linewidth]{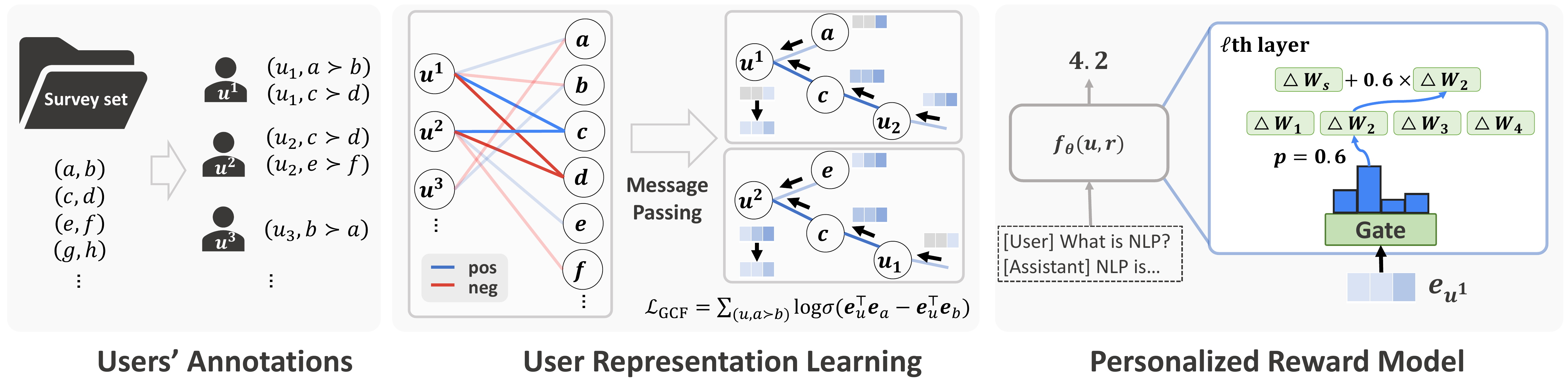} 
    \caption{An overview of \ours{}. To learn user representations, the GCF model is trained on a user-response bipartite graph.
    To build a personalized reward model, \ours{} uses the learned representations to select a user-specific expert from MoLE, enabling effective modeling of diverse preferences.}
    \label{fig:copl-overview} 
\end{figure*}

\begin{figure}[t!]
    \centering
    \includegraphics[width=\linewidth]{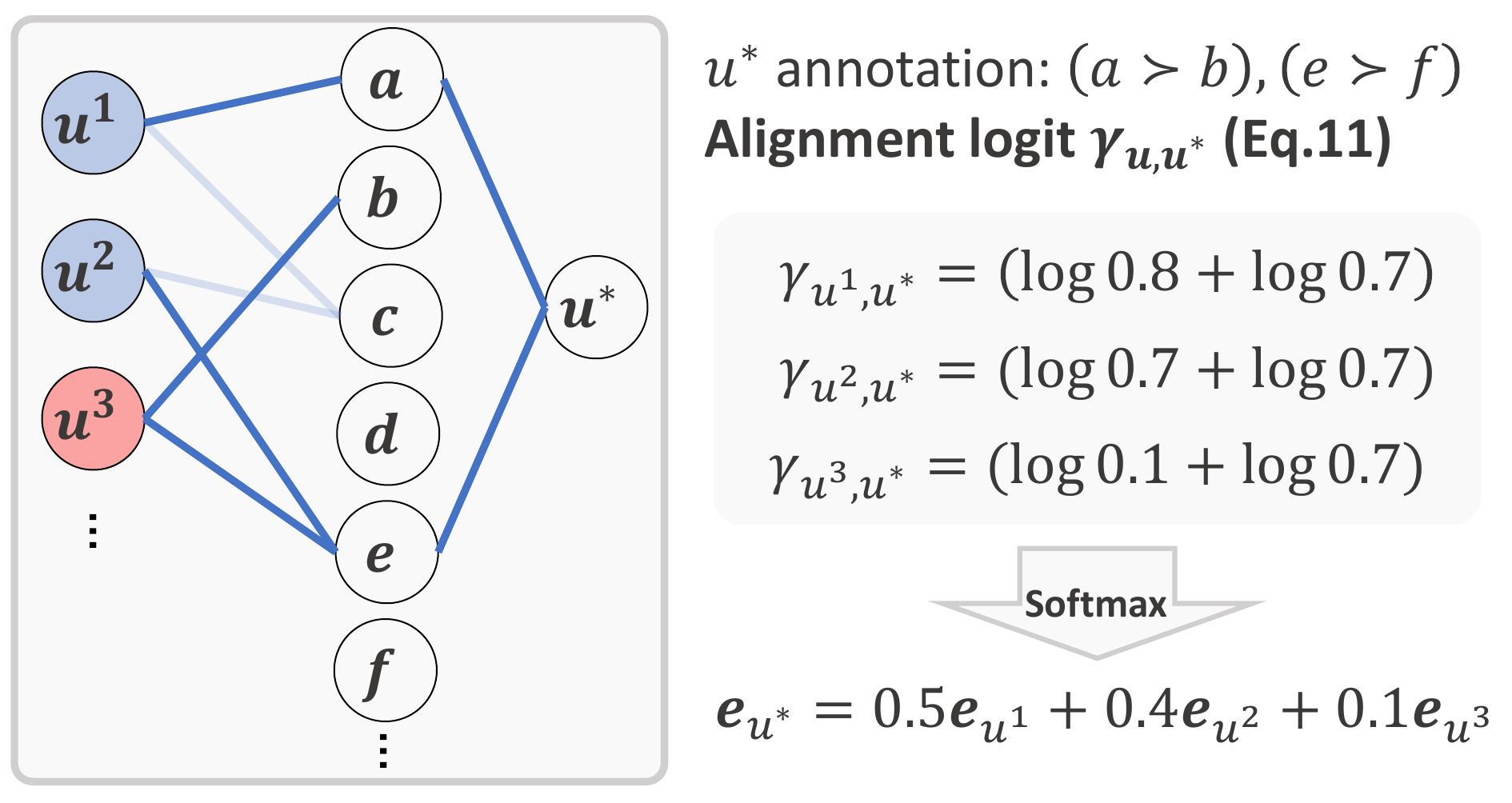} 
    \caption{Illustration of unseen user adaptation. Blue nodes are users who have similar preferences to $u^*$, and red nodes are users who have dissimilar preferences.}
    
    \label{fig:adapation-vis} 
\end{figure}

In this section, we describe our Collaborative Preference Learning (CoPL). Our approach consists of three steps: learning user representations given preference data, construction of personalized reward models, and adaptation to unseen (new) users at test time.  \Cref{fig:copl-overview} illustrates the first two steps, and \autoref{fig:adapation-vis} the last step.

\subsection{User Representation Learning}

Users who share similar preferences are likely to respond to similar responses. When the number of annotated responses is very small, it is unlikely to annotate the same responses between users. However, if we exploit multi-hop relations between users and responses, we may estimate user preference accurately. In fact, the exploitation of the relationship between users and items is the key idea behind graph-based collaborative filtering (GCF).

The preference dataset for all users can be naturally converted into a bipartite graph, where each user and response is represented as a node, and an edge between a user and a response represents the user's preference over the response, as illustrated in \cref{fig:copl-overview}. The edge can have two different types: positive or negative, indicating whether a user prefers the response or not.

Given a bipartite graph, we design a message-passing algorithm to update user and response representations. Let $\embedding{e}_u \in \mathbb{R}^d$ be an embedding vector of user $u$, and $\embedding{e}_r \in \mathbb{R}^d$ be an embedding vector of response $r$. Since there are two different edge types, we use different parameterizations for each type. Let $\set{N}^+_u$ be a set of positive edges and $\set{N}^-_u$ be a set of negative edges from user $u$. Similary, we can define $\set{N}^+_r$ and $\set{N}^-_r$ for response $r$. Given user and response embeddings at layer $\ell$, the message passing computes a message from neighborhood responses to the user as

\begin{align} \scriptscriptstyle
&\embedding{m}_u^{+} = \sum_{r\in \set{N}^+_u}
\alpha_{u, r}\Bigl( 
W_1^{(\ell)} \embedding{e}_r^{(\ell)} +W_2^{(\ell)} (\embedding{e}_r^{(\ell)} \odot \embedding{e}_u^{(\ell)})  \Bigr), \nonumber\\
&\embedding{m}_u^{-} = \sum_{r \in \set{N}^-_u} \beta_{u,r}\Bigl( 
W_3^{(\ell)} \embedding{e}_r^{(\ell)} + 
W_4^{(\ell)} (\embedding{e}_r^{(\ell)} \odot \embedding{e}_u^{(\ell)}) \Bigr), \nonumber\\
&\embedding{m}^{(\ell)}_u = W_{\text{self}}^{(\ell)}\,\embedding{e}_u^{(\ell)} \;+\; \embedding{m}_u^{+} \;+\; \embedding{m}_u^{-},
\label{eq:message}
\end{align}
where $ W_1^{(\ell)}, W_2^{(\ell)}, W_3^{(\ell)}, W_4^{(\ell)}, W_{\text{self}}^{(\ell)}\in\mathbb{R}^{d\times d}$ are parameter matrices, $\odot$ is element-wise multiplication, and $\alpha_{u,r}$ and $\beta_{u,r}$ are normalization factors, set to $\frac{1}{\sqrt{|\set{N}^+_u||\set{N}^+_r|}}$ and $\frac{1}{\sqrt{|\set{N}^-_u||\set{N}^-_r|}}$, respectively. 
Then, the user embedding is updated with the aggregated message $\embedding{m}^{(\ell)}_u$:
\begin{align}
\embedding{e}_u^{(\ell+1)} = {\psi} \bigl(\embedding{m}^{(\ell)}_u\bigr),
\label{eq:update-user}
\end{align}
where $\psi(\cdot)$ is a non-linear activation. The response embedding $\embedding{e}_r^{(\ell)}$ is updated with analogous process. We randomly initialize the user and response embeddings at the first layer and then fine-tune the embeddings through training. The update steps for the response embeddings are provided in \cref{appendix.message}.

After $L$ propagation steps, user and response embeddings accumulate information from their local neighborhood.
Given the final user embedding $\embedding{e}_u^{(L)}$ and response embedding $\embedding{e}_r^{(L)}$, we use the inner product between the embeddings as a predicted preference :
\begin{equation}
s_{u,r} \;=\; \bigl(\embedding{e}_u^{(L)}\bigr)^\top \bigl(\embedding{e}_r^{(L)}\bigr).
\label{eq:user-item-score}
\end{equation}
With the score function, the GNN is trained on preference data $\mathcal{D}_{{u}}$ {for all users} by minimizing the following loss function:
\begin{align}
\label{eqn:gcf}
&\mathcal{L}_\text{GCF}(\theta) := \\\nonumber
&\sum_{u \in \set{U}}\sum_{(a \succ b)\in\mathcal{D}_u}
- \log \sigma\left(s_{u,a} - s_{u,b}\right)
+ \lambda \|\theta\|_2^2,
\end{align}
where {$\sigma(\cdot)$ denotes a sigmoid function,} $\lambda$ is a regularization hyper-parameter and $\theta$ represents all trainable parameters, including weights of the propagation layers and initial embeddings of the users $\embedding{e}_u^{(0)}$ and responses $\embedding{e}_r^{(0)}$.

\subsection{Personalized Reward Model with User Representations}
\label{sec:prm}
Based on the learned user embeddings $\embedding{e}_u^{(L)}$, we build a reward model that can accommodate the preferences of diverse users. We use an LLM-based reward function:
\begin{align}
   f_\phi(\embedding{e}_u, r): \mathbb{R}^d \times \set{X} \rightarrow \mathbb{R}
\end{align}
where $f$ is an LLM parameterized by $\phi$ taking user embedding $\embedding{e}_u$ and  the response $r$ as inputs and predicts preference score. Unlike the response, the user embedding is not used as an input token. Instead, it is used in the gating mechanism described below. To learn the reward model, we can employ the BTL model, resulting in the maximum likelihood objective:
\begin{align}
\set{L}_\text{RM}(\phi) = \sum_{u}\sum_{(a \succ b) \in \set{D}_u } \log p_\phi(a \succ b \mid \embedding{e}_u)
\end{align}
However, naively optimizing this objective starting from a pretrained LLM requires fine-tuning billions of parameters. Moreover, different preferences of users result in conflicting descent directions of the model parameters, resembling a multi-task learning scenario.

\paragraph{Mixture of LoRA experts for personalized reward function.} For an efficient parameter update while minimizing the negative effect of diverse preferences, we adopt the mixture of LoRA experts (MoLE)~\citep{hu2021lora, 10.1145/3626772.3657722} into our framework. 
MoLE is proposed to maximize the benefit of the mixture of experts (MoE) while maintaining efficient parameter updates. With MoLE, the model parameter matrix $W$ is decomposed into pretrained and frozen $W_0$ and trainable $\Delta W$, i.e., $W = W_0 + \Delta W$. $\Delta W$ is further decomposed into a shared LoRA expert $A_s \in\mathbb{R}^{d_\text{out}\times n}, B_s \in\mathbb{R}^{n\times d_\text{in}}$, which is used across all users, and $M$ individual LoRA experts $\{A_i, B_i\}_{i=1}^{M}$ with the same dimensionality of the shared expert.
Formally, this can be written as
\begin{align}
\Delta W_u = A_sB_s + \sum_{i=1}^M w_i A_iB_i,
\end{align}
where $w_i \in [0,1]$ denotes the importance of expert $i$.

To adopt the different preferences of users, we define a user-dependent gating mechanism to model the importance parameter $w_i$. 
For each user $u$, a gating function \(g: \mathbb{R}^d \to \mathbb{R}^M\) maps \(\embedding{e}_u^{(L)}\) to expert-selection logits:
\begin{equation}
\mathbf{z} = g\bigl(\embedding{e}_u^{(L)}\bigr).
\end{equation}
We convert these logits $\mathbf{z}$ into gating weight $w_i$ by selecting the top one expert from the logits:
\begin{align}
    w_i = 
        \begin{cases} 
            \frac{\exp(z_i/\tau)}{\sum_{j=1}^{M} \exp(z_j/\tau)} & \text{if}\; i = \arg\max_i z_i \\
            0 &\text{otherwise,}
        \end{cases}
\end{align}
where $\tau$ is a temperature parameter. In practice, one can use top-$k$ experts, but we could not find a significant difference in our experiments. For computational efficiency, we keep the top one expert.

\subsection{Optimization-free User Adaptation} 
While we can predict a preference score of unseen responses for a known user, the reward model trained in \cref{sec:prm} cannot be used to predict the preference of users who have not been observed during training. To estimate the embeddings of unseen users, we propose an optimization-free adaptation approach.


Let $u^*$ be an unseen user who annotates a small set of response pairs. Under the assumption that users who have similar responses have similar preferences, we can estimate the embedding of an unseen user by taking an embedding of users with similar tastes. For example, if both user $u^*$ and $u$ share positive preference over the same response $r$, then we can use the embedding of $u$ to approximate that of $u^*$.
Based on this intuition, we propose the following optimization-free adaptation strategy for unseen user embedding:
\begin{align}
&\embedding{e}_{u^*}^{(L)} \;=\; 
\sum_{u \in \set{N}^{+}_{u{^*}}(k
)} w_{u,u^*} \embedding{e}_{u}^{(L)},
\end{align}
where $\set{N}^{+}_{u{^*}}(k)$ is a set of $k$-hop neighborhood\footnote{$k$ must be an even number to aggregate only the user embeddings.} of user $u^*$ connected by only positive edges, and $w_{u,u^*}$ is a normalized alignment score between $u$ and $u^*$.
The normalized alignment score $w_{u,u^*}$ is defined as
$$
w_{u,u^*} = \frac{\exp(\gamma_{u,u^*}/\kappa)}{\sum_{\tilde{u}\in \set{N}^{+}_{u^*}(k)} \exp(\gamma_{\tilde{u},u^*}/\kappa)},
$$
where
$$
\gamma_{u,u^*} = \sum_{(a\succ b)\in \set{D}_{u^*}} \log\sigma(s_{u,a} - s_{u,b}),
$$
where $s_{u,i}$ is an inner product between user and response embeddings, $\kappa$ is a temperature parameter, and $\gamma_{u,u^*}$ is an alignment score between user $u$ and $u^*$. 
Intuitively, $\gamma_{u,u^*}$ measures how well the \emph{predicted preference} of user $u$ aligns with the \emph{annotated preference} provided by user $u^*$. If the preferences of both users align well, $\gamma_{u,u^*}$ is large. Consequently, their embeddings become similar to each other.
By collecting embeddings of well-aligned neighborhood users, we can obtain embeddings of user $u^*$ without having further optimization.

\section{Experiments}
In this section, we empirically verify the performance of CoPL across various scenarios.

\begin{table}[t]
\centering
\resizebox{\columnwidth}{!}{
\begin{tabular}{lcccc}
\toprule
\textbf{Dataset} & \textbf{TL;DR} & \textbf{UF-P-2} & \textbf{UF-P-4} & \textbf{PersonalLLM} \\
\midrule
Size of survey set              & 19,824 & 25,993 & 25,993 & 14,435 \\
\# of preference groups     & 2      & 2      & 4      & $\infty$      \\
\# of annotations per user      & 8      & 8      & 16     & 16     \\
\# of users per group           & 5,000  & 5,000  & 2,500  & -  \\
\bottomrule
\end{tabular}
}
\caption{Statistics of the datasets. We report the \emph{average} number of annotations per user. All users have different preferences in PersonalLLM.}
\label{tab:dataset_statistic}
\end{table}

\subsection{Experimental Settings}

\paragraph{Datasets.}
We employ three datasets, including {TL;DR}~\citep{stiennon2020learning, chen2024palpluralisticalignmentframework}, {UltraFeedback-P (UF-P)}~\citep{poddar2024personalizing}, and {PersonalLLM}~\citep{zollo2024personalllm}, that explicitly capture diverse user preferences rather than assuming a single dominant preference.
We briefly describe the key characteristics of these datasets below.

Following prior work~\citep{chen2024palpluralisticalignmentframework, li2024personalized}, we define two user groups in the TL;DR dataset: one group prefers short summaries, and the other favors long summaries. 
We create two environments with the UF-P dataset: UF-P-2, dividing users into two groups based on their preference, and UF-P-4, dividing users into four groups.
In PersonalLLM~\citep{zollo2024personalllm}, user preferences are modeled as a mixture of four preference dimensions where weight vectors are drawn from a Dirichlet distribution with $\alpha=0.1$.
Additional details on their construction and properties can be found in \cref{appendix.dataset}.

\begin{table*}[t!]
\centering
\resizebox{\textwidth}{!}{
\begin{tabular}{llrrrrrrrr}
\toprule
& & \multicolumn{2}{c}{TL;DR} & \multicolumn{2}{c}{UF-P-2} & \multicolumn{2}{c}{UF-P-4} & \multicolumn{2}{c}{PersonalLLM} \\
\cmidrule(lr){3-4}\cmidrule(lr){5-6}\cmidrule(lr){7-8}\cmidrule(lr){9-10}
& & \multicolumn{1}{c}{ALL} & \multicolumn{1}{c}{AVG} & \multicolumn{1}{c}{ALL} & \multicolumn{1}{c}{AVG} & \multicolumn{1}{c}{ALL} & \multicolumn{1}{c}{AVG} & \multicolumn{1}{c}{ALL} & \multicolumn{1}{c}{AVG} \\
\midrule
\multirow{7}{*}{\rotatebox[origin=c]{90}{Seen}} 
& G-Oracle   
& $73.06_{\pm0.23}$ & $73.06_{\pm0.23}$ 
& $64.53_{\pm0.14}$ & $64.53_{\pm0.14}$ 
& $61.52_{\pm0.13}$ & $61.52_{\pm0.13}$ 
& N/A & N/A \\ \cmidrule(lr){2-10}

& Uniform  
& $49.62_{\pm0.09}$ & $49.62_{\pm0.09}$ 
& $61.82_{\pm0.16}$ & $61.82_{\pm0.16}$ 
& $56.15_{\pm0.22}$ & $56.15_{\pm0.22}$ 
& $62.91_{\pm0.07}$ & $62.91_{\pm0.07}$ \\

& I2E      
& $49.93_{\pm0.23}$ & $49.74_{\pm0.06}$ 
& $61.48_{\pm0.18}$ & $61.49_{\pm0.70}$ 
& $57.21_{\pm0.37}$ & $57.44_{\pm0.37}$ 
& $65.74_{\pm0.04}$ & $65.77_{\pm0.05}$ \\

& I2E$_{\text{proxy}}$
& $49.80_{\pm0.16}$ & $49.54_{\pm0.13}$ 
& $61.43_{\pm0.56}$ & $61.33_{\pm0.61}$ 
& $56.78_{\pm0.14}$ & $57.14_{\pm0.31}$ 
& $65.66_{\pm0.11}$ & $65.77_{\pm0.05}$ \\

& VPL      
& $49.52_{\pm0.14}$ & $49.44_{\pm0.21}$ 
& $61.11_{\pm0.16}$ & $61.86_{\pm0.84}$ 
& $56.04_{\pm1.71}$ & $56.77_{\pm0.38}$ 
& $70.84_{\pm0.18}$ & $67.95_{\pm0.21}$ \\

& PAL      
& $50.12_{\pm0.13}$ & $50.15_{\pm0.15}$ 
& $59.95_{\pm0.04}$ & $61.53_{\pm0.22}$ 
& $56.95_{\pm0.13}$ & $57.37_{\pm0.14}$ 
& $66.25_{\pm0.35}$ & $66.29_{\pm0.06}$ \\

& \ours{}  
& $\mathbf{96.58}_{\pm0.09}$ & $\mathbf{96.19}_{\pm0.02}$
& $\mathbf{63.81}_{\pm0.16}$ & $\mathbf{63.45}_{\pm0.38}$
& $\mathbf{62.57}_{\pm0.38}$ & $\mathbf{62.08}_{\pm0.27}$
& $\mathbf{74.85}_{\pm0.17}$ & $\mathbf{74.37}_{\pm0.03}$ \\
\midrule \midrule

\multirow{7}{*}{\rotatebox[origin=c]{90}{Unseen}} 
& G-Oracle   
& $72.55_{\pm1.79}$ & $72.55_{\pm1.79}$ 
& $64.66_{\pm1.10}$ & $64.66_{\pm1.10}$ 
& $61.33_{\pm0.35}$ & $61.33_{\pm0.35}$ 
& N/A & N/A \\ \cmidrule(lr){2-10}

& Uniform  
& $50.11_{\pm0.36}$ & $50.11_{\pm0.36}$ 
& $62.82_{\pm0.59}$ & $62.82_{\pm0.59}$ 
& $55.65_{\pm0.61}$ & $55.65_{\pm0.61}$ 
& $62.97_{\pm0.07}$ & $62.97_{\pm0.07}$ \\

& I2E      
& $49.85_{\pm0.38}$ & $49.16_{\pm0.82}$ 
& $61.67_{\pm0.82}$ & $59.52_{\pm0.51}$ 
& $56.42_{\pm0.41}$ & $56.75_{\pm0.68}$ 
& $65.79_{\pm0.18}$ & $66.11_{\pm0.24}$ \\

& I2E$_{\text{proxy}}$
& $49.75_{\pm0.94}$ & $49.12_{\pm0.57}$ 
& $62.30_{\pm0.54}$ & $61.70_{\pm0.63}$ 
& $56.00_{\pm1.15}$ & $56.50_{\pm0.34}$ 
& $65.49_{\pm0.10}$ & $65.79_{\pm0.04}$ \\

& VPL      
& $49.40_{\pm0.88}$ & $49.31_{\pm0.57}$ 
& $60.83_{\pm0.40}$ & $62.62_{\pm0.49}$ 
& $54.03_{\pm1.54}$ & $56.13_{\pm0.57}$ 
& $71.31_{\pm0.58}$ & $68.55_{\pm0.47}$ \\

& PAL      
& $49.48_{\pm0.86}$ & $49.64_{\pm0.55}$ 
& $59.83_{\pm0.69}$ & $61.71_{\pm0.31}$ 
& $57.07_{\pm0.22}$ & $57.13_{\pm0.33}$ 
& $65.94_{\pm0.11}$ & $66.40_{\pm0.03}$ \\

& \ours{}
& $\mathbf{96.71}_{\pm0.25}$ & $\mathbf{96.21}_{\pm0.14}$
& $\mathbf{63.92}_{\pm0.54}$ & $\mathbf{63.26}_{\pm0.51}$
& $\mathbf{61.62}_{\pm0.10}$ & $\mathbf{61.97}_{\pm0.35}$
& $\mathbf{75.69}_{\pm0.22}$ & $\mathbf{75.49}_{\pm0.03}$ \\
\bottomrule
\end{tabular}
}
\caption{Accuracy of reward models on unseen annotated pairs. The results report performance on \emph{Seen users} encountered during training and on \emph{Unseen users}. \textbf{Bold} represents the best result, except for G-Oracle. These results are based on \texttt{gemma-2b-it}. Additional results using \texttt{gemma-7b-it} and \texttt{gemma2-27b-it} are represented in \cref{tab:combined_7b} and \cref{ablation.large-rm}, respectively.}
\label{tab:combined_2b}
\end{table*}

We divide $10,000$ users evenly into the predefined number of preference groups. For all datasets, we curate two different versions, denoted as ALL and AVG, representing two different annotation sampling strategies.
For TL;DR/UF-P-2 (ALL), each user provides exactly 8 annotations, while for TL;DR/UF-P-2 (AVG), each user’s annotation count is uniformly sampled from 1 to 15, averaging to 8. 
Similarly, in UF-P-4/PersonalLLM (ALL), each user provides exactly 16 annotations, and in UF-P-4/PersonalLLM (AVG), the count is uniformly sampled from 1 to 31, averaging to 16.
\autoref{tab:dataset_statistic} summarizes the key statistics.

\paragraph{Baselines.} We evaluate six baselines to benchmark. First, we use a uniform preference model (Uniform) trained on all annotations via BTL. Additionally, we consider four personalized reward models: I2E, I2E$_\text{{proxy}}$~\citep{li2024personalized}, VPL~\citep{poddar2024personalizing}, and PAL~\cite{chen2024palpluralisticalignmentframework}. Finally, we include a group-wise Oracle (G-Oracle), which has access to user group information and all annotations in the survey set, and trains a separate reward function in \cref{eqn:reward_fn} for each preference group. Note that we do not have the G-Oracle for PersonalLLM since the users are not categorized into a fixed number of preference groups. The details of each model are provided in \cref{appendix.baseline}.

\paragraph{Training and evaluation details.}
For reward function training, we utilize two LLM backbones: \texttt{gemma-2b-it} and \texttt{gemma-7b-it}~\citep{team2024gemma}. 
Our model uses one shared LoRA, eight LoRA experts, each with a rank of eight, and a two-layer MLP for the gating function. The other baselines, e.g., Uniform, I2E, VPL, PAL, and G-Oracle, use a LoRA rank of 64. Other training details, such as hyper-parameters and model architecture, are provided in  \cref{appendix.training}. All experiments, including additional analysis, are repeated three times with different seeds.

We report reward model accuracy on unseen test pairs that are not in the survey set. 
We evaluate performance for both seen and unseen users. For seen user experiments, each user is assigned 10 test pairs, and accuracy is calculated over all seen users. We fix the number of unseen users at 100, evenly distributed across preference groups. To adapt the reward model for each unseen user, we provide 8 annotations in TL;DR/UF-P-2 (ALL/AVG) and 16 annotations in UF-P-4/PersonalLLM (ALL/AVG), followed by evaluation on 50 test pairs per unseen user. \ours{} uses 2-hop neighbors for unseen user adaptation.

    

\begin{table*}
\resizebox{\textwidth}{!}{
\centering
\begin{tabular}{lrrrrrrr}
\toprule
& \multicolumn{1}{c}{G-Oracle} & \multicolumn{1}{c}{Uniform} & \multicolumn{1}{c}{I2E} & \multicolumn{1}{c}{I2E$_\text{{proxy}}$} & \multicolumn{1}{c}{VPL} &  \multicolumn{1}{c}{PAL} & \multicolumn{1}{c}{\ours{}} \\ \midrule
Common &  $71.86_{\pm0.14}$  &   $\textbf{74.52}_{\pm0.45}$  &   $73.94_{\pm0.21}$   &  $74.15_{\pm1.53}$ &  $72.73_{\pm1.00}$   & $70.82_{\pm0.17}$ & $71.23_{\pm1.63}$   \\
Controversial &   $57.68_{\pm0.27}$   &  $49.86_{\pm0.30}$  &  $49.61_{\pm0.05}$  &  $49.86_{\pm0.06}$   &  $50.26_{\pm0.44}$ & $49.79_{\pm0.12}$ & $\textbf{56.89}_{\pm1.56}$  \\ \midrule
Total &   $64.53_{\pm0.14}$   &  $61.82_{\pm0.16}$  &  $61.48._{\pm0.18}$  &  $61.59_{\pm0.79}$   & $61.11_{\pm0.32}$ &  $59.95_{\pm0.04}$  & $\textbf{63.81}_{\pm0.15}$  \\ \bottomrule
\end{tabular}}
\caption{Accuracy of reward models on UF-P-2 (ALL) with \texttt{gemma-2b-it}, broken down by pair type. 
\emph{Common} refers to pairs for which the two preference groups provide the same preference label, 
\emph{Controversial} refers to pairs labeled differently by the two groups, 
and \emph{Total} encompasses all pairs. These categories reflect how diverse user preferences affect the performance of reward models. \textbf{Bold} represents the best result, except with G-Oracle.}
\label{tab:Analysis.con-common}
\end{table*}

\subsection{Results}

\cref{tab:combined_2b} presents accuracy for both seen and unseen users. \ours{}  consistently outperforms other baselines, except for G-Oracle, in both seen user and unseen user experiments. Notably, \ours{} surpasses the performance of G-Oracle on TL;DR and UF-P-4, demonstrating the advantage of multi-task learning. In the PersonalLLM, \ours{} remains robust across the ALL and AVG, whereas VPL suffers from performance degradation in a more realistic AVG setting. These findings are consistent with \citet{ju2024does}, which theoretically shows that message-passing can help users with limited interactions in collaborative filtering. In unseen user experiments, \ours{} achieves accuracy comparable to the seen user setting, indicating the effectiveness of our unseen user adaptation. 

\cref{fig:simple_embedding_space} illustrates the learned user embeddings for UF-P-4 (AVG), selected as the most challenging environment among those with distinct groups. The figure shows that GNN-based representation learning successfully captures preference similarities, despite the limited annotations per user.

\subsection{Analysis}

\paragraph{Analysis of performance in UF-P-2.}
In \cref{tab:combined_2b}, all models appear capable of representing diverse preferences, surprisingly including the uniform models in UF-P-2 (ALL/AVG). To investigate further, we divide the test pairs of UF-P-2 into \emph{common} and \emph{controversial} categories, where common pairs have identical annotations from both preference groups, and controversial pairs differ. Focusing on the seen user results in UF-P-2 (ALL) with \texttt{gemma-2b-it} from \cref{tab:combined_2b}, we break down the accuracy in \cref{tab:Analysis.con-common}. The results indicate that baselines, except G-Oracle, struggle with controversial pairs, suggesting a tendency to capture only the common preference across all users. By contrast, our method achieves comparable performance to G-Oracle on controversial pairs while preserving high accuracy on common pairs.

\begin{figure}[t!]
    \centering
    \begin{subfigure}[b]{0.49\columnwidth}
        \centering
        \includegraphics[width=\textwidth]{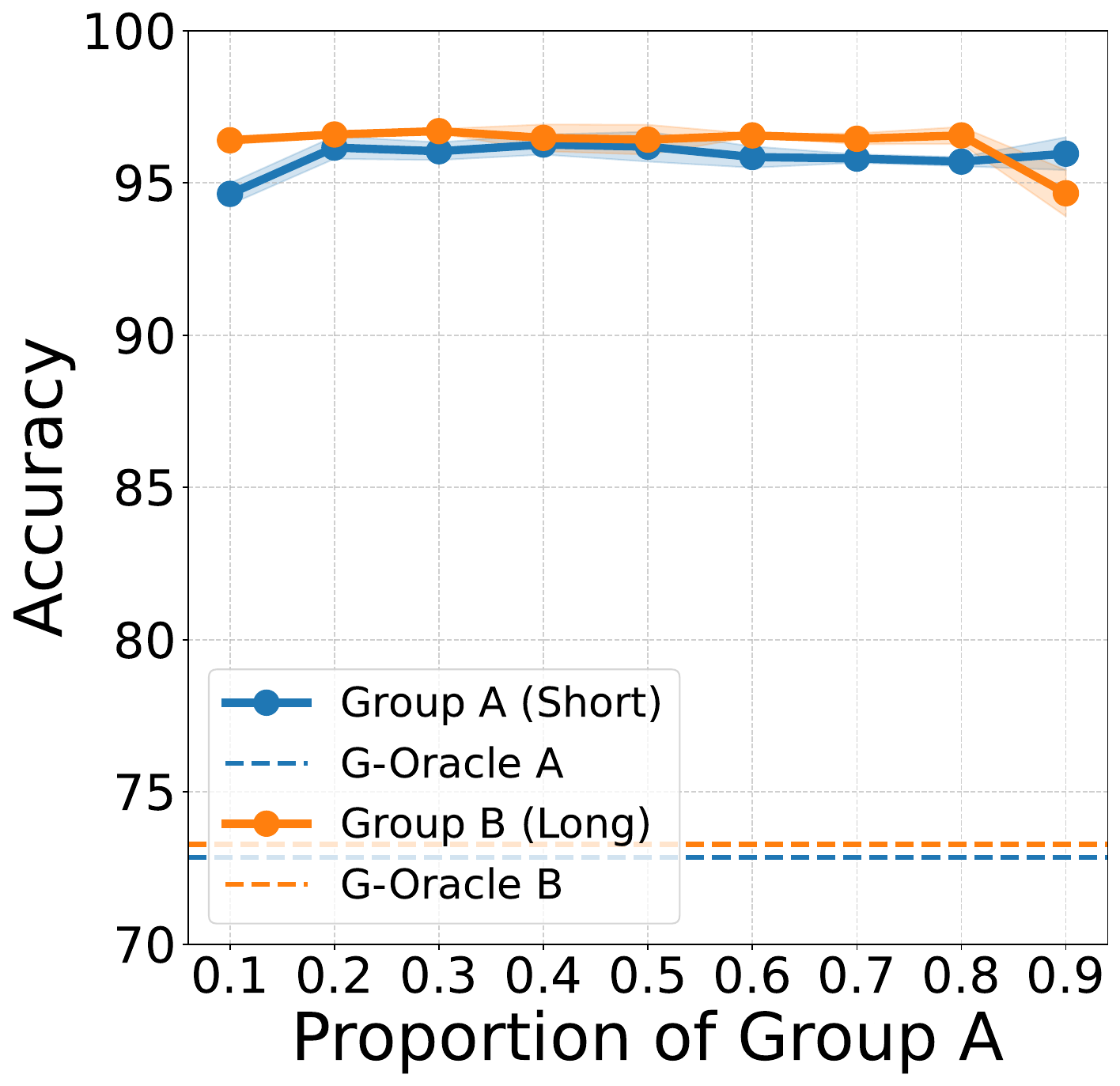}
        \caption{TL;DR (AVG)}
    \end{subfigure}
    \hfill
    \begin{subfigure}[b]{0.475\columnwidth}
        \centering
        \includegraphics[width=\textwidth]{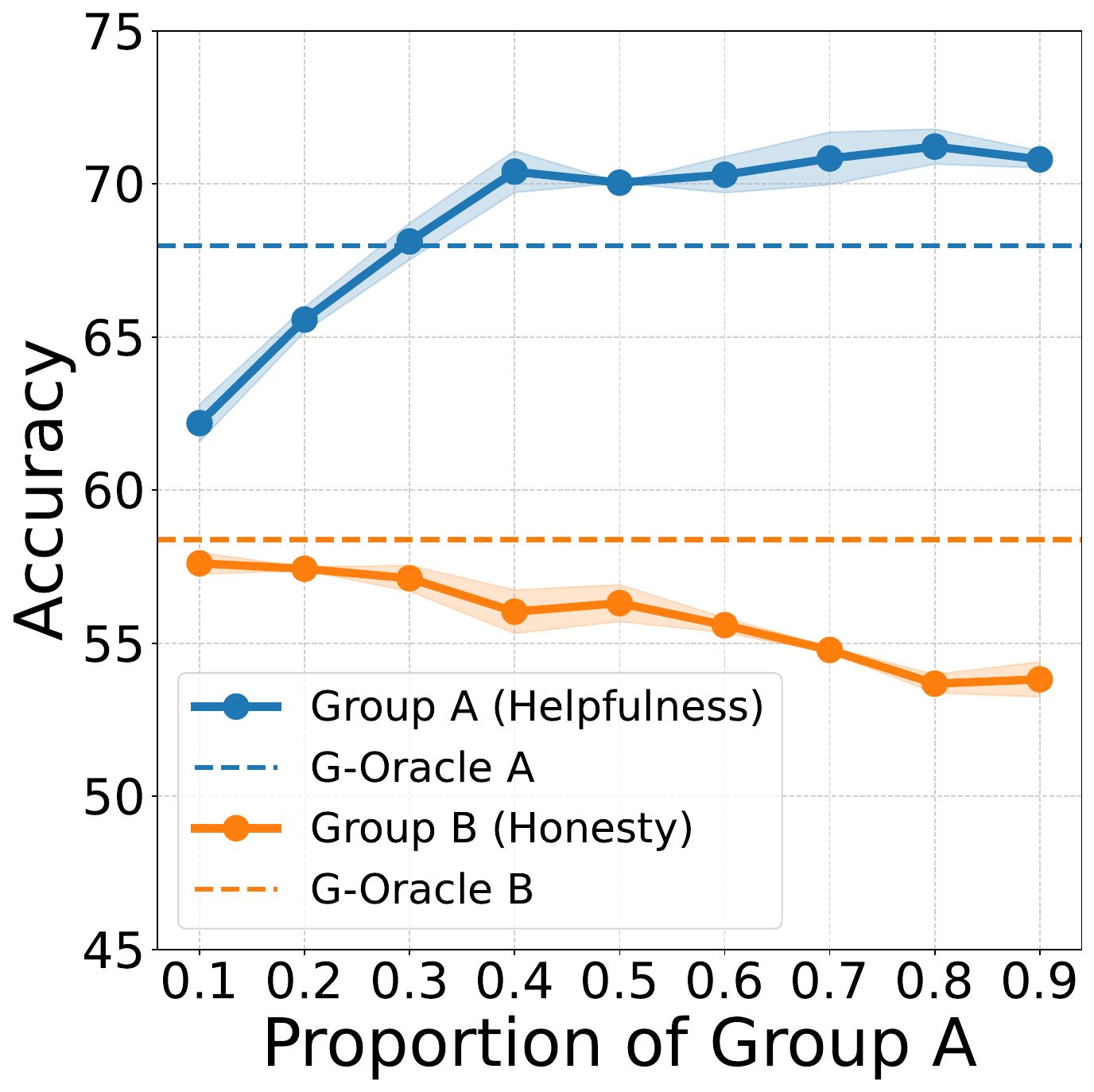}
        \caption{UF-P-2 (AVG)}
    \end{subfigure}
\caption{{Group-wise accuracy of reward models with Gemma-2b-it in TL;DR (AVG) and UF-P-2 (AVG), varying the ratio of group size (A:B) from 1:9 to 9:1 with the total number of users fixed at 10,000.}}
\label{fig:ablation.group-ratio}
\end{figure}

\paragraph{{Performance under imbalanced group distributions.}}
We vary the group proportion from 1:9 to 9:1 on the TL;DR (AVG) and UF-P-2 (AVG) datasets. As shown in \cref{fig:ablation.group-ratio}, CoPL consistently captures both majority and minority preferences, maintaining stable accuracy for the short- and long-summary groups on TL;DR. On UF-P-2, CoPL still reflects diverse preferences, but the gap relative to the balanced 5{:}5 setting widens as the ratio becomes more skewed. Majority accuracy rises while minority accuracy falls, showing majority bias under imbalance. The lower absolute accuracy for the honesty group reflects the inherent difficulty of that preference, which remains consistent with the G-oracle results. The difference between TL;DR and UF-P-2 is also explained by UF-P-2 containing common pairs on which both groups agree, which reduces the distinct signal from the minority.

    {\cref{fig:embedding_space_ratios,fig:group-ratio-ufp2} show the learned user embeddings for TL;DR and UF-P-2. CoPL preserves well-separated clusters aligned with group identities even under extreme imbalance. Thus, while minority group accuracy may drop, the representation space remains robust to group structure. To mitigate majority bias, we can apply loss reweighting, such as focal loss \citep{lin2017focal,subramanian2021fairness}, to emphasize underrepresented groups in the reward model training. }

{In the four-group setting, additional UF-P-4 (AVG) results exhibit the same trend, as reported in \cref{appendix.experimental_results}.}

\begin{figure}[t]
    \centering
    \begin{subfigure}[b]{0.48\columnwidth}
        \centering
        \includegraphics[width=\textwidth]{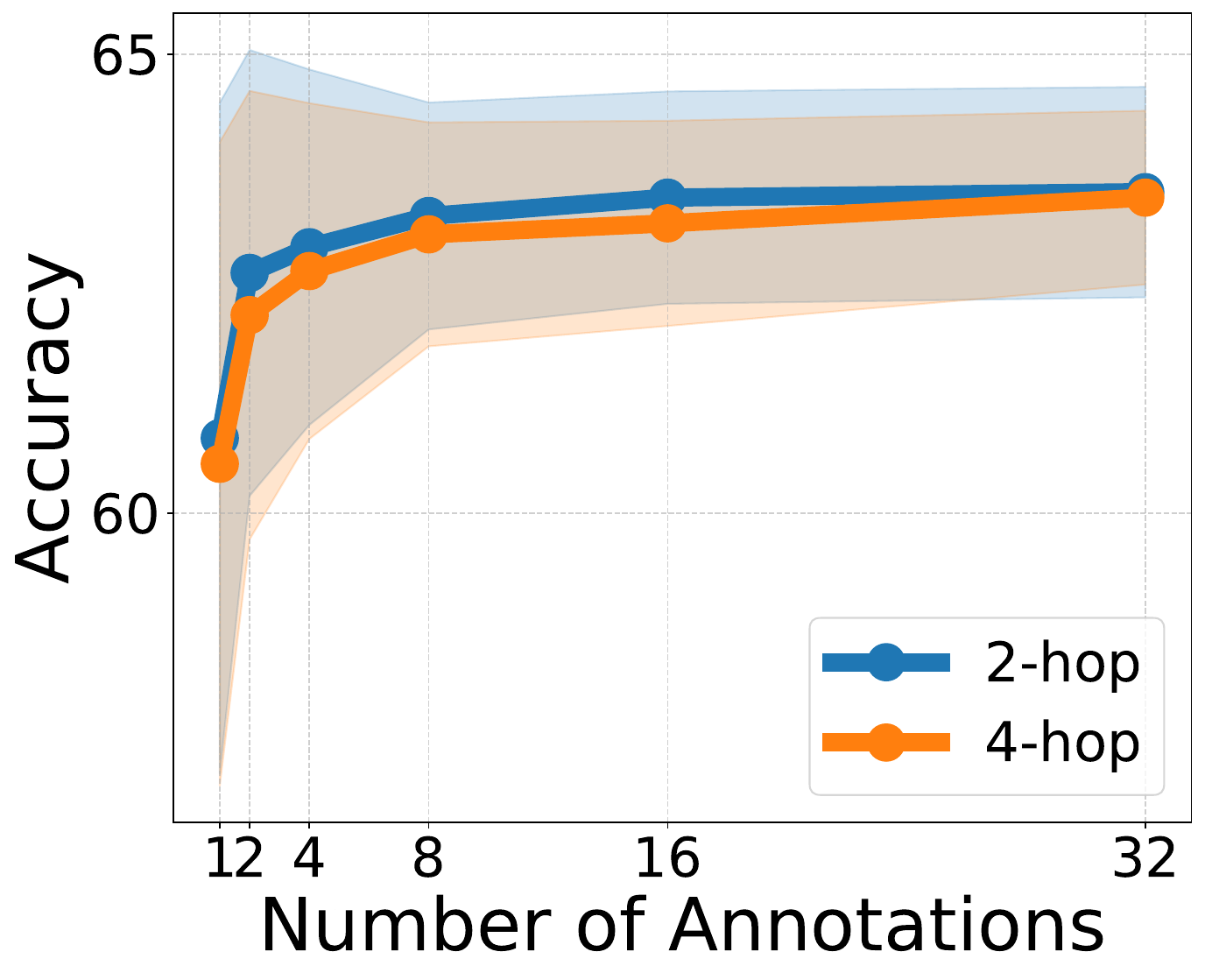}
        \caption{UF-P-2 (AVG)}
    \end{subfigure}
    \hfill
    \begin{subfigure}[b]{0.48\columnwidth}
        \centering
        \includegraphics[width=\textwidth]{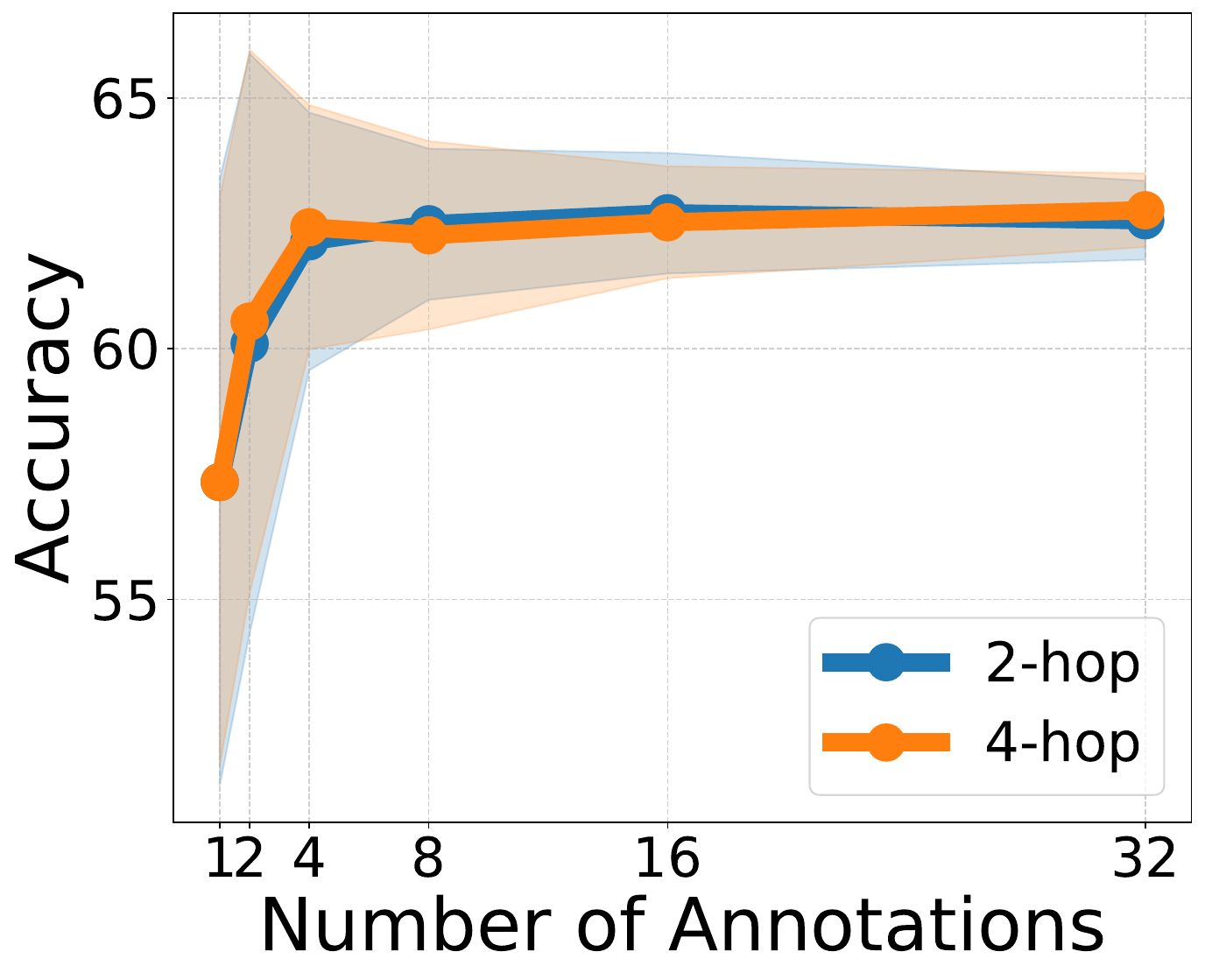}
        \caption{UF-P-4 (AVG)}
    \end{subfigure}
    \caption{Accuracy of unseen user adaptation as the number of provided annotation sets increases, evaluated on UF-P-2/4 (AVG) with \texttt{gemma-2b-it}. \emph{2-hop} and \emph{4-hop} indicates 2-hop and 4-hop adaptation, respectively. } 
    \label{fig:acc_w_num_context}

\end{figure}

\paragraph{Effect of the number of annotations in unseen user adaptation.} \cref{fig:acc_w_num_context} shows accuracy as the number of provided annotations increases in UF-P-2 (AVG) and UF-P-4 (AVG). We observe that additional annotations lead to more accurate preference predictions for unseen users in general. However, in practice, even eight annotations are sufficient, enabling accurate inference of each user's preference. We also compare two-hop and four-hop adaptations, but there is no significant difference.

\begin{figure}[t]
    \centering
    \includegraphics[width=\columnwidth]{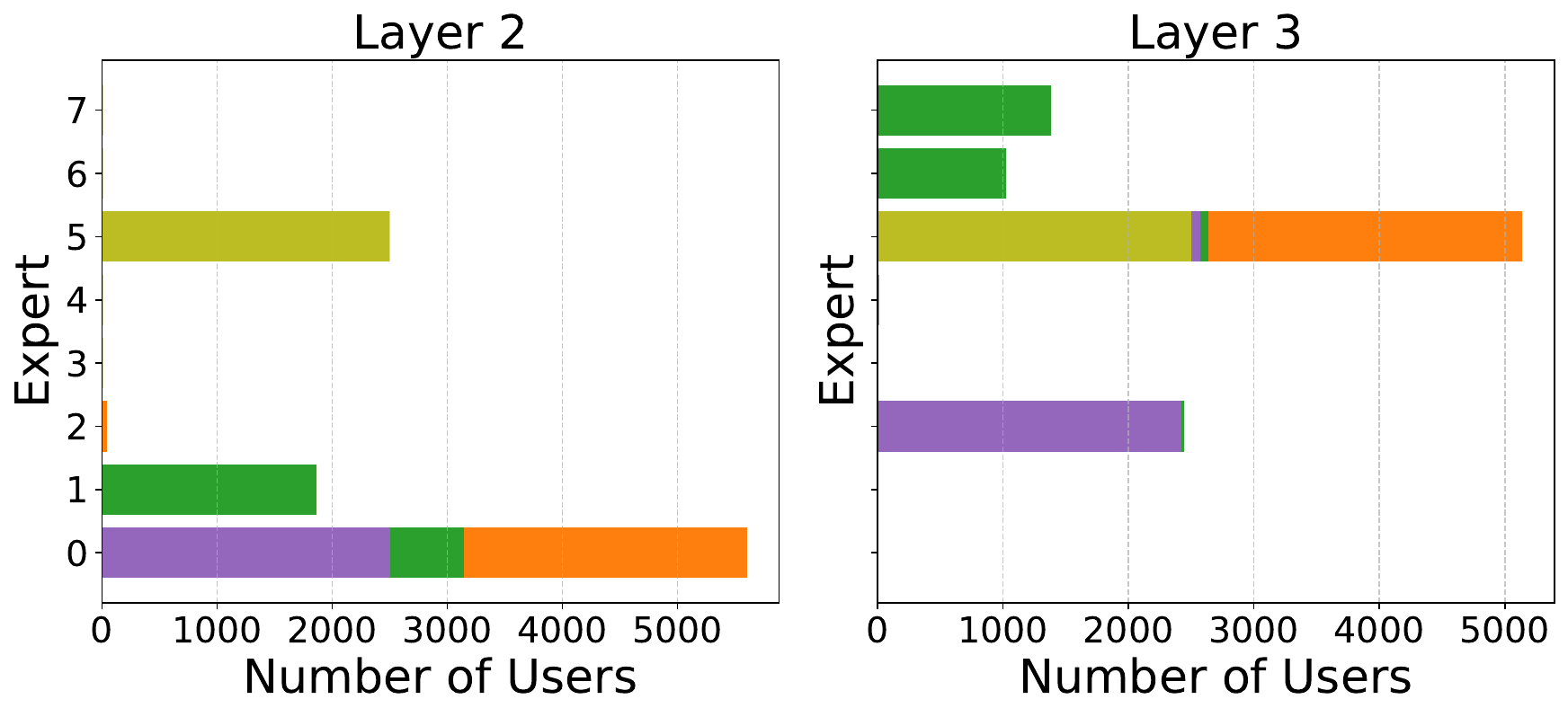} 
    \caption{Expert allocation at layers 2 and 3 in UF-P-4 (ALL) with \texttt{gemma-2b-it}. Colors indicate preference groups. Users with similar preference groups are mapped to the same expert.} 
    \label{fig:mole-vis} 
\end{figure}

\begin{table}[t]
\centering
\resizebox{\columnwidth}{!}{
\begin{tabular}{lcc}
\toprule
 &     UF-P-2 (ALL)      &   UF-P-4 (ALL)     \\ \midrule
\ours{} & \multicolumn{1}{r}{$\textbf{63.81}_{\pm0.16}$} & \multicolumn{1}{r}{$\textbf{62.57}_{\pm0.38}$} \\ \midrule
w/o GNN embedding & \multicolumn{1}{r}{$62.09_{\pm0.38}$} & \multicolumn{1}{r}{$56.75_{\pm0.30}$} \\
w/o MoLE $(n=64)$ & \multicolumn{1}{r}{$62.69_{\pm0.86}$} & \multicolumn{1}{r}{$62.28_{\pm0.33}$} \\ 
w/o MoLE $(n=16)$ & \multicolumn{1}{r}{$62.43_{\pm0.69}$} & \multicolumn{1}{r}{$62.13_{\pm0.12}$} \\

\bottomrule
\end{tabular}
}
\caption{Ablation study of \ours{} in UF-P-2/4 (ALL) with \texttt{gemma-2b-it}. \emph{w/o GNN embedding} replaces user embeddings from GNN with learnable user embeddings.
\emph{w/o MoLE} removes the MoLE and projects user embeddings into the token space. The symbol \(n\) denotes the LoRA rank.}
\label{tab:ablation.mole}
\end{table}

\begin{table}[t]
\centering
\resizebox{0.8\linewidth}{!}{%
    \begin{tabular}{lcc}
    \toprule
     &     UF-P-4 (ALL)      &   UF-P-4 (AVG)     \\ \midrule
    \ours{} & \multicolumn{1}{r}{$\textbf{61.62}_{\pm0.10}$} & \multicolumn{1}{r}{$\textbf{61.97}_{\pm0.35}$} \\ \midrule
    Naive Avg. & \multicolumn{1}{r}{$59.91_{\pm0.59}$} & \multicolumn{1}{r}{$59.39_{\pm0.50}$} \\ 
    User Opt. & \multicolumn{1}{r}{$59.24_{\pm0.71}$} & \multicolumn{1}{r}{$59.45_{\pm0.72}$} \\ 
    \bottomrule
    \end{tabular}
}
\caption{Accuracy of unseen-user adaptation in UF-P-4 (ALL/AVG) with \texttt{gemma-2b-it}. 
\emph{Naive Avg.} computes the unseen user's embedding as the unweighted average of 2-hop neighbors, 
while \ours{} applies a weighted average. \emph{User Opt.} represents an optimization-based approach that learns a parameterized user embedding by maximizing the likelihood of the given annotations.}
\label{tab:ablation.unseen_method}
\end{table}

\paragraph{Ablation study of \ours{}.}
\cref{tab:ablation.mole} presents an ablation study of \ours{}, focusing on GNN-derived user embeddings and the MoLE architecture. When GNN embeddings are removed, user representations become learnable parameters. Without MoLE, user embeddings are projected into the token space and passed as an additional token to the reward model. The results indicate that components of \ours{} are effective. Specifically, GNN-based embeddings are a crucial component of \ours{}, and the MoLE architecture further enhances accuracy. Notably, \ours{} uses fewer activated parameters than w/o MoLE \((n=64)\).

\cref{fig:mole-vis} depicts expert allocation across layers two and three, where the user-conditioned gating mechanism partitions users differently at each layer. We can observe that users with the same preferences tend to be routed to the same expert.

We provide the ablation study of the number of experts in \cref{appendix.experimental_results}.

\paragraph{Ablation study of unseen user adaptation.}
We conduct an ablation study to evaluate the effectiveness of the unseen user adaptation strategy, comparing it to two baselines, Naive Avg and User Opt.
Naive Avg assigns each unseen user embedding as the unweighted average of 2-hop seen user embeddings. 
User Opt replaces $\embedding{e}_u^{(L)}$ with a parameterized embedding learned by minimizing Equation~(\ref{eqn:gcf}) on the provided annotations. \cref{tab:ablation.unseen_method} reports results in UF-P-4-ALL/AVG with \texttt{gemma-2b-it}, showing that \ours{} outperforms both alternatives while achieving better computational efficiency than the optimization-based User Opt.

\cref{fig:comparison} illustrates that naive averaging places unseen users away from identical preference group users, whereas our method clusters them more closely with users who share the same preferences.

\begin{figure}[t]
    \centering
    \begin{subfigure}[b]{0.49\columnwidth}
        \centering
        \includegraphics[width=\textwidth]{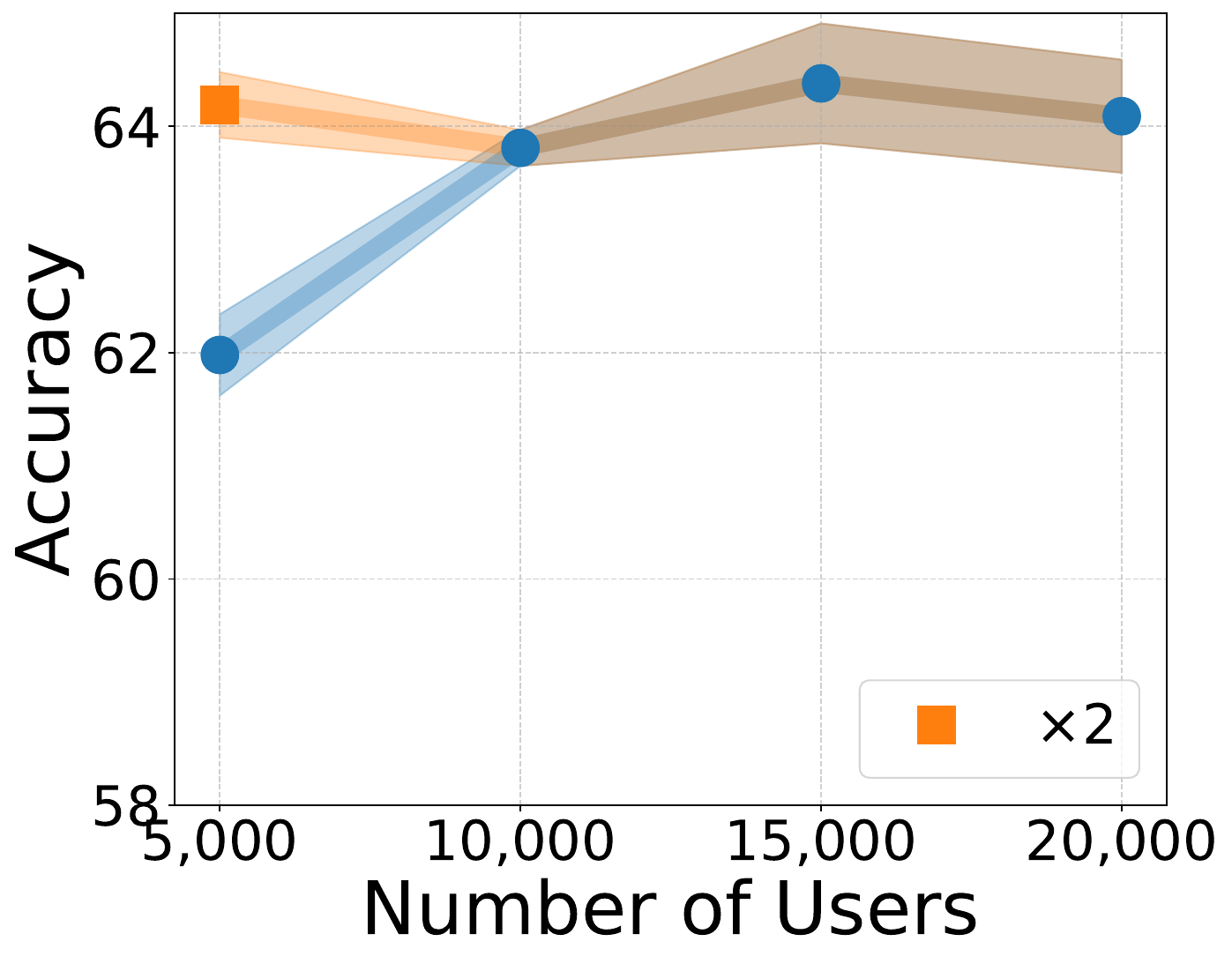}
        \caption{UF-P-2 (ALL)}
    \end{subfigure}
    \hfill
    \begin{subfigure}[b]{0.49\columnwidth}
        \centering
        \includegraphics[width=\textwidth]{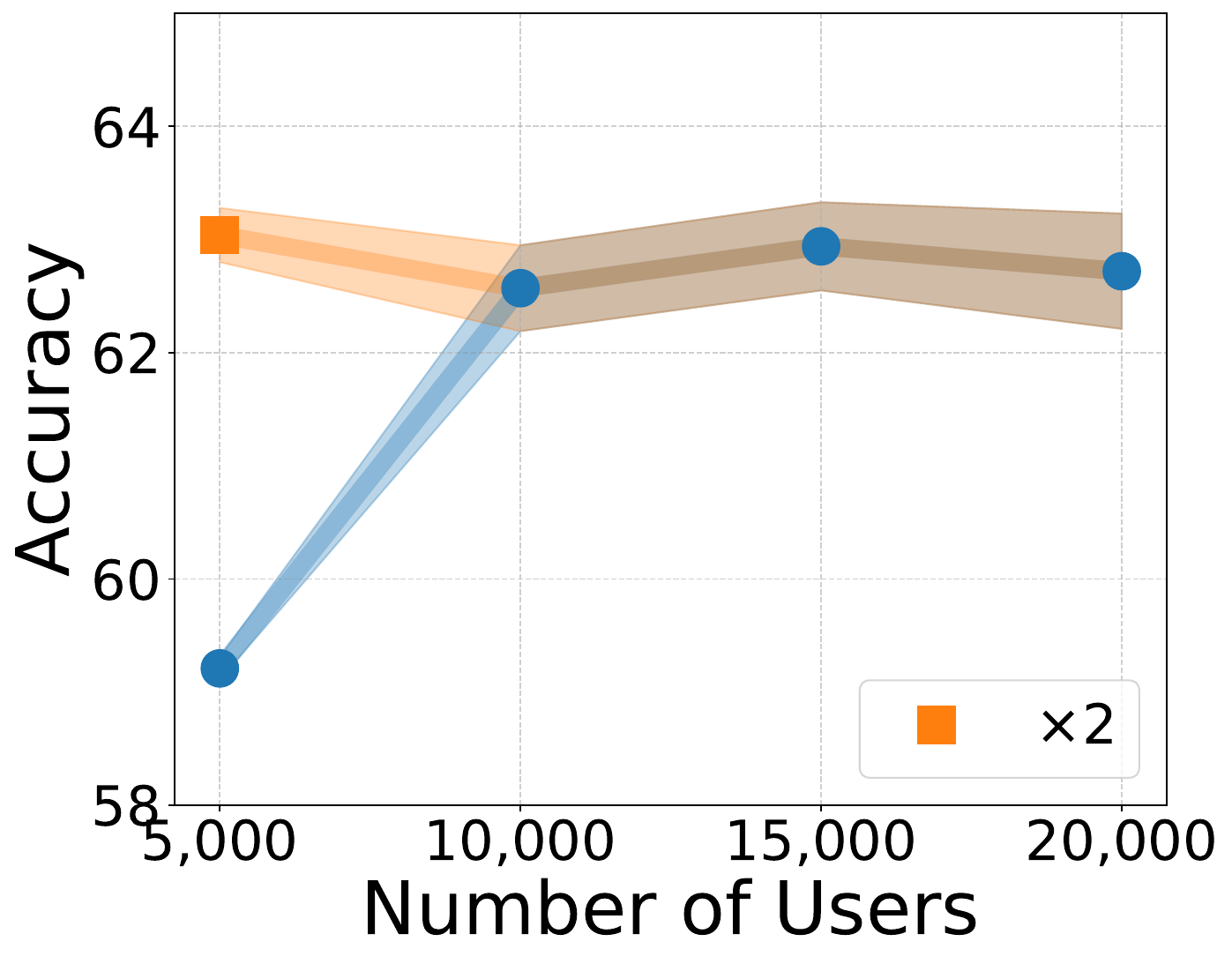}
        \caption{UF-P-4 (ALL)}
    \end{subfigure}
\caption{Accuracy of reward models on UF-P-2 and UF-P-4 (ALL) with \texttt{gemma-2b-it} with varying number of seen users. The number of annotations per user remains constant except in the case with ``$\times2$,'' where we double the per-user annotations only for $5,000$ users, making the total number of annotations $10,000$.}
\label{fig:ablation.num_users}
\end{figure}

\paragraph{Ablation study of the number of users.} We conduct an ablation study of CoPL by varying the number of users and report the performance in \cref{fig:ablation.num_users}. 
The performance of the model is consistent except for the case where there are only $5,000$ users in the training set. The performance with 5,000 users becomes comparable when we double the number of annotations $(2\times)$, indicating the need for a sufficient amount of annotations to capture diverse preferences.

\paragraph{Training reward models with GNN.}
\cref{tab:GCN-ACC} reports GNN accuracy on seen users and responses for test pairs excluded from the training dataset. The results demonstrate that GNN can accurately predict labels for unannotated pairs with sparse annotations. We provide the additional ablation study of message-passing in \cref{appendix.experimental_results}. 

\cref{tab:analysis.GCN-aug} examines the impact of training with GNN-based pseudo labels, allowing the model to leverage additional preference data. Although the pseudo-labeled pairs increase the dataset size, performance is slightly worse than using only user-provided annotations, suggesting that noise degrades model accuracy.  

To investigate the effect of noise further, a user-specific reward model is trained on pseudo labels for a random sample of 10 users per group. The results are considerably worse than the G-Oracle, indicating that noisy labels introduce training instability. This observation aligns with \citet{wang2024reward}, which notes that noisy preference labels can lead to training instability and performance degradation. 

\begin{table}[t]
\resizebox{\columnwidth}{!}{
\begin{tabular}{cccc}
\toprule
 \multicolumn{2}{c}{UF-P-2}                        & \multicolumn{2}{c}{UF-P-4}                        \\ \midrule
     ALL      &   AVG     & ALL    &   AVG      \\ \midrule
 \multicolumn{1}{r}{$84.84_{\pm0.83}$} & \multicolumn{1}{r}{$84.32_{\pm0.09}$} & \multicolumn{1}{r}{$90.01_{\pm0.35}$} & \multicolumn{1}{r}{$87.74_{\pm0.19}$} \\ \bottomrule
\end{tabular}
}
\caption{Test accuracy of the GNN. We evaluate the model using the same users from training but with annotation pairs that are not reflected in the graph. }
\label{tab:GCN-ACC}
\end{table}
\begin{table}[t]
\centering
\resizebox{0.8\linewidth}{!}{%
    \begin{tabular}{lcc}
    \toprule
     &     UF-P-2 (ALL)      &   UF-P-4 (ALL)     \\ \midrule
    \ours{} & \multicolumn{1}{r}{$63.81_{\pm0.16}$} & \multicolumn{1}{r}{$62.57_{\pm0.38}$} \\ 
    Pseudo label & \multicolumn{1}{r}{$62.77_{\pm0.70}$} & \multicolumn{1}{r}{$62.26_{\pm0.27}$} \\ \midrule
    G-Oracle & \multicolumn{1}{r}{$64.53_{\pm0.14}$} & \multicolumn{1}{r}{$61.52_{\pm0.13}$} \\ 
    User-specific & \multicolumn{1}{r}{$58.09_{\pm1.73}$} & \multicolumn{1}{r}{$55.30_{\pm3.30}$} \\
    \bottomrule
    \end{tabular}
}
\caption{Accuracy of reward model trained by using a pre-trained GNN in UF-P-2/4 (ALL) with \texttt{gemma-2b-it}. The \emph{"pseudo-label"} trains a reward model on all seen user–response pairs, with annotations provided by GNN-predicted labels.  The \emph{"user-specific"} refers to a BTL model trained with pseudo-labels for each user. Only 10 users per group are sampled due to computational cost.}
\label{tab:analysis.GCN-aug}
\end{table}

\section{Conclusion}


In this work, we introduced CoPL, a novel approach for personalizing LLMs through graph-based collaborative filtering and MoLE. Unlike existing methods that treat user preferences independently or require predefined clusters, our approach leverages multi-hop user-response relationships to improve preference estimation, even in sparse annotation settings. By integrating user-specific embeddings into the reward modeling process with MoLE, CoPL effectively predicts an individual preference.

\section*{Limitations}
This work demonstrates how GCF-based user embeddings enable personalization in sparse settings, but we do not extensively explore other GNN architectures that could further reduce sample complexity. Additionally, although \ours{} employs a gating mechanism for user-specific expert allocation, we did not apply load-balancing loss, which induces more even activation among experts. As a result, some experts remain inactive in \cref{fig:mole-vis}. Future work may investigate different GNN designs and incorporate load-balancing techniques to fully leverage the potential of GNN and MoLE, respectively.

The group-wise oracle model may appear underwhelming, likely because our smaller backbone LLM struggles to capture subtle stylistic differences between responses. Larger-scale models (over 30B parameters) could better handle these nuances; however, constraints in our current setup prevent such experiments, and we defer them to future work.

{Although CoPL is robust in sparse regimes compared to prior methods, it still depends on having sufficient annotation overlap to train the graph-based collaborative filtering model. In cases where the overlap is exceedingly limited, this reliance may constrain the model's flexibility. While existing preference datasets often contain such overlap \citep{wang2024helpsteer,stiennon2020learning,zhang2024diverging,bai2022training}, relaxing this requirement is an important next step. A promising approach is to construct user–response graphs from semantic similarity computed with sentence embeddings or other textual similarity measures, which would extend CoPL to settings without explicit overlap.}

{The effectiveness of our adaptation procedure depends on the informativeness of a new user’s annotations. When annotated pairs from a user mostly involve common pairs, they contain little information about that user’s preferences, thereby degrading adaptation performance. Integrating active learning to select informative pairs for annotation could mitigate this issue and reduce sample complexity.}

{Finally, \cref{fig:ablation.group-ratio} and \cref{tab:groupwise_accuracy} show that  CoPL can favor majority groups under severe imbalance, even though it captures diverse preferences overall. Exploring loss reweighting is a promising direction. Methods such as focal loss \citep{lin2017focal, subramanian2021fairness}, which increase the weight on high-error or underrepresented examples, may reduce majority bias and improve robustness.}

\section*{Acknowledgements}
{This work was supported by the National Research Foundation of Korea (NRF) grant funded by the Korea government(MSIT) (RS-2024-00337955; RS-2023-00217286) and Institute of Information \& communications Technology Planning \& Evaluation (IITP) grant funded by the Korea government(MSIT) (RS-2024-00457882, National AI Research Lab Project; RS-2019-II191906, Artificial Intelligence Graduate School Program(POSTECH)).}

\bibliography{custom}

\begin{thebibliography}{46}
\providecommand{\natexlab}[1]{#1}

\bibitem[{Achiam et~al.(2023)Achiam, Adler, Agarwal, Ahmad, Akkaya, Aleman, Almeida, Altenschmidt, Altman, Anadkat et~al.}]{achiam2023gpt}
Josh Achiam, Steven Adler, Sandhini Agarwal, Lama Ahmad, Ilge Akkaya, Florencia~Leoni Aleman, Diogo Almeida, Janko Altenschmidt, Sam Altman, Shyamal Anadkat, et~al. 2023.
\newblock Gpt-4 technical report.
\newblock \emph{arXiv preprint arXiv:2303.08774}.

\bibitem[{Bai et~al.(2022)Bai, Jones, Ndousse, Askell, Chen, DasSarma, Drain, Fort, Ganguli, Henighan et~al.}]{bai2022training}
Yuntao Bai, Andy Jones, Kamal Ndousse, Amanda Askell, Anna Chen, Nova DasSarma, Dawn Drain, Stanislav Fort, Deep Ganguli, Tom Henighan, et~al. 2022.
\newblock Training a helpful and harmless assistant with reinforcement learning from human feedback.
\newblock \emph{arXiv preprint arXiv:2204.05862}.

\bibitem[{Barreto et~al.(2025)Barreto, Dumoulin, Mao, Perez-Nieves, Shahriari, Dauphin, Precup, and Larochelle}]{barreto2025capturing}
Andr{\'e} Barreto, Vincent Dumoulin, Yiran Mao, Nicolas Perez-Nieves, Bobak Shahriari, Yann Dauphin, Doina Precup, and Hugo Larochelle. 2025.
\newblock Capturing individual human preferences with reward features.
\newblock \emph{arXiv preprint arXiv:2503.17338}.

\bibitem[{Bradley and Terry(1952)}]{bradley1952rank}
Ralph~Allan Bradley and Milton~E Terry. 1952.
\newblock Rank analysis of incomplete block designs: I. the method of paired comparisons.
\newblock \emph{Biometrika}, 39(3/4):324--345.

\bibitem[{Chen et~al.(2024{\natexlab{a}})Chen, Chen, Rege, and Vinayak}]{chen2024palpluralisticalignmentframework}
Daiwei Chen, Yi~Chen, Aniket Rege, and Ramya~Korlakai Vinayak. 2024{\natexlab{a}}.
\newblock \href {https://arxiv.org/abs/2406.08469} {Pal: Pluralistic alignment framework for learning from heterogeneous preferences}.
\newblock \emph{Preprint}, arXiv:2406.08469.

\bibitem[{Chen et~al.(2020)Chen, Wu, Hong, Zhang, and Wang}]{chen2020revisiting}
Lei Chen, Le~Wu, Richang Hong, Kun Zhang, and Meng Wang. 2020.
\newblock Revisiting graph based collaborative filtering: A linear residual graph convolutional network approach.
\newblock In \emph{Proceedings of the AAAI conference on artificial intelligence}.

\bibitem[{Chen et~al.(2024{\natexlab{b}})Chen, Zheng, Wang, Jin, Huang, Ye, Zhang, Zhou, Xi, Gui et~al.}]{chen2024improving}
Lu~Chen, Rui Zheng, Binghai Wang, Senjie Jin, Caishuang Huang, Junjie Ye, Zhihao Zhang, Yuhao Zhou, Zhiheng Xi, Tao Gui, et~al. 2024{\natexlab{b}}.
\newblock Improving discriminative capability of reward models in rlhf using contrastive learning.
\newblock In \emph{Proceedings of the 2024 Conference on Empirical Methods in Natural Language Processing}, pages 15270--15283.

\bibitem[{Chen et~al.(2024{\natexlab{c}})Chen, Jie, and Ma}]{chen2024llava}
Shaoxiang Chen, Zequn Jie, and Lin Ma. 2024{\natexlab{c}}.
\newblock Llava-mole: Sparse mixture of lora experts for mitigating data conflicts in instruction finetuning mllms.
\newblock \emph{arXiv preprint arXiv:2401.16160}.

\bibitem[{Chen et~al.(2023)Chen, Wang, Wang, Liu, Yin, Liu, Sheng, Ouyang, and Shao}]{chenoctavius}
Zeren Chen, Ziqin Wang, Zhen Wang, Huayang Liu, Zhenfei Yin, Si~Liu, Lu~Sheng, Wanli Ouyang, and Jing Shao. 2023.
\newblock Octavius: Mitigating task interference in mllms via lora-moe.
\newblock In \emph{ICLR}.

\bibitem[{Cui et~al.(2023)Cui, Yuan, Ding, Yao, Zhu, Ni, Xie, Liu, and Sun}]{cui2023ultrafeedback}
Ganqu Cui, Lifan Yuan, Ning Ding, Guanming Yao, Wei Zhu, Yuan Ni, Guotong Xie, Zhiyuan Liu, and Maosong Sun. 2023.
\newblock \href {https://arxiv.org/abs/2310.01377} {Ultrafeedback: Boosting language models with high-quality feedback}.
\newblock \emph{Preprint}, arXiv:2310.01377.

\bibitem[{Dai et~al.(2023)Dai, Pan, Sun, Ji, Xu, Liu, Wang, and Yang}]{dai2023safe}
Josef Dai, Xuehai Pan, Ruiyang Sun, Jiaming Ji, Xinbo Xu, Mickel Liu, Yizhou Wang, and Yaodong Yang. 2023.
\newblock Safe rlhf: Safe reinforcement learning from human feedback.
\newblock \emph{arXiv preprint arXiv:2310.12773}.

\bibitem[{Guan et~al.(2025)Guan, Wu, Li, Cheng, and Wu}]{guan2025survey}
Jian Guan, Junfei Wu, Jia-Nan Li, Chuanqi Cheng, and Wei Wu. 2025.
\newblock A survey on personalized alignment--the missing piece for large language models in real-world applications.
\newblock \emph{arXiv preprint arXiv:2503.17003}.

\bibitem[{He and Chua(2017)}]{he2017neuralfactorizationmachinessparse}
Xiangnan He and Tat-Seng Chua. 2017.
\newblock \href {https://arxiv.org/abs/1708.05027} {Neural factorization machines for sparse predictive analytics}.
\newblock \emph{Preprint}, arXiv:1708.05027.

\bibitem[{He et~al.(2020)He, Deng, Wang, Li, Zhang, and Wang}]{he2020lightgcn}
Xiangnan He, Kuan Deng, Xiang Wang, Yan Li, Yongdong Zhang, and Meng Wang. 2020.
\newblock Lightgcn: Simplifying and powering graph convolution network for recommendation.
\newblock In \emph{Proceedings of the 43rd International ACM SIGIR conference on research and development in Information Retrieval}, pages 639--648.

\bibitem[{Hu et~al.(2021)Hu, Shen, Wallis, Allen-Zhu, Li, Wang, Wang, and Chen}]{hu2021lora}
Edward~J Hu, Yelong Shen, Phillip Wallis, Zeyuan Allen-Zhu, Yuanzhi Li, Shean Wang, Lu~Wang, and Weizhu Chen. 2021.
\newblock Lora: Low-rank adaptation of large language models.
\newblock \emph{arXiv preprint arXiv:2106.09685}.

\bibitem[{Jang et~al.(2023)Jang, Kim, Lin, Wang, Hessel, Zettlemoyer, Hajishirzi, Choi, and Ammanabrolu}]{jang2023personalized}
Joel Jang, Seungone Kim, Bill~Yuchen Lin, Yizhong Wang, Jack Hessel, Luke Zettlemoyer, Hannaneh Hajishirzi, Yejin Choi, and Prithviraj Ammanabrolu. 2023.
\newblock Personalized soups: Personalized large language model alignment via post-hoc parameter merging.
\newblock \emph{arXiv preprint arXiv:2310.11564}.

\bibitem[{Ju et~al.(2024)Ju, Shiao, Guo, Ye, Liu, Shah, and Zhao}]{ju2024does}
Mingxuan Ju, William Shiao, Zhichun Guo, Yanfang Ye, Yozen Liu, Neil Shah, and Tong Zhao. 2024.
\newblock How does message passing improve collaborative filtering?
\newblock \emph{arXiv preprint arXiv:2404.08660}.

\bibitem[{Lambert et~al.(2024)Lambert, Pyatkin, Morrison, Miranda, Lin, Chandu, Dziri, Kumar, Zick, Choi et~al.}]{lambert2024rewardbench}
Nathan Lambert, Valentina Pyatkin, Jacob Morrison, LJ~Miranda, Bill~Yuchen Lin, Khyathi Chandu, Nouha Dziri, Sachin Kumar, Tom Zick, Yejin Choi, et~al. 2024.
\newblock Rewardbench: Evaluating reward models for language modeling.
\newblock \emph{arXiv preprint arXiv:2403.13787}.

\bibitem[{Li et~al.(2025)Li, Guan, Wu, Wu, and Yan}]{li20251}
Jia-Nan Li, Jian Guan, Songhao Wu, Wei Wu, and Rui Yan. 2025.
\newblock From 1,000,000 users to every user: Scaling up personalized preference for user-level alignment.
\newblock \emph{arXiv preprint arXiv:2503.15463}.

\bibitem[{Li et~al.(2022)Li, Zhao, Li, Chan, Faloutsos, Karypis, Pantel, and McAuley}]{li2022coarse}
Jiacheng Li, Tong Zhao, Jin Li, Jim Chan, Christos Faloutsos, George Karypis, Soo-Min Pantel, and Julian McAuley. 2022.
\newblock Coarse-to-fine sparse sequential recommendation.
\newblock In \emph{Proceedings of the 45th international ACM SIGIR conference on research and development in information retrieval}, pages 2082--2086.

\bibitem[{Li et~al.(2024)Li, Lipton, and Leqi}]{li2024personalized}
Xinyu Li, Zachary~C Lipton, and Liu Leqi. 2024.
\newblock Personalized language modeling from personalized human feedback.
\newblock \emph{arXiv preprint arXiv:2402.05133}.

\bibitem[{Lin et~al.(2017)Lin, Goyal, Girshick, He, and Doll{\'a}r}]{lin2017focal}
Tsung-Yi Lin, Priya Goyal, Ross Girshick, Kaiming He, and Piotr Doll{\'a}r. 2017.
\newblock Focal loss for dense object detection.
\newblock In \emph{Proceedings of the IEEE international conference on computer vision}, pages 2980--2988.

\bibitem[{Lin et~al.(2022)Lin, Tian, Hou, and Zhao}]{10.1145/3485447.3512104}
Zihan Lin, Changxin Tian, Yupeng Hou, and Wayne~Xin Zhao. 2022.
\newblock \href {https://doi.org/10.1145/3485447.3512104} {Improving graph collaborative filtering with neighborhood-enriched contrastive learning}.
\newblock In \emph{Proceedings of the ACM Web Conference 2022}, WWW '22, page 2320–2329, New York, NY, USA. Association for Computing Machinery.

\bibitem[{Liu et~al.(2025)Liu, Qiu, Li, Dai, Zhu, Hu, Yang, and King}]{liu2025survey}
Jiahong Liu, Zexuan Qiu, Zhongyang Li, Quanyu Dai, Jieming Zhu, Minda Hu, Menglin Yang, and Irwin King. 2025.
\newblock A survey of personalized large language models: Progress and future directions.
\newblock \emph{arXiv preprint arXiv:2502.11528}.

\bibitem[{Liu et~al.(2024)Liu, Wu, Zhao, Zhu, Xu, Tian, and Zheng}]{10.1145/3626772.3657722}
Qidong Liu, Xian Wu, Xiangyu Zhao, Yuanshao Zhu, Derong Xu, Feng Tian, and Yefeng Zheng. 2024.
\newblock \href {https://doi.org/10.1145/3626772.3657722} {When moe meets llms: Parameter efficient fine-tuning for multi-task medical applications}.
\newblock In \emph{Proceedings of the 47th International ACM SIGIR Conference on Research and Development in Information Retrieval}, SIGIR '24. Association for Computing Machinery.

\bibitem[{Loshchilov(2017)}]{loshchilov2017decoupled}
I~Loshchilov. 2017.
\newblock Decoupled weight decay regularization.
\newblock \emph{arXiv preprint arXiv:1711.05101}.

\bibitem[{Molina et~al.(2024)Molina, Montalvo, Ochoa, Denny, and Porter}]{molina2024leveraging}
Ismael~Villegas Molina, Audria Montalvo, Benjamin Ochoa, Paul Denny, and Leo Porter. 2024.
\newblock Leveraging llm tutoring systems for non-native english speakers in introductory cs courses.
\newblock \emph{arXiv preprint arXiv:2411.02725}.

\bibitem[{Oh et~al.(2024)Oh, Lee, and Ok}]{oh2024activepreferencebasedlearningmultidimensional}
Minhyeon Oh, Seungjoon Lee, and Jungseul Ok. 2024.
\newblock \href {https://arxiv.org/abs/2411.00524} {Active preference-based learning for multi-dimensional personalization}.
\newblock \emph{Preprint}, arXiv:2411.00524.

\bibitem[{Poddar et~al.(2024)Poddar, Wan, Ivison, Gupta, and Jaques}]{poddar2024personalizing}
Sriyash Poddar, Yanming Wan, Hamish Ivison, Abhishek Gupta, and Natasha Jaques. 2024.
\newblock Personalizing reinforcement learning from human feedback with variational preference learning.
\newblock \emph{arXiv preprint arXiv:2408.10075}.

\bibitem[{Rendle et~al.(2012)Rendle, Freudenthaler, Gantner, and Schmidt-Thieme}]{rendle2012bpr}
Steffen Rendle, Christoph Freudenthaler, Zeno Gantner, and Lars Schmidt-Thieme. 2012.
\newblock Bpr: Bayesian personalized ranking from implicit feedback.
\newblock \emph{arXiv preprint arXiv:1205.2618}.

\bibitem[{Shi et~al.(2024)Shi, Li, Ma, Yang, Ma, and Li}]{shi2024chops}
Jingzhe Shi, Jialuo Li, Qinwei Ma, Zaiwen Yang, Huan Ma, and Lei Li. 2024.
\newblock Chops: Chat with customer profile systems for customer service with llms.
\newblock \emph{arXiv preprint arXiv:2404.01343}.

\bibitem[{Siththaranjan et~al.(2024)Siththaranjan, Laidlaw, and Hadfield-Menell}]{siththaranjan2024DPL}
Anand Siththaranjan, Cassidy Laidlaw, and Dylan Hadfield-Menell. 2024.
\newblock \href {https://arxiv.org/abs/2312.08358} {Distributional preference learning: Understanding and accounting for hidden context in rlhf}.
\newblock In \emph{ICLR}.

\bibitem[{Sorensen et~al.(2024)Sorensen, Moore, Fisher, Gordon, Mireshghallah, Rytting, Ye, Jiang, Lu, Dziri et~al.}]{sorensen2024roadmap}
Taylor Sorensen, Jared Moore, Jillian Fisher, Mitchell Gordon, Niloofar Mireshghallah, Christopher~Michael Rytting, Andre Ye, Liwei Jiang, Ximing Lu, Nouha Dziri, et~al. 2024.
\newblock A roadmap to pluralistic alignment.
\newblock \emph{arXiv preprint arXiv:2402.05070}.

\bibitem[{Stiennon et~al.(2020)Stiennon, Ouyang, Wu, Ziegler, Lowe, Voss, Radford, Amodei, and Christiano}]{stiennon2020learning}
Nisan Stiennon, Long Ouyang, Jeffrey Wu, Daniel Ziegler, Ryan Lowe, Chelsea Voss, Alec Radford, Dario Amodei, and Paul~F Christiano. 2020.
\newblock Learning to summarize with human feedback.
\newblock \emph{Advances in neural information processing systems}, 33:3008--3021.

\bibitem[{Subramanian et~al.(2021)Subramanian, Rahimi, Baldwin, Cohn, and Frermann}]{subramanian2021fairness}
Shivashankar Subramanian, Afshin Rahimi, Timothy Baldwin, Trevor Cohn, and Lea Frermann. 2021.
\newblock Fairness-aware class imbalanced learning.
\newblock \emph{arXiv preprint arXiv:2109.10444}.

\bibitem[{Team et~al.(2024{\natexlab{a}})Team, Mesnard, Hardin, Dadashi, Bhupatiraju, Pathak, Sifre, Rivi{\`e}re, Kale, Love et~al.}]{team2024gemma}
Gemma Team, Thomas Mesnard, Cassidy Hardin, Robert Dadashi, Surya Bhupatiraju, Shreya Pathak, Laurent Sifre, Morgane Rivi{\`e}re, Mihir~Sanjay Kale, Juliette Love, et~al. 2024{\natexlab{a}}.
\newblock Gemma: Open models based on gemini research and technology.
\newblock \emph{arXiv preprint arXiv:2403.08295}.

\bibitem[{Team et~al.(2024{\natexlab{b}})Team, Riviere, Pathak, Sessa, Hardin, Bhupatiraju, Hussenot, Mesnard, Shahriari, Ram{\'e} et~al.}]{team2024gemma2}
Gemma Team, Morgane Riviere, Shreya Pathak, Pier~Giuseppe Sessa, Cassidy Hardin, Surya Bhupatiraju, L{\'e}onard Hussenot, Thomas Mesnard, Bobak Shahriari, Alexandre Ram{\'e}, et~al. 2024{\natexlab{b}}.
\newblock Gemma 2: Improving open language models at a practical size.
\newblock \emph{arXiv preprint arXiv:2408.00118}.

\bibitem[{Venkatraman et~al.(2024)Venkatraman, Tripto, and Lee}]{venkatraman2024collabstory}
Saranya Venkatraman, Nafis~Irtiza Tripto, and Dongwon Lee. 2024.
\newblock Collabstory: Multi-llm collaborative story generation and authorship analysis.
\newblock \emph{arXiv preprint arXiv:2406.12665}.

\bibitem[{Wang et~al.(2024{\natexlab{a}})Wang, Zheng, Chen, Xi, Shen, Zhou, Yan, Gui, Zhang, and Huang}]{wang2024reward}
Binghai Wang, Rui Zheng, Lu~Chen, Zhiheng Xi, Wei Shen, Yuhao Zhou, Dong Yan, Tao Gui, Qi~Zhang, and Xuan-Jing Huang. 2024{\natexlab{a}}.
\newblock Reward modeling requires automatic adjustment based on data quality.
\newblock In \emph{Findings of the Association for Computational Linguistics: EMNLP 2024}, pages 4041--4064.

\bibitem[{Wang et~al.(2019)Wang, He, Wang, Feng, and Chua}]{wang2019neural}
Xiang Wang, Xiangnan He, Meng Wang, Fuli Feng, and Tat-Seng Chua. 2019.
\newblock Neural graph collaborative filtering.
\newblock In \emph{Proceedings of the 42nd international ACM SIGIR conference on Research and development in Information Retrieval}, pages 165--174.

\bibitem[{Wang et~al.(2024{\natexlab{b}})Wang, Dong, Delalleau, Zeng, Shen, Egert, Zhang, Sreedhar, and Kuchaiev}]{wang2024helpsteer}
Zhilin Wang, Yi~Dong, Olivier Delalleau, Jiaqi Zeng, Gerald Shen, Daniel Egert, Jimmy Zhang, Makesh~Narsimhan Sreedhar, and Oleksii Kuchaiev. 2024{\natexlab{b}}.
\newblock Helpsteer 2: Open-source dataset for training top-performing reward models.
\newblock \emph{Advances in Neural Information Processing Systems}, 37:1474--1501.

\bibitem[{Wang et~al.(2023)Wang, Dong, Zeng, Adams, Sreedhar, Egert, Delalleau, Scowcroft, Kant, Swope et~al.}]{wang2023helpsteer}
Zhilin Wang, Yi~Dong, Jiaqi Zeng, Virginia Adams, Makesh~Narsimhan Sreedhar, Daniel Egert, Olivier Delalleau, Jane~Polak Scowcroft, Neel Kant, Aidan Swope, et~al. 2023.
\newblock Helpsteer: Multi-attribute helpfulness dataset for steerlm.
\newblock \emph{arXiv preprint arXiv:2311.09528}.

\bibitem[{Yang et~al.(2024{\natexlab{a}})Yang, Tao, Lyu, Ge, Chen, Shen, Zhu, and Li}]{Yang_2024_CVPR}
Kai Yang, Jian Tao, Jiafei Lyu, Chunjiang Ge, Jiaxin Chen, Weihan Shen, Xiaolong Zhu, and Xiu Li. 2024{\natexlab{a}}.
\newblock Using human feedback to fine-tune diffusion models without any reward model.
\newblock In \emph{Proceedings of the IEEE/CVF Conference on Computer Vision and Pattern Recognition (CVPR)}, pages 8941--8951.

\bibitem[{Yang et~al.(2024{\natexlab{b}})Yang, Pan, Luo, Qiu, Zhong, Yu, and Chen}]{yang2024rewards}
Rui Yang, Xiaoman Pan, Feng Luo, Shuang Qiu, Han Zhong, Dong Yu, and Jianshu Chen. 2024{\natexlab{b}}.
\newblock Rewards-in-context: Multi-objective alignment of foundation models with dynamic preference adjustment.
\newblock \emph{arXiv preprint arXiv:2402.10207}.

\bibitem[{Zhang et~al.(2024)Zhang, Wang, Hwang, Dong, Delalleau, Choi, Choi, Ren, and Pyatkin}]{zhang2024diverging}
Michael~JQ Zhang, Zhilin Wang, Jena~D Hwang, Yi~Dong, Olivier Delalleau, Yejin Choi, Eunsol Choi, Xiang Ren, and Valentina Pyatkin. 2024.
\newblock Diverging preferences: When do annotators disagree and do models know?
\newblock \emph{arXiv preprint arXiv:2410.14632}.

\bibitem[{Zollo et~al.(2024)Zollo, Siah, Ye, Li, and Namkoong}]{zollo2024personalllm}
Thomas~P Zollo, Andrew Wei~Tung Siah, Naimeng Ye, Ang Li, and Hongseok Namkoong. 2024.
\newblock Personalllm: Tailoring llms to individual preferences.
\newblock \emph{arXiv preprint arXiv:2409.20296}.

\end{thebibliography}

\clearpage
\appendix
\section*{Appendix}
\renewcommand{\thefigure}{A\arabic{figure}}
\setcounter{figure}{0}
\renewcommand{\thetable}{A\arabic{table}}
\setcounter{table}{0}

\section{Message Passing for Response Embeddings}
\label{appendix.message}

Given user and response embeddings at layer $\ell$, a message from neighborhood users to the response as
\begin{align}
&\embedding{m}_r^{+} = \sum_{u\in \mathcal{N}^+_r}
\alpha_{u, r}\Bigl( 
\hat{W}_1^{(\ell)} \embedding{e}_u^{(\ell)} +\hat{W}_2^{(\ell)} (\embedding{e}_u^{(\ell)} \odot \embedding{e}_r^{(\ell)})  \Bigr), \nonumber\\
&\embedding{m}_r^{-} = \sum_{u \in \mathcal{N}^-_r} \beta_{u,r}\Bigl( 
\hat{W}_3^{(\ell)} \embedding{e}_u^{(\ell)} +
\hat{W}_4^{(\ell)} (\embedding{e}_u^{(\ell)} \odot \embedding{e}_r^{(\ell)}) \Bigr), \nonumber\\
&\embedding{m}^{(\ell)}_r = \hat{W}_{\text{self}}^{(\ell)}\,\embedding{e}_r^{(\ell)} \;+\; \embedding{m}_r^{+} \;+\; \embedding{m}_r^{-},
\end{align}
where $ \hat{W}_1^{(\ell)}, \hat{W}_2^{(\ell)}, \hat{W}_3^{(\ell)}, \hat{W}_4^{(\ell)}, \hat{W}_{\text{self}}^{(\ell)}\in\mathbb{R}^{d\times d}$ are parameter matrices, $\odot$ is element-wise multiplication, and $\alpha_{u,r}$ and $\beta_{u,r}$ are normalization factors, set to $\frac{1}{\sqrt{|\mathcal{N}^+_u|\cdot|\mathcal{N}^+_r|}}$ and $\frac{1}{\sqrt{|\mathcal{N}^-_u|\cdot|\mathcal{N}^-_r|}}$, respectively. 

Then, the response embedding is updated with the aggregated message $\embedding{m}^{(\ell)}_r$:
\begin{align}
\embedding{e}_r^{(\ell+1)} = {\psi} \bigl(\embedding{m}^{(\ell)}_r\bigr),
\end{align}
where $\psi(\cdot)$ is a non-linear activation.

\section{Method Baselines}
\label{appendix.baseline}

\paragraph{Uniform.}
The uniform model is a standard approach for pairwise preference comparisons. We train the uniform model with all annotation pairs, which will capture the common preference.

\paragraph{Oracle.}
For an oracle model of our setting, we train the model with the true group membership of all users. A separate uniform model is trained for each group by aggregating annotations from the users in that group.

\paragraph{I2E \citep{li2024personalized}.}
I2E is a framework that uses DPO to personalize LLM. However, it can be easily extended to reward modeling. I2E trains a model that maps the user index into a learnable embedding. It appends each user embedding as an additional input token to the LLM, providing user-specific signals for reward prediction.

\paragraph{I2E$_\text{proxy}$ \citep{li2024personalized}.}
A variant of I2E that introduces $N$ proxy embeddings. A weighted combination of these proxies forms the final user embedding, which is passed to the LLM for reward prediction. In our experiments, we use $N=10$.

\paragraph{VPL \citep{poddar2024personalizing}.}
Variational Preference Learning (VPL) encodes user-specific annotations into user embeddings. The user embeddings are then combined with sentence representations via an MLP to predict reward scores. To capture the user preferences effectively, VPL uses a variational approach that maps the user annotations into a prior distribution.

\paragraph{PAL \citep{chen2024palpluralisticalignmentframework}.}
Pluralistic Alignment (PAL) applies an ideal-point model, where the distance between the user and the response determines the reward. The ideal point of the user is represented by $N$ proxies, set to $N=10$ in this work. Among variants of PAL, we use PAL-A with logistic loss.

\section{Related Works}
\label{appendix.related_works}

\paragraph{Personalized alignment.}

With the growth of generative models, alignment has emerged as a crucial strategy for mitigating undesirable outcomes, such as biased or harmful outputs, and ensuring that the model works with human preference~\citep{dai2023safe, Yang_2024_CVPR}. Alignment methods often rely on reward models. They typically build on the BTL framework, which relies on pairwise comparisons from various annotators. However, previous research has often focused on the average preference of annotators~\citep{achiam2023gpt}, ignoring the diverse preferences.

To address preference diversity, recent works~\citep{jang2023personalized, oh2024activepreferencebasedlearningmultidimensional, yang2024rewards} view this problem as a soft clustering problem, where user-specific preferences are treated as mixtures of predefined preference types. Although this approach effectively handles diverse preferences, it relies on specifying several preference types in advance.

Another line of work introduces user latent variable in the BTL framework~\citep{poddar2024personalizing, li2024personalized, chen2024palpluralisticalignmentframework}.
Although extending the BTL framework with latent user variables can address diverse preferences, the main challenge lies in obtaining user representations. One approach is to treat each user embedding as learnable parameters, ~\citep{li2024personalized, chen2024palpluralisticalignmentframework}, and the other strategy is to train an encoder that infers embeddings from the small set of annotated pairs provided by each user~\citep{poddar2024personalizing}.

\paragraph{Preference learning with sparse interactions.}

Preference learning with sparse interactions is a well-studied challenge in recommendation systems, where each user typically interacts with only a small fraction of the available items. Despite these limited interactions, the system should infer the preference of each user and recommend additional items accordingly~\citep{he2017neuralfactorizationmachinessparse, chen2020revisiting, li2022coarse, 10.1145/3485447.3512104}. Collaborative filtering (CF) is a widely adopted solution that assumes users with similar interaction histories will exhibit similar preferences.

Graph-based CF (GCF)~\citep{wang2019neural,he2020lightgcn} has been considered one of the most advanced algorithms for a recommendation system. GCF leverages graph neural networks (GNNs) to capture preference through the connectivity among users and items. Many GCFs are developed based on an implicit feedback assumption~\citep{rendle2012bpr}, where an edge between a user and an item reveals a preferable relation. Whereas in our setting, users provide explicit feedback given a pair of responses, making direct application of GCF unsuitable.



\section{Experimental Details}

In this section, we provide a detailed explanation of dataset construction and hyper-parameters.

\subsection{Datasets}
\label{appendix.dataset}

\paragraph{TL;DR.}
The TL;DR dataset~\citep{stiennon2020learning} contains Reddit posts alongside concise summaries and annotator IDs. 
Prior works~\citep{li2022coarse,chen2024palpluralisticalignmentframework} employ a modified version of this dataset by defining two simulated preference groups: one group favors shorter summaries, while the other prefers longer ones. The two groups provide different annotations for each summary pair. To focus on the most active annotators, they retain only the ten users with the highest number of annotations. We adopt the resulting set of annotation pairs from these ten users as our survey set.

\paragraph{Ultrafeedback-P.}

\citet{poddar2024personalizing} proposes the Ultrafeedback-P (UF-P) benchmark for personalized reward modeling, based on the Ultrafeedback (UF) dataset~\cite{cui2023ultrafeedback}, which provides response pairs rated on four attributes: helpfulness, honesty, instruction following, and truthfulness. In UF-P, each attribute corresponds to a distinct preference. For instance, a user belonging to the helpfulness group annotates pairs, solely considering the helpfulness score.

\textbf{UF-P-2} employs only two attributes and removes pairs that both user groups label identically, focusing on controversial cases where preferences differ. In \textbf{UF-P-4}, all four attributes are retained as preference dimensions, which allows for partial agreement among groups and hence increases complexity. Although \citet{poddar2024personalizing} also excludes pairs fully agreed upon by all users, the remaining set is larger and exhibits more variety than UF-P-2.

In \citet{poddar2024personalizing}, each user is given a small context sample from a limited set of unannotated pairs to infer the user’s preference. In contrast, we leverage every available pair in the dataset to infer each user’s preferences. For our dataset construction, we use UF-P-4 dataset.

\paragraph{PersonalLLM.}  
{PersonalLLM~\citep{zollo2024personalllm} is built with 10,402 open-ended prompts that were sampled from a larger pool of 37,919 conversational questions drawn from public RLHF and preference benchmarks such as Anthropic HH-RLHF~\citep{bai2022training}, NVIDIA HelpSteer~\citep{wang2023helpsteer}, and RewardBench~\citep{lambert2024rewardbench}. For each prompt, they used eight frontier chat models to generate a diverse response set that minimizes obvious quality gaps while covering latent preference dimensions. The resulting (prompt, response1, response2, ..., response8) tuples are split into 9,402 training and 1,000 test items.}

{Each response is evaluated by ten strong open-source reward models with heterogeneous alignment objectives. These reward models assign scalar scores capturing distinct value dimensions for every response. Storing the full 10×8 matrix of scores per prompt provides a dense, model-agnostic preference signal that later steps can recombine to reflect arbitrary preferences. To simulate a large user base, they treat the preference of a user as a weighted ensemble over the ten reward models. The weight is sampled from a Dirichlet distribution, where varying the concentration parameter controls preference diversity.}

{We use $\alpha=0.1$ for Dirichlet distribution. Due to computational constraints, we simplify the dataset by selecting three responses per prompt and considering only four reward dimensions. Following \citet{poddar2024personalizing}, we remove \emph{non-controversial} response pairs---those in which one response is strictly ranked below the other across all preference dimensions---to ensure the heterogeneity.}

\subsection{Hyper-parameters}
\label{appendix.training}

We describe the training details of GNN, a reward model, and unseen user adaptation, such as model architecture and hyper-parameters.

\paragraph{GNN.} The model consists of four message-passing layers, each with user and response embeddings of dimension 512. We use Leaky ReLU as a non-linear activation function to update user and response embeddings. Training proceeds for 300 epochs using the AdamW optimizer~\cite{loshchilov2017decoupled} with a learning rate of \(1\times10^{-4}\) and a cosine scheduler with warmup ratio \(0.1\). The batch size is 1024, and all experiments are conducted on an RTX 4090 GPU.

\paragraph{Reward models.}
\ours{} comprises an LLM backbone and a MoLE adapter. We use \texttt{gemma-2b-it} or \texttt{gemma-7b-it} as the LLM backbone. MoLE includes one shared expert and eight LoRA experts with a rank of eight. A two-layer MLP with a hidden dimension of 256 and ReLU activation serves as the gating mechanism, with a temperature set to 1. 

We train the reward models using the AdamW optimizer with a learning rate of \(5\times10^{-5}\) and a cosine scheduler with warmup ratio \(0.03\). Four GPUs, such as RTX6000ADA, L40S, and A100-PCIE-40GB, are employed with a batch size of 32 per GPU for \texttt{gemma-2b-it} and 16 per GPU for \texttt{gemma-7b-it}.

Baseline models use LoRA with rank 64. They also trained with an AdamW optimizer and a cosine scheduler with a warmup ratio \(0.03\). We search the learning rate from $[1\times 10^{-4}, 5 \times 10^{-5}, 1\times 10^{-5}, 5\times 10^{-6}]$.

\paragraph{User adaptation.}
We use a two-hop seen user and $0.07$ as temperature for unseen user adaptation of CoPL. For I2E, each learnable user representation is mapped into each user. For I2E$_\text{proxy}$ and PAL, user representations are determined by $N=10$ proxies. Adapting to an unseen user requires parameter optimization for unseen users, typically through several gradient steps. To optimize the parameters for unseen users, 50 gradient steps are applied during adaptation.

\section{Additional Experimental Results}
\label{appendix.experimental_results}

\begin{figure*}[t]
    \centering
    \begin{subfigure}[b]{0.32\textwidth}
        \centering
        \includegraphics[width=\linewidth]{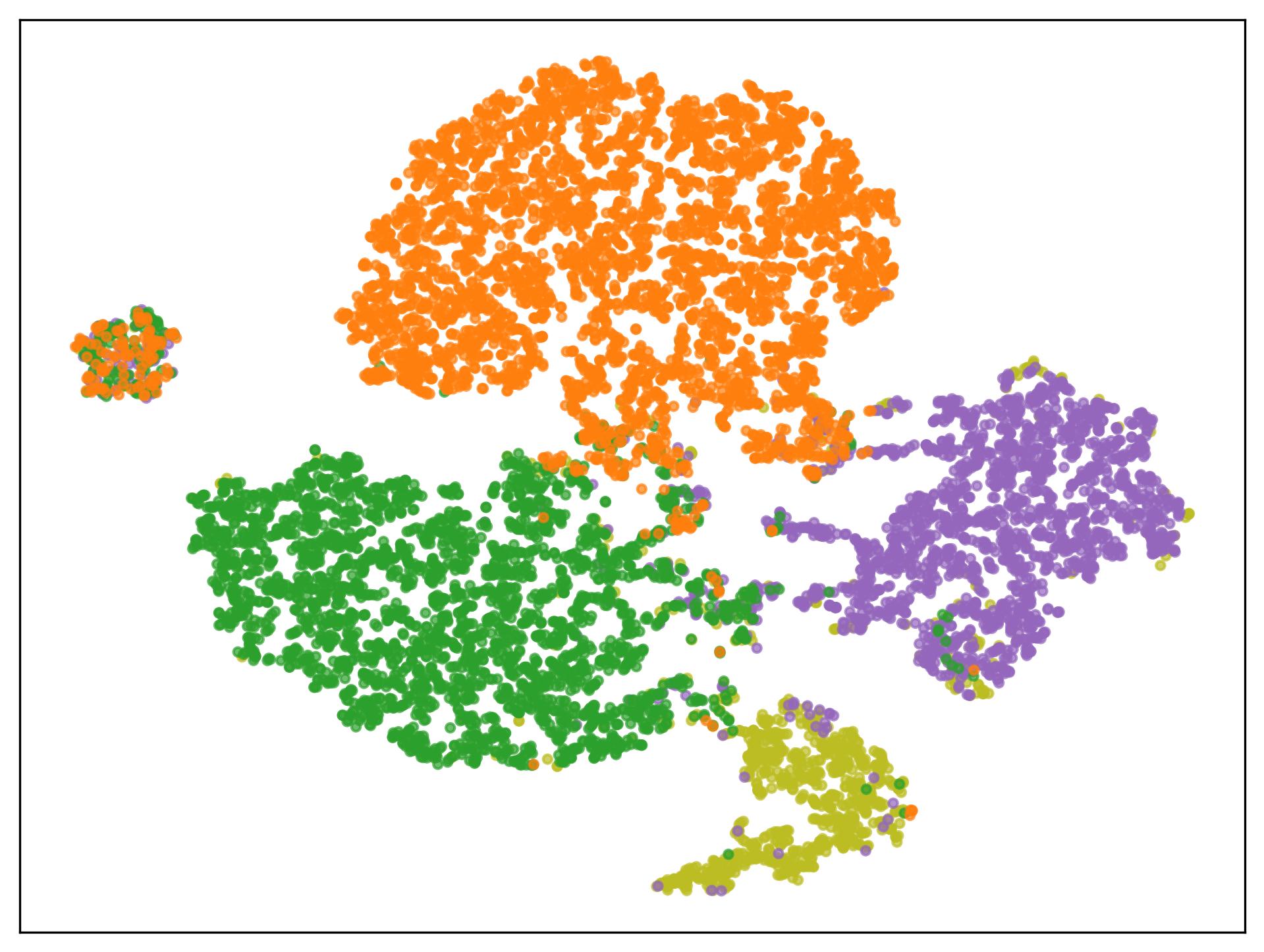}
        \caption{1:2:3:4}
    \end{subfigure}
    \hfill
    \begin{subfigure}[b]{0.32\textwidth}
        \centering
        \includegraphics[width=\linewidth]{fig/UF-P-4-AVG-ours_s.jpg}
        \caption{1:1:1:1}
    \end{subfigure}
    \hfill
    \begin{subfigure}[b]{0.32\textwidth}
        \centering
        \includegraphics[width=\linewidth]{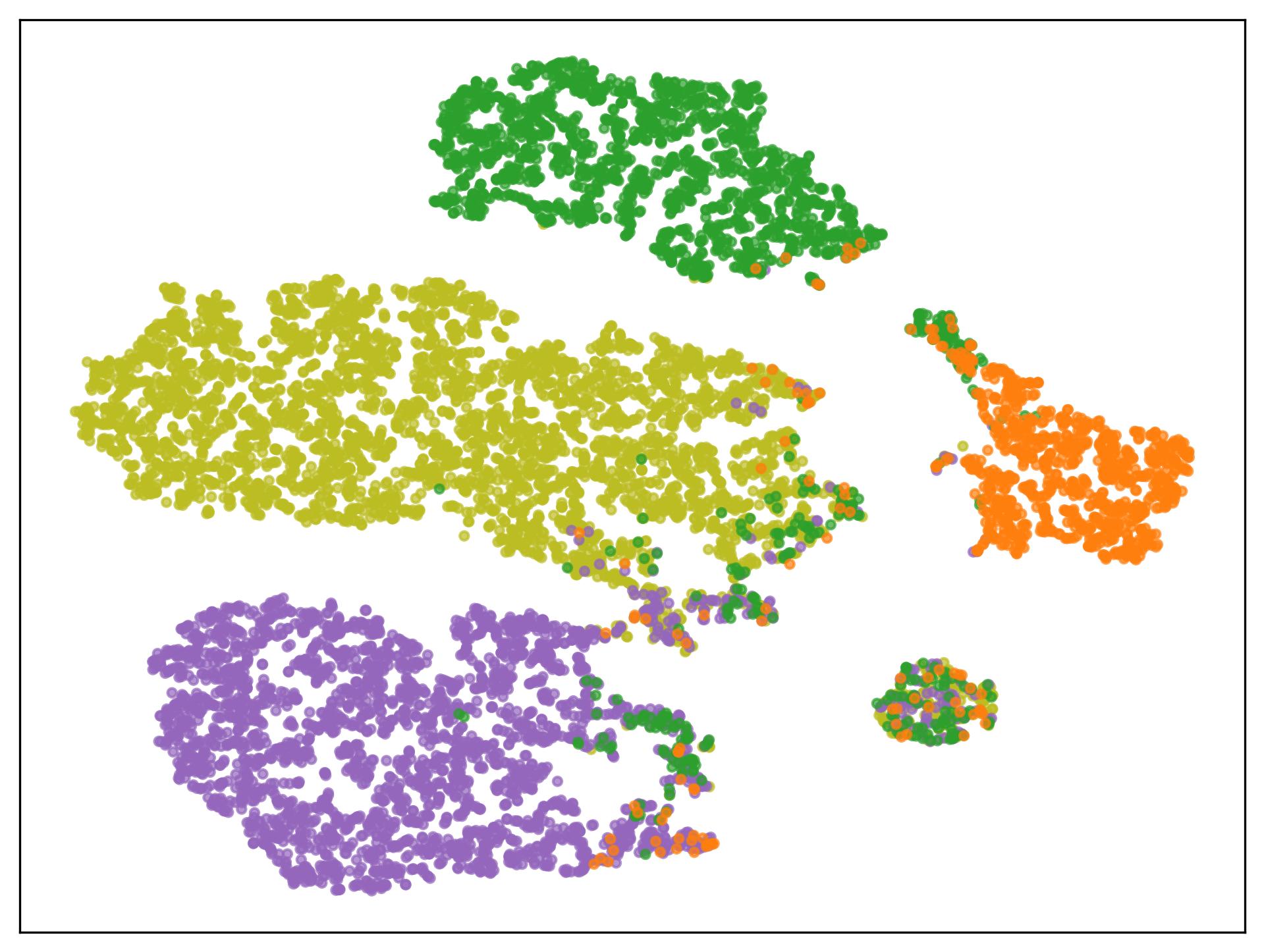}
        \caption{4:3:2:1}
    \end{subfigure}
    \caption{T-SNE visualization of user embeddings on UF-P-4 (AVG) under group ratios 1{:}2{:}3{:}4, 1{:}1{:}1{:}1, 4{:}3{:}2{:}1. Points are colored by preference group.}
    \label{fig:groupratio-ufp4}
\end{figure*}

\paragraph{Performance under the imbalanced group distribution with UF-P-4 (AVG).}
{\cref{tab:groupwise_accuracy} reports group-wise accuracies for the four group UF-P-4 (AVG) setting under selected imbalance configurations. The results exhibit the same trend seen in the two-group setting. CoPL continues to capture diverse user preferences across all groups. As the distribution departs from the balanced 1{:}1{:}1{:}1 setting, the gap from the balanced baseline widens. The lower absolute accuracy of some groups is largely due to the intrinsic difficulty of their preferences rather than the imbalance itself. This interpretation is supported by the G-Oracle. \cref{fig:groupratio-ufp4} visualizes the learned user embeddings. The embeddings form well-separated clusters aligned with group identities even under strong imbalance, which suggests that the representation remains stable, although predictive performance on minority groups drops.}

\begin{figure}[!t]
    \centering
    \begin{subfigure}[b]{0.48\columnwidth}
        \centering
        \includegraphics[width=\textwidth]{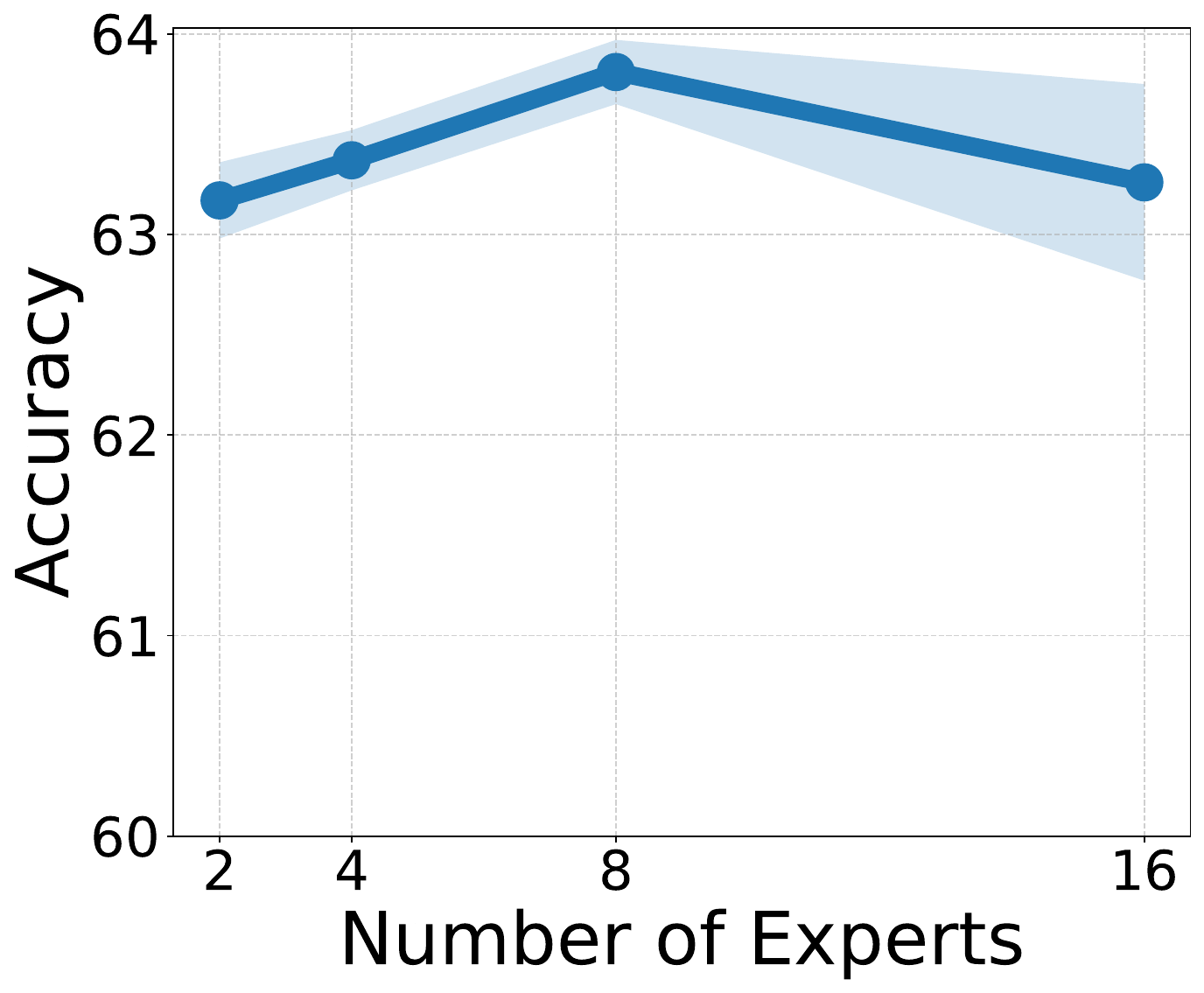}
        \caption{UF-P-2 (ALL)}
    \end{subfigure}
    \hfill
    \begin{subfigure}[b]{0.48\columnwidth}
        \centering
        \includegraphics[width=\textwidth]{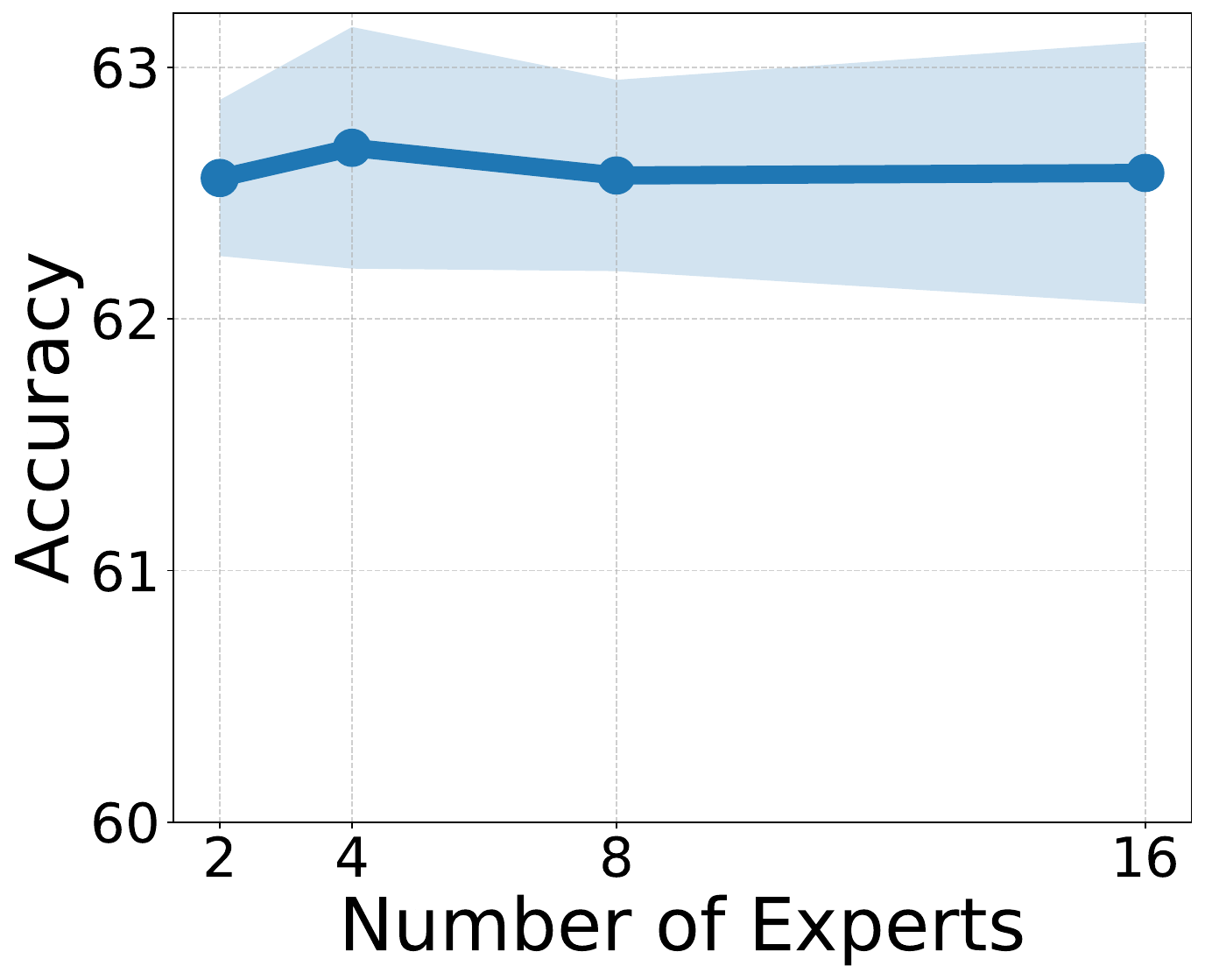}
        \caption{UF-P-4 (ALL)}
    \end{subfigure}
\caption{Ablation study on the number of experts in UF-P-2 and UF-P-4 (ALL) with \texttt{gemma-2b-it}.}
\label{fig:ablation.num_experts}
\end{figure}

\paragraph{Ablation study of the number of users.}
{\cref{fig:ablation.num_experts} shows that \ours{} performs robustly across different expert counts. This indicates that a moderate number of experts is generally sufficient to capture diverse user preferences.} 

\paragraph{{Performance with large-scale LLM.}}
{To assess scalability, we instantiate CoPL with gemma-2-27B-it~\citep{team2024gemma2} and evaluate on UF-P-4 (ALL) and PersonalLLM (ALL). We use a single seed due to hardware limits and compare with VPL, the strongest baseline. As shown in \cref{ablation.large-rm}, CoPL surpasses VPL on both datasets, indicating that the gains carry over to larger model scales. These results support the scalability of CoPL beyond the settings used in the main experiments.}

\paragraph{Ablation study of message-passing.}
Inspired by the previous work~\cite{he2020lightgcn} in recommendation systems, we first omit the non-linear activation and feature transformation matrix used in \cref{eq:message}, and also investigate the effectiveness of negative edges. As shown in ~\cref{tab:ablation}, incorporating negative edges consistently improves accuracy. Notably, our proposed message-passing achieves the highest accuracy, highlighting both the effectiveness of our message-passing operation and the advantage of modeling negative edges.

\begin{table*}[t!]
\centering
\resizebox{\textwidth}{!}{
\begin{tabular}{llrrrrrrrr}
\toprule
& & \multicolumn{2}{c}{TL;DR} & \multicolumn{2}{c}{UF-P-2} & \multicolumn{2}{c}{UF-P-4} & \multicolumn{2}{c}{PersonalLLM} \\
\cmidrule(lr){3-4}\cmidrule(lr){5-6}\cmidrule(lr){7-8}\cmidrule(lr){9-10}
& & \multicolumn{1}{c}{ALL} & \multicolumn{1}{c}{AVG} & \multicolumn{1}{c}{ALL} & \multicolumn{1}{c}{AVG} & \multicolumn{1}{c}{ALL} & \multicolumn{1}{c}{AVG} & \multicolumn{1}{c}{ALL} & \multicolumn{1}{c}{AVG} \\
\midrule
\multirow{7}{*}{\rotatebox[origin=c]{90}{Seen}} 
& G-Oracle   
& $77.21_{\pm0.28}$ & $77.21_{\pm0.28}$ 
& $66.80_{\pm0.17}$ & $66.80_{\pm0.17}$ 
& $62.17_{\pm0.09}$ & $62.17_{\pm0.09}$ 
& N/A & N/A \\ \cmidrule(lr){2-10}

& Uniform  
& $49.39_{\pm0.52}$ & $49.39_{\pm0.52}$ 
& $61.96_{\pm0.07}$ & $61.96_{\pm0.07}$ 
& $56.80_{\pm0.12}$ & $56.80_{\pm0.12}$ 
& $63.64_{\pm0.30}$ & $63.64_{\pm0.30}$ \\

& I2E      
& $49.40_{\pm0.77}$ & $49.66_{\pm0.31}$ 
& $62.10_{\pm0.28}$ & $61.43_{\pm0.23}$ 
& $57.90_{\pm0.21}$ & $58.50_{\pm0.09}$ 
& $66.40_{\pm0.38}$ & $65.86_{\pm0.12}$ \\

& I2E$_{\text{proxy}}$
& $49.50_{\pm0.73}$ & $49.95_{\pm0.34}$ 
& $62.03_{\pm0.30}$ & $62.27_{\pm0.09}$ 
& $57.54_{\pm0.16}$ & $58.12_{\pm0.14}$ 
& $66.58_{\pm0.35}$ & $65.70_{\pm0.02}$ \\

& VPL      
& $49.14_{\pm0.72}$ & $49.17_{\pm0.67}$ 
& $62.39_{\pm0.10}$ & $62.59_{\pm0.24}$ 
& $58.87_{\pm0.25}$ & $57.55_{\pm1.00}$ 
& $70.55_{\pm0.16}$ & $66.18_{\pm0.01}$ \\

& PAL      
& $49.57_{\pm0.09}$ & $49.75_{\pm0.27}$ 
& $62.59_{\pm0.06}$ & $62.47_{\pm0.13}$ 
& $57.17_{\pm0.22}$ & $56.27_{\pm0.13}$ 
& $66.46_{\pm0.49}$ & $65.43_{\pm0.43}$ \\

& \ours{}  
& $\mathbf{97.85}_{\pm0.07}$ & $\mathbf{97.88}_{\pm0.01}$
& $\mathbf{63.90}_{\pm0.07}$ & $\mathbf{63.48}_{\pm0.13}$
& $\mathbf{62.90}_{\pm0.05}$ & $\mathbf{61.93}_{\pm0.02}$
& $\mathbf{74.87}_{\pm0.19}$ & $\mathbf{74.76}_{\pm0.01}$ \\
\midrule

\multirow{7}{*}{\rotatebox[origin=c]{90}{Unseen}} 
& G-Oracle   
& $77.54_{\pm0.49}$ & $77.54_{\pm0.49}$ 
& $67.43_{\pm0.65}$ & $67.43_{\pm0.65}$ 
& $62.01_{\pm0.04}$ & $62.01_{\pm0.04}$ 
& N/A & N/A \\ \cmidrule(lr){2-10}

& Uniform  
& $49.03_{\pm0.76}$ & $49.03_{\pm0.76}$ 
& $62.23_{\pm0.06}$ & $62.23_{\pm0.06}$ 
& $57.02_{\pm0.27}$ & $57.02_{\pm0.27}$ 
& $63.30_{\pm0.08}$ & $63.30_{\pm0.08}$ \\

& I2E      
& $49.64_{\pm0.98}$ & $49.56_{\pm0.49}$ 
& $62.62_{\pm0.95}$ & $61.88_{\pm0.21}$ 
& $57.62_{\pm0.92}$ & $58.12_{\pm0.98}$ 
& $65.75_{\pm0.38}$ & $65.74_{\pm0.37}$ \\

& I2E$_{\text{proxy}}$
& $49.68_{\pm1.35}$ & $49.19_{\pm1.06}$ 
& $61.99_{\pm0.33}$ & $62.84_{\pm0.40}$ 
& $57.69_{\pm0.70}$ & $57.73_{\pm0.32}$ 
& $66.47_{\pm0.08}$ & $66.13_{\pm0.33}$ \\

& VPL      
& $49.07_{\pm0.65}$ & $48.92_{\pm0.72}$ 
& $62.69_{\pm0.99}$ & $63.67_{\pm0.12}$ 
& $58.49_{\pm1.22}$ & $56.85_{\pm0.84}$ 
& $69.93_{\pm0.33}$ & $65.72_{\pm0.42}$ \\

& PAL      
& $49.71_{\pm0.44}$ & $49.68_{\pm0.34}$ 
& $63.08_{\pm0.73}$ & $62.52_{\pm0.58}$ 
& $57.15_{\pm0.48}$ & $56.44_{\pm0.67}$ 
& $66.57_{\pm0.08}$ & $65.92_{\pm0.25}$ \\

& \ours{}  
& $\mathbf{97.95}_{\pm0.15}$ & $\mathbf{98.19}_{\pm0.06}$
& $\mathbf{64.08}_{\pm0.71}$ & $\mathbf{64.38}_{\pm1.00}$
& $\mathbf{62.77}_{\pm1.32}$ & $\mathbf{62.08}_{\pm0.64}$
& $\mathbf{74.84}_{\pm0.18}$ & $\mathbf{75.64}_{\pm0.05}$ \\
\bottomrule
\end{tabular}
}
\caption{Accuracy of reward models on unseen annotated pairs. The results report performance on \emph{Seen users} encountered during training and on \emph{Unseen users}, which consist of 100 new users evenly distributed across preference groups. Unseen users provide 8 annotations under TL;DR/UF-P-2 (ALL/AVG) and 16 annotations under UF-P-4/PersonalLLM (ALL/AVG). \textbf{Bold} represents the best result, except for G-Oracle. N/A indicates that training reward models for each group is infeasible for PersonalLLM, as this dataset does not clearly partition users into discrete groups. All experiments run on three seeds. These results are based on \texttt{gemma-7b-it}.}
\label{tab:combined_7b}
\end{table*}

\begin{figure*}[t!]
    \centering
    \begin{subfigure}{0.19\textwidth}
        \centering
        \includegraphics[width=\linewidth]{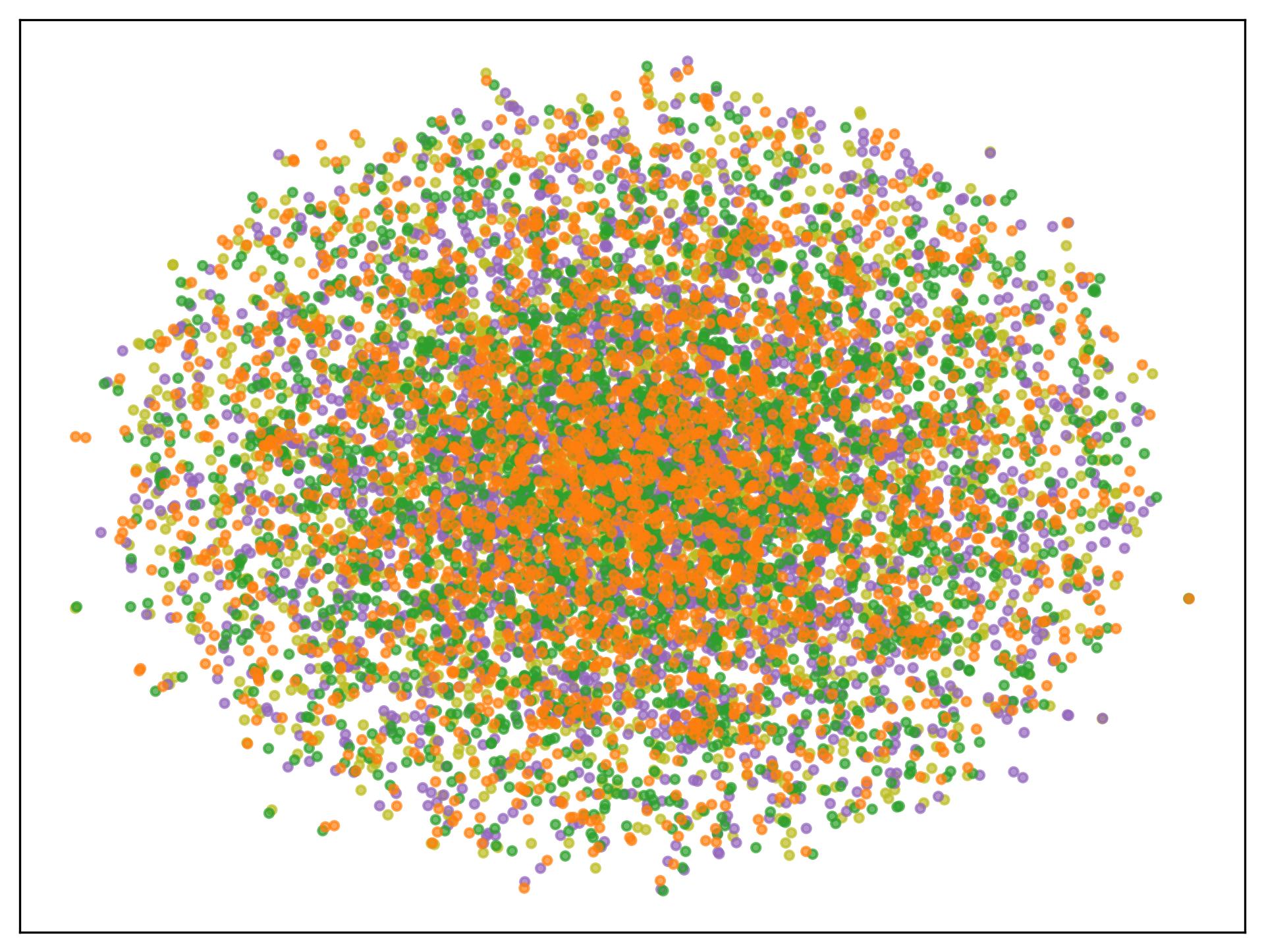}
        \caption{I2E}
    \end{subfigure}
    \hfill
    \begin{subfigure}{0.19\textwidth}
        \centering
        \includegraphics[width=\linewidth]{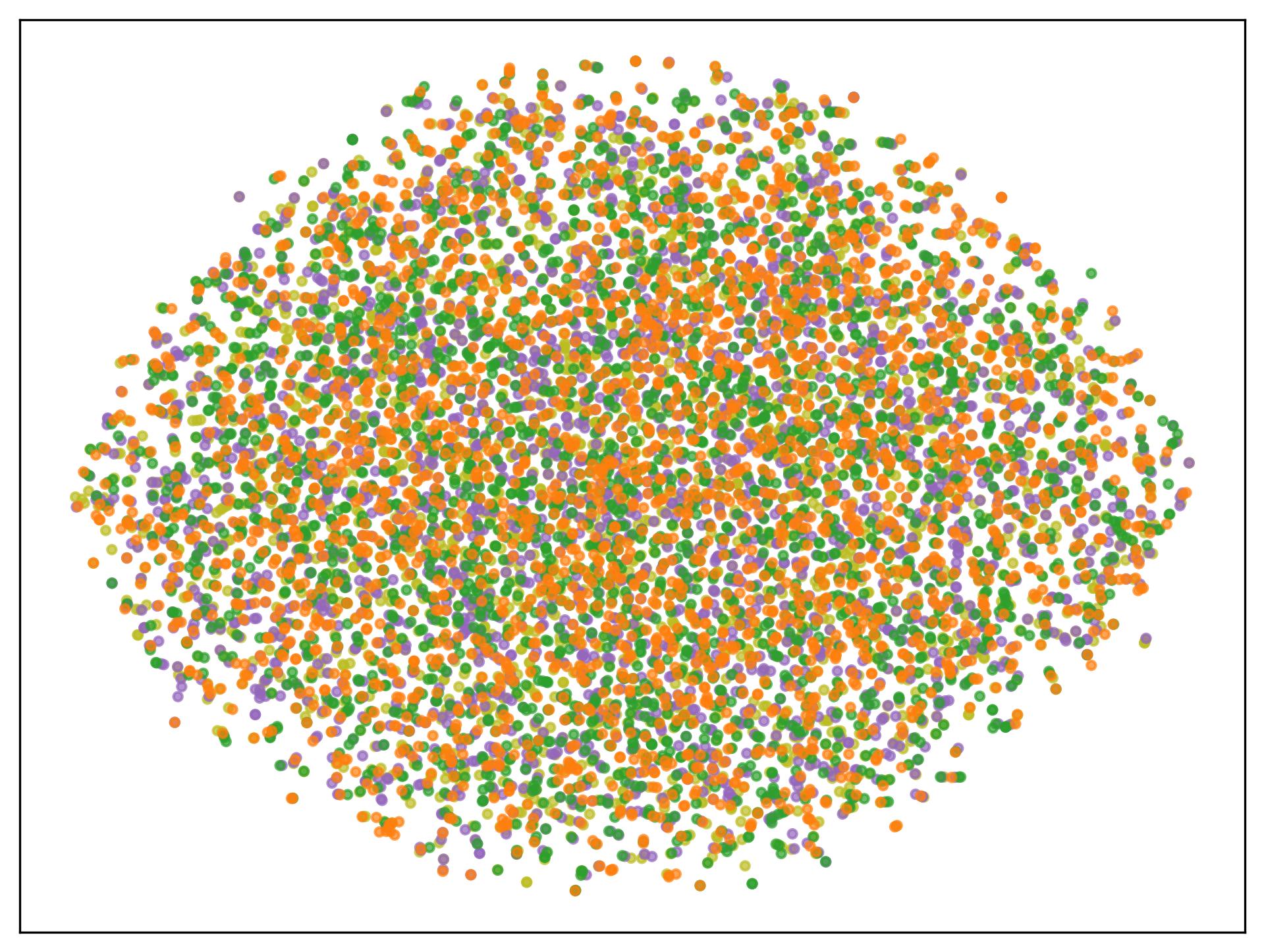}
        \caption{I2E$_\text{{proxy}}$}
    \end{subfigure}
    \hfill
    \begin{subfigure}{0.19\textwidth}
        \centering
        \includegraphics[width=\linewidth]{fig/UF-P-4-AVG-VPL_s.jpg}
        \caption{VPL}
    \end{subfigure}
    \hfill
    \begin{subfigure}{0.19\textwidth}
        \centering
        \includegraphics[width=\linewidth]{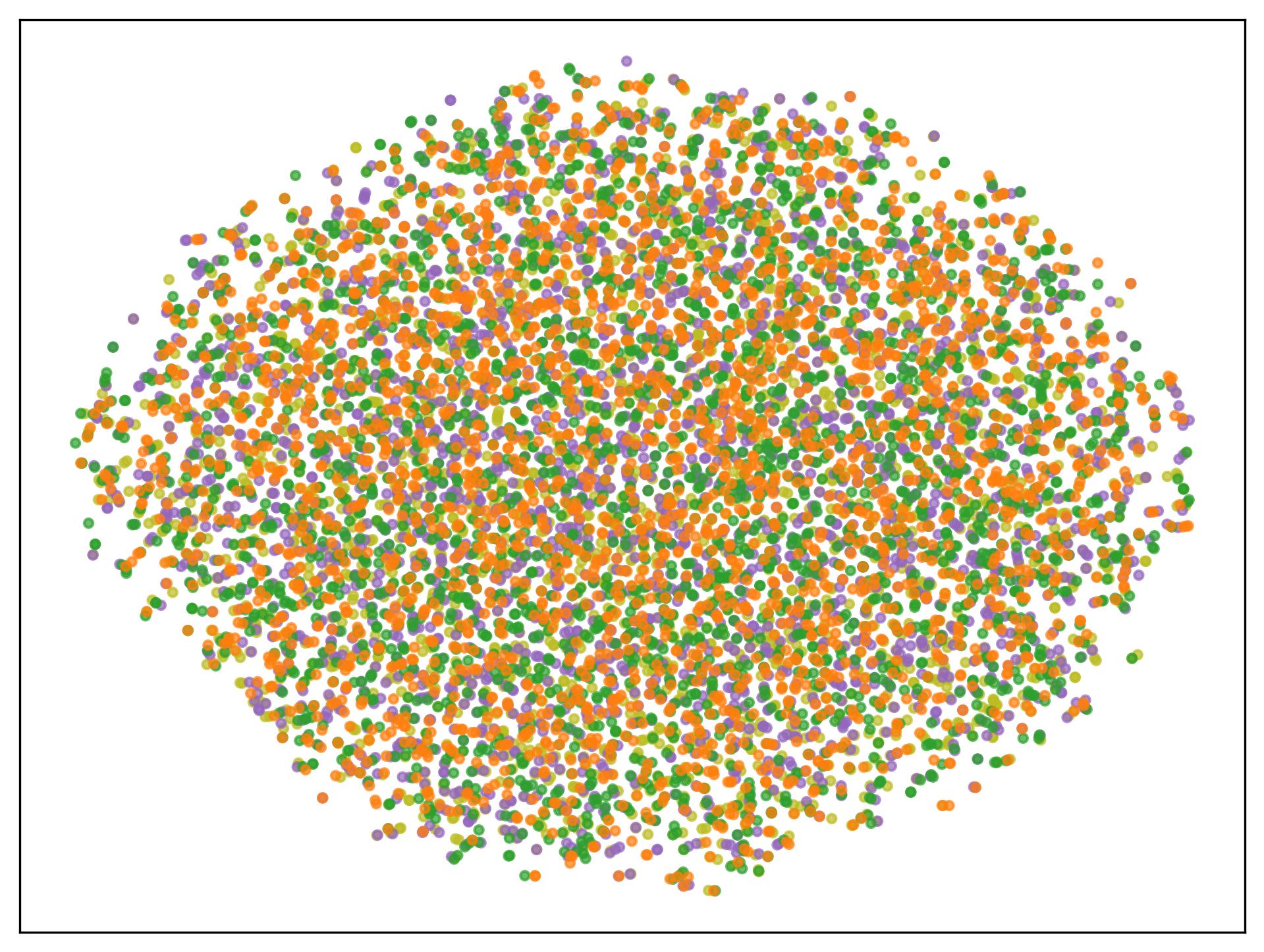}
        \caption{PAL}
    \end{subfigure}
    \hfill
    \begin{subfigure}{0.19\textwidth}
        \centering
        \includegraphics[width=\linewidth]{fig/UF-P-4-AVG-ours_s.jpg}
        \caption{\ours{}}
    \end{subfigure}
    \caption{T-SNE visualization of seen user embeddings in UF-P-4 (AVG) with \texttt{gemma-2b-it}. Points are colored by their preference group.  Our method clusters users in the same group more effectively, whereas other baselines fail to cluster users by their preference groups in the user embedding space.}
    \label{fig:embedding_space}
\end{figure*}

\begin{figure*}[t!]
    \centering
    \begin{subfigure}[b]{0.32\textwidth}
        \centering
        \includegraphics[width=\linewidth]{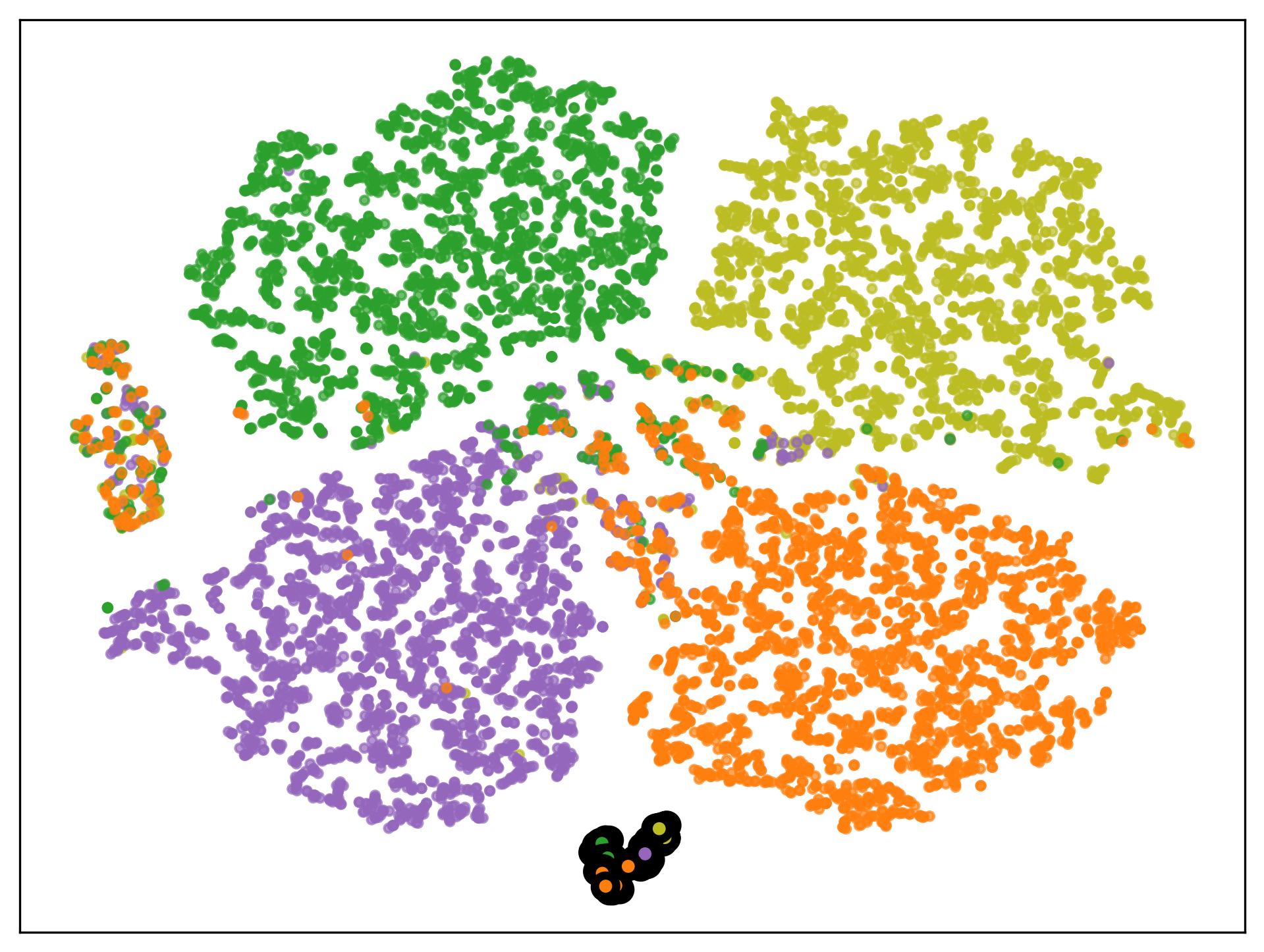}
        \caption{Naive Avg.}
    \end{subfigure}
    \hfill
    \begin{subfigure}[b]{0.32\textwidth}
        \centering
        \includegraphics[width=\linewidth]{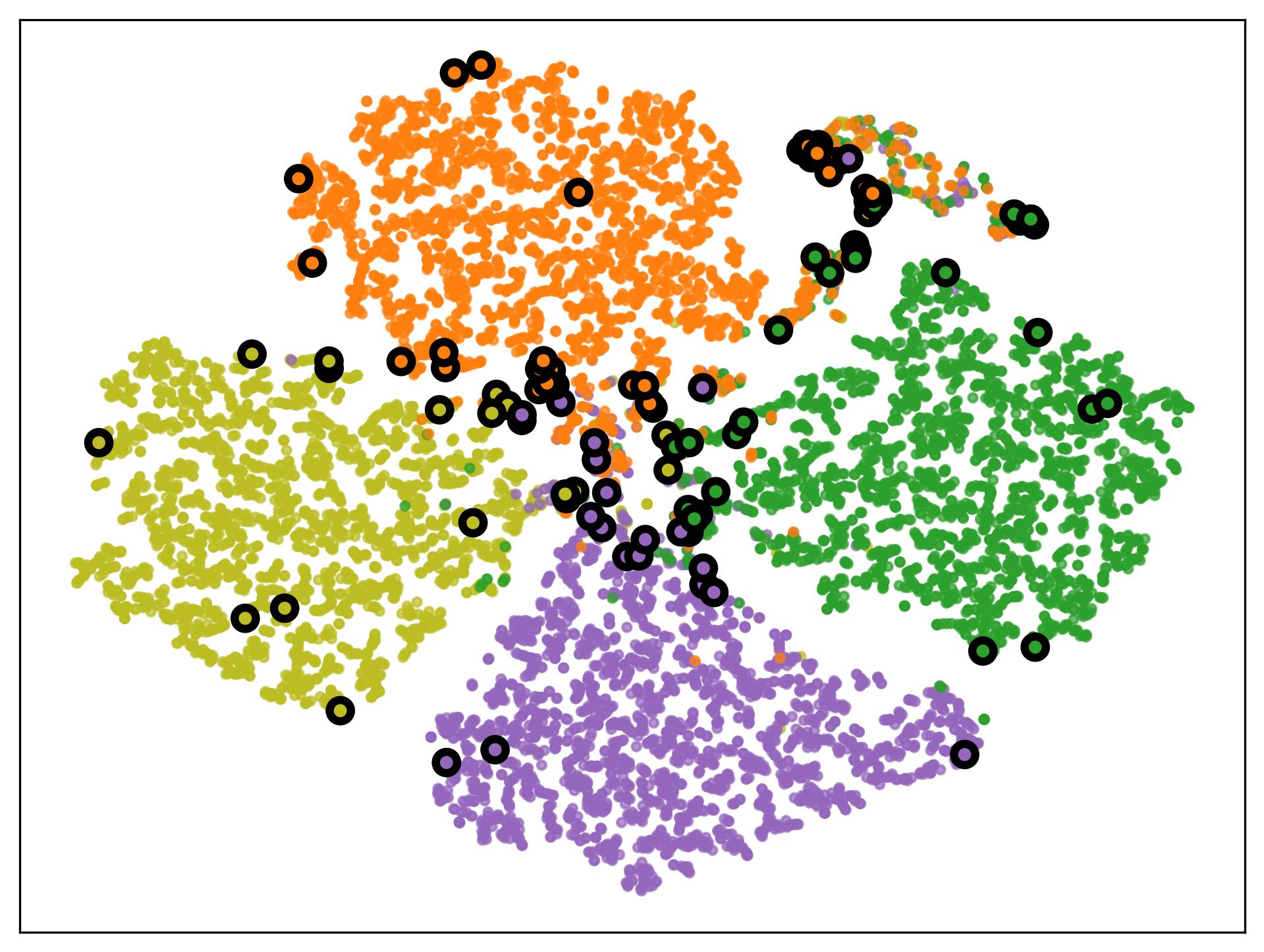}
        \caption{User Opt.}
    \end{subfigure}
    \hfill
    \begin{subfigure}[b]{0.32\textwidth}
        \centering
        \includegraphics[width=\linewidth]{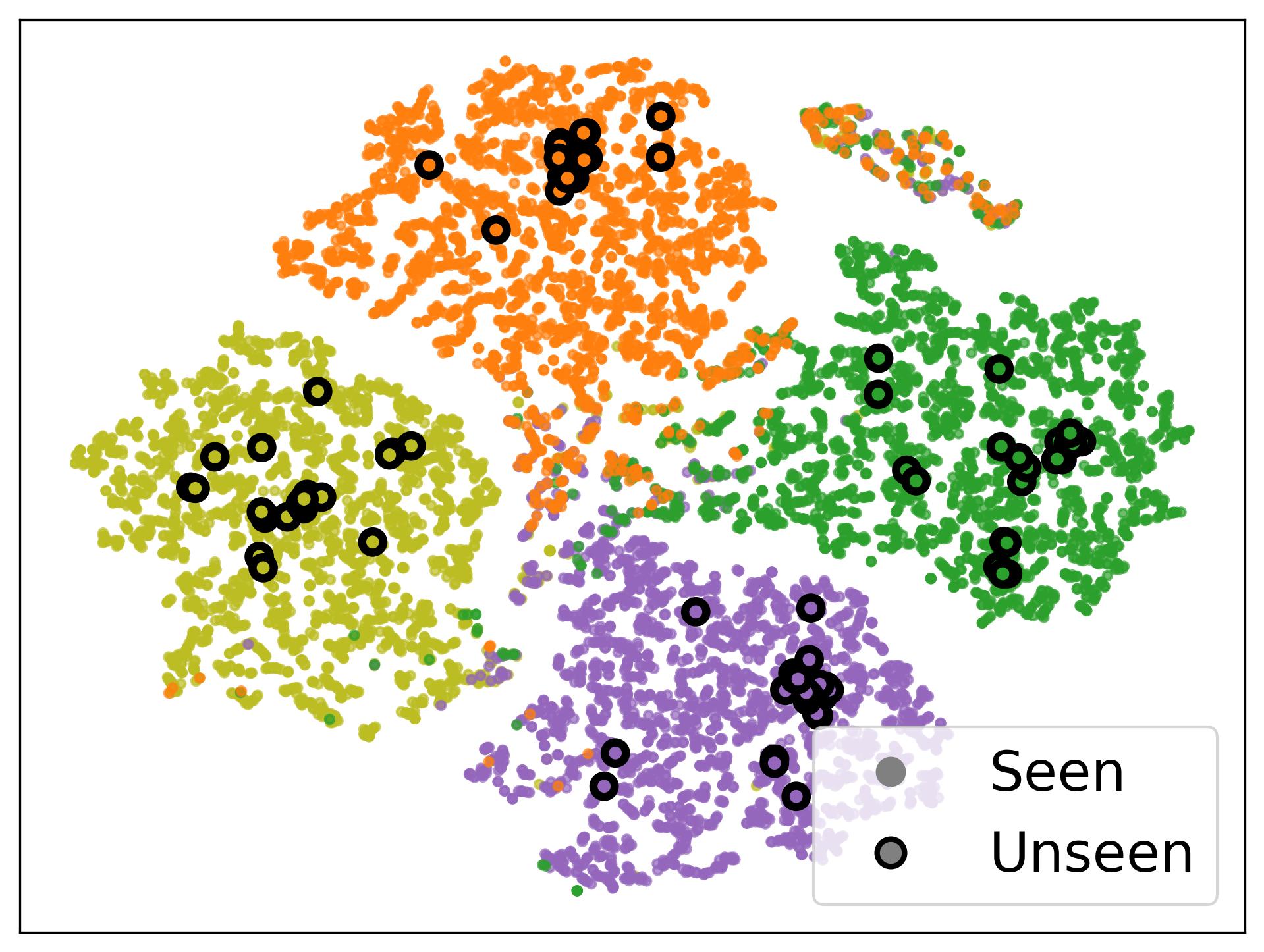}
        \caption{\ours{}}
    \end{subfigure}
    \caption{T-SNE visualization of seen and unseen user embeddings in UF-P-4-AVG. \emph{Naive Avg.} computes unseen user embeddings as the unweighted mean of 2-hop neighbor embeddings. \emph{User Opt.} represents an optimization-based approach that learns a parameterized user embedding by maximizing the likelihood of the given annotations. Colors indicate preference groups, and points with black edges represent unseen users. Unseen users adapted by our method align with their respective preference groups.}
    \label{fig:comparison}
\end{figure*}

\begin{table}[h]
\centering
\resizebox{\columnwidth}{!}{
\begin{tabular}{lcccc}
\toprule
          & \textit{Helpfulness}     & \textit{Honesty}         & \textit{I.F.} & \textit{Truthfulness}    \\
\midrule
G-Oracle    & $67.99_{\pm 0.52}$ & $58.38_{\pm 0.38}$ & $61.00_{\pm 0.16}$      & $58.73_{\pm 0.40}$ \\
\midrule
1:2:3:4   & $64.20_{\pm 1.15}$ & $57.56_{\pm 0.03}$ & $62.28_{\pm 1.01}$      & $58.40_{\pm 0.04}$ \\
1:1:1:1   & $71.56_{\pm 0.67}$ & $57.46_{\pm 0.28}$ & $61.55_{\pm 0.17}$      & $57.17_{\pm 0.25}$ \\
4:3:2:1   & $71.25_{\pm 0.25}$ & $56.58_{\pm 0.57}$ & $61.08_{\pm 0.29}$      & $52.55_{\pm 0.36}$ \\
\bottomrule
\end{tabular}
}
\caption{Group-wise accuracy of reward models with Gemma-2b-it in UF-P-4 (AVG), varying the ratio of group size with the total number of users fixed at 10,000. \textit{I.F.} means \textit{Instruction Following}.}
\label{tab:groupwise_accuracy}
\end{table}
\begin{table}[h]
\centering
\begin{tabular}{lcc}
\toprule
       & UF-P-4 (ALL) & PersonalLLM (ALL) \\
\midrule
VPL    & 58.96        & 70.92             \\
CoPL   & \textbf{63.17} & \textbf{74.30}  \\
\bottomrule
\end{tabular}
\caption{Accuracy of reward models with Gemma-2-27b-it in UF-P-4 (ALL) and PersonalLLM (ALL).}
\label{ablation.large-rm}
\end{table}


\begin{table}[h]
\centering
\begin{tabular}{l c}
\toprule 
  \ours{}  &  $\textbf{84.84}_{\pm 0.83}$  \\ \midrule
   w/o N.E.   & $72.94_{\pm 0.61}$ \\
   w/o Act. \& Trans.  & $80.61_{\pm 0.32}$ \\
   w/o Act. \& Trans. \& N.E. & $72.15_{\pm 1.02}$ \\
   \bottomrule
\end{tabular}
\caption{Test accuracy of GNN in UF-P-2-ALL. ``N.E.'' denotes the negative edges. ``Act.'' denotes the non-linear activation. ``Trans.'' denotes the feature transformation matrix.}
\label{tab:ablation}
\end{table}

\begin{figure*}[!t]
    \centering
    \begin{subfigure}{0.3\textwidth}
        \centering
        \includegraphics[width=\linewidth]{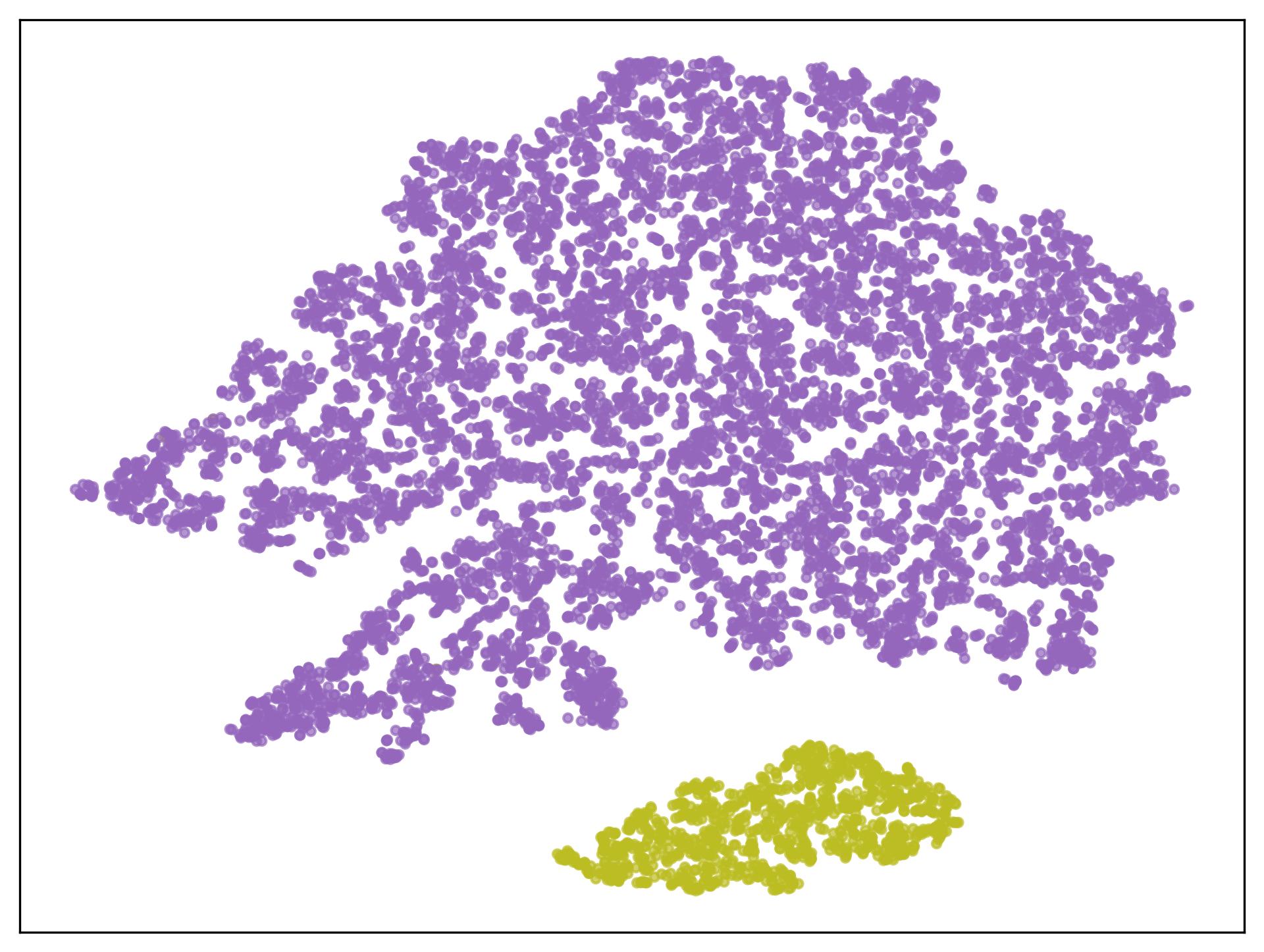}
        \caption{1:9}
    \end{subfigure}
    \hfill
    \begin{subfigure}{0.3\textwidth}
        \centering
        \includegraphics[width=\linewidth]{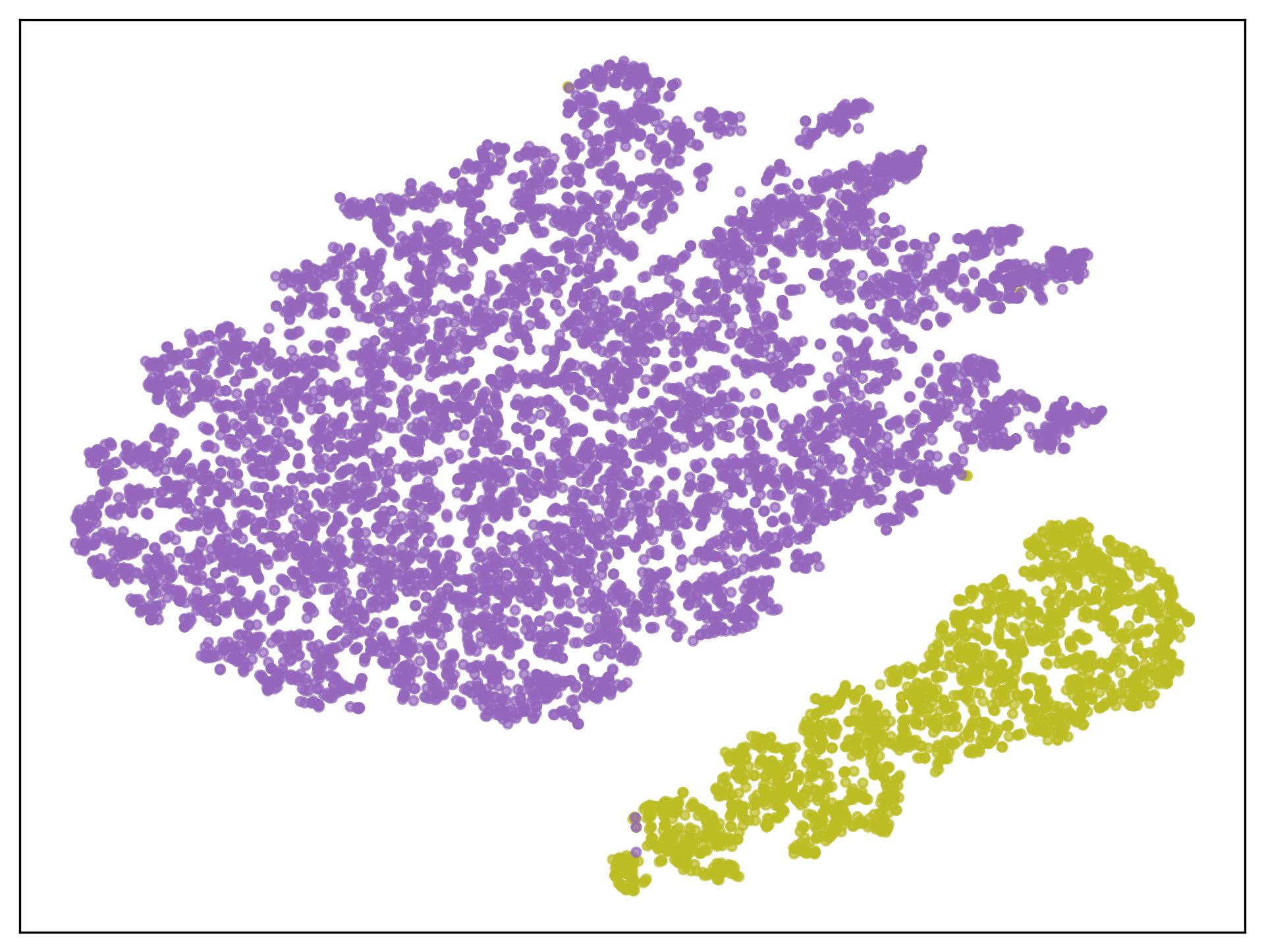}
        \caption{2:8}
    \end{subfigure}
    \hfill
    \begin{subfigure}{0.3\textwidth}
        \centering
        \includegraphics[width=\linewidth]{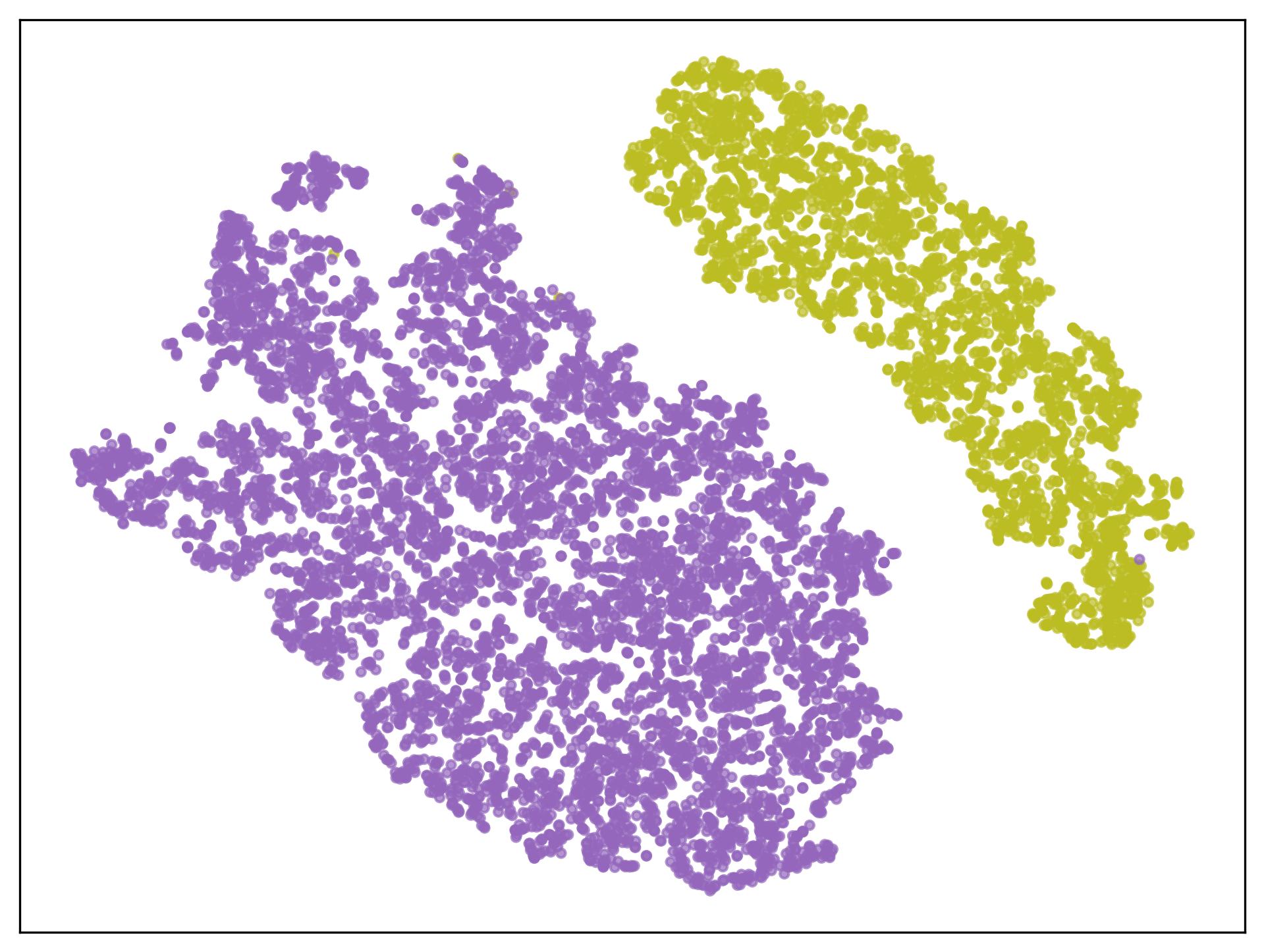}
        \caption{3:7}
    \end{subfigure}

    \vskip\baselineskip
    \begin{subfigure}{0.3\textwidth}
        \centering
        \includegraphics[width=\linewidth]{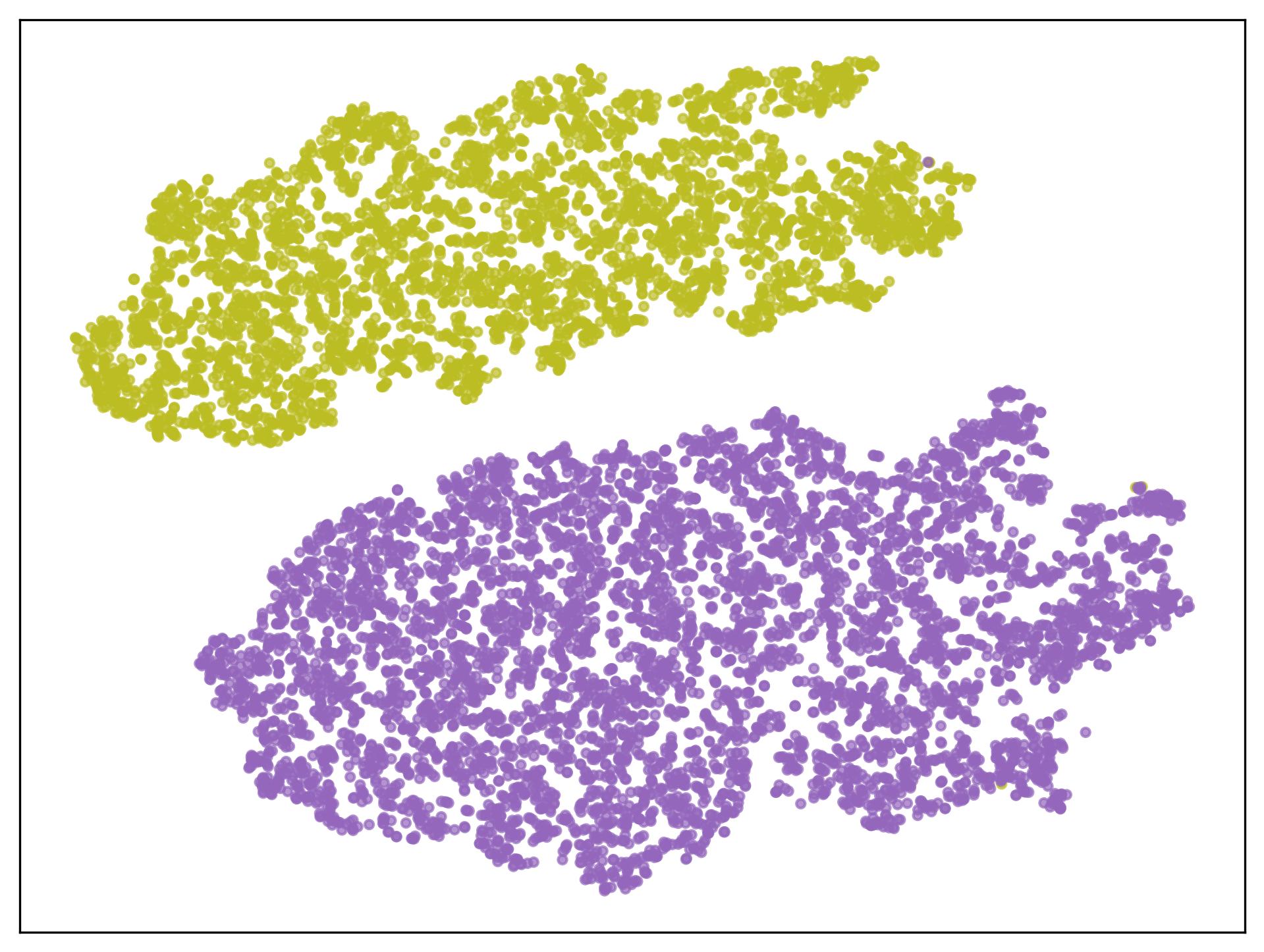}
        \caption{4:6}
    \end{subfigure}
    \hfill
    \begin{subfigure}{0.3\textwidth}
        \centering
        \includegraphics[width=\linewidth]{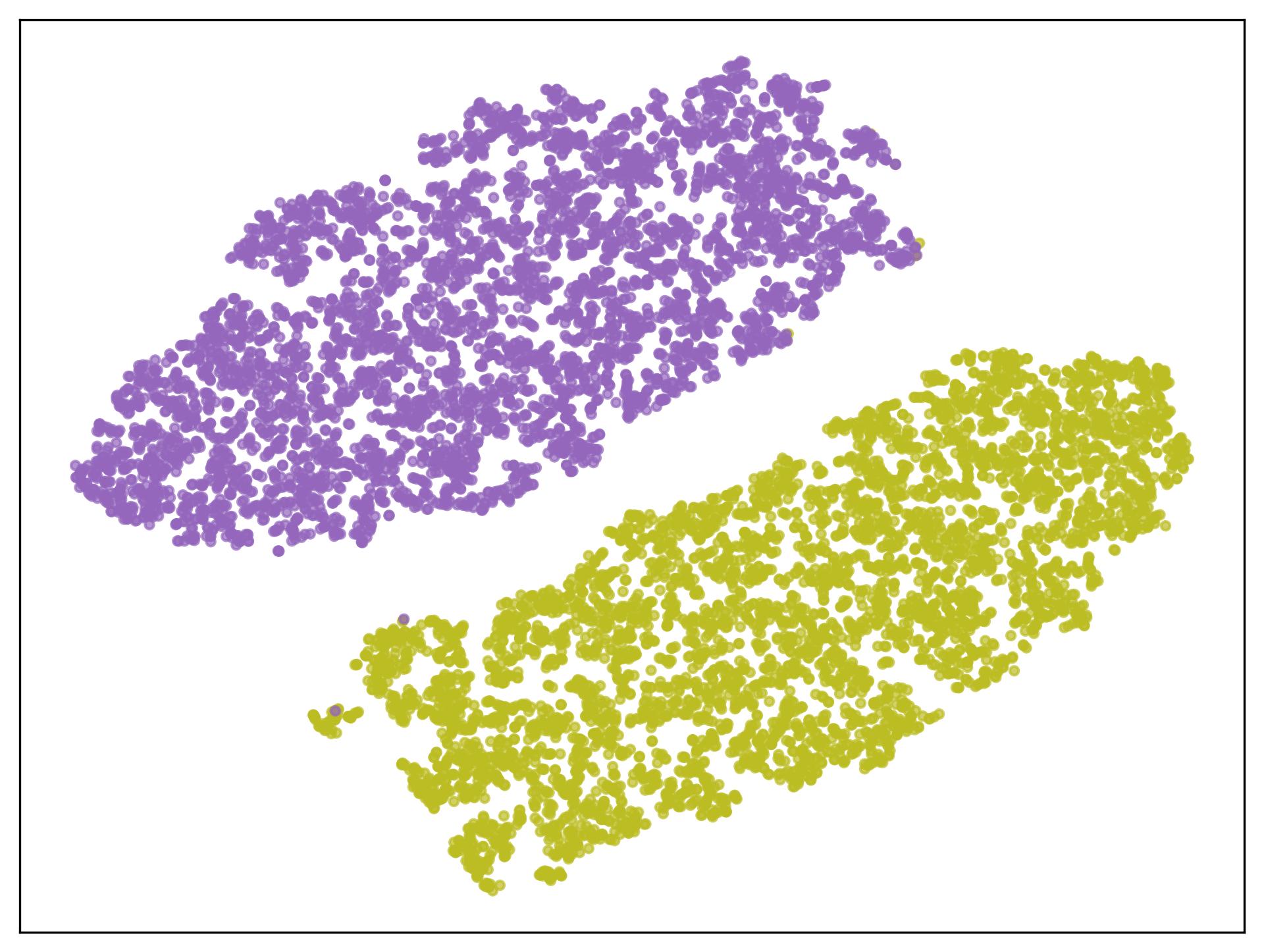}
        \caption{5:5}
    \end{subfigure}
    \hfill
    \begin{subfigure}{0.3\textwidth}
        \centering
        \includegraphics[width=\linewidth]{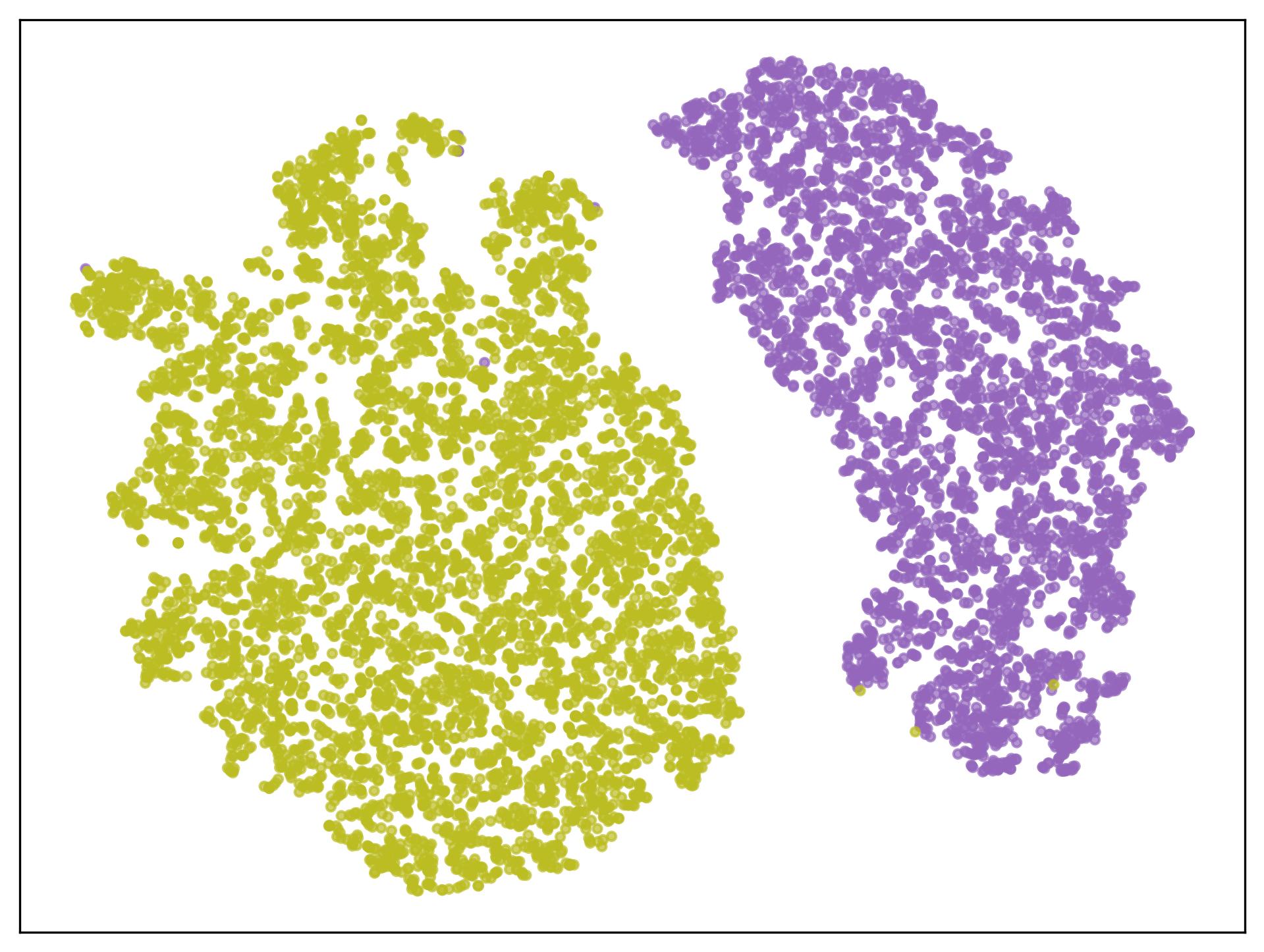}
        \caption{6:4}
    \end{subfigure}

    \vskip\baselineskip
    \begin{subfigure}{0.3\textwidth}
        \centering
        \includegraphics[width=\linewidth]{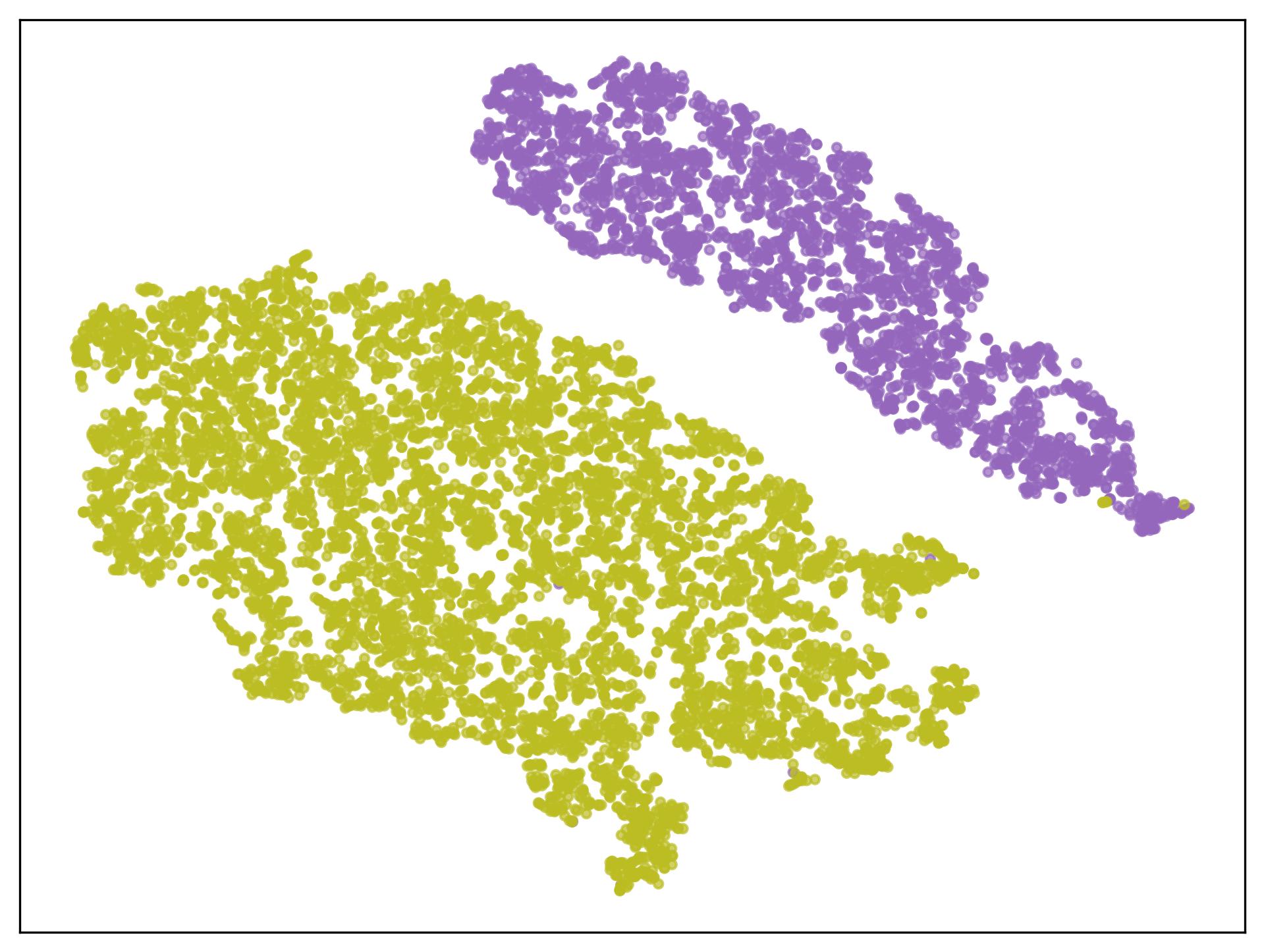}
        \caption{7:3}
    \end{subfigure}
    \hfill
    \begin{subfigure}{0.3\textwidth}
        \centering
        \includegraphics[width=\linewidth]{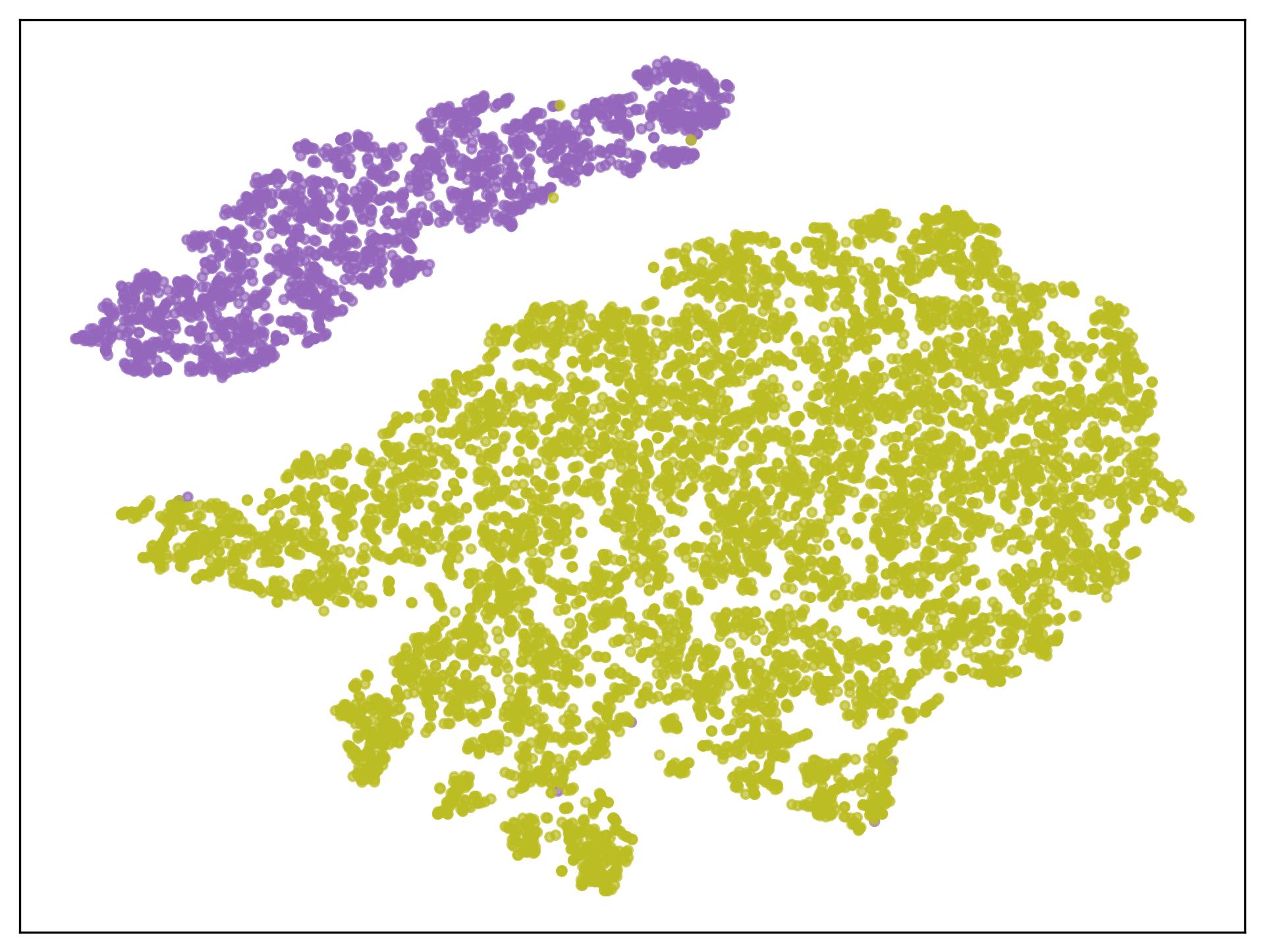}
        \caption{8:2}
    \end{subfigure}
    \hfill
    \begin{subfigure}{0.3\textwidth}
        \centering
        \includegraphics[width=\linewidth]{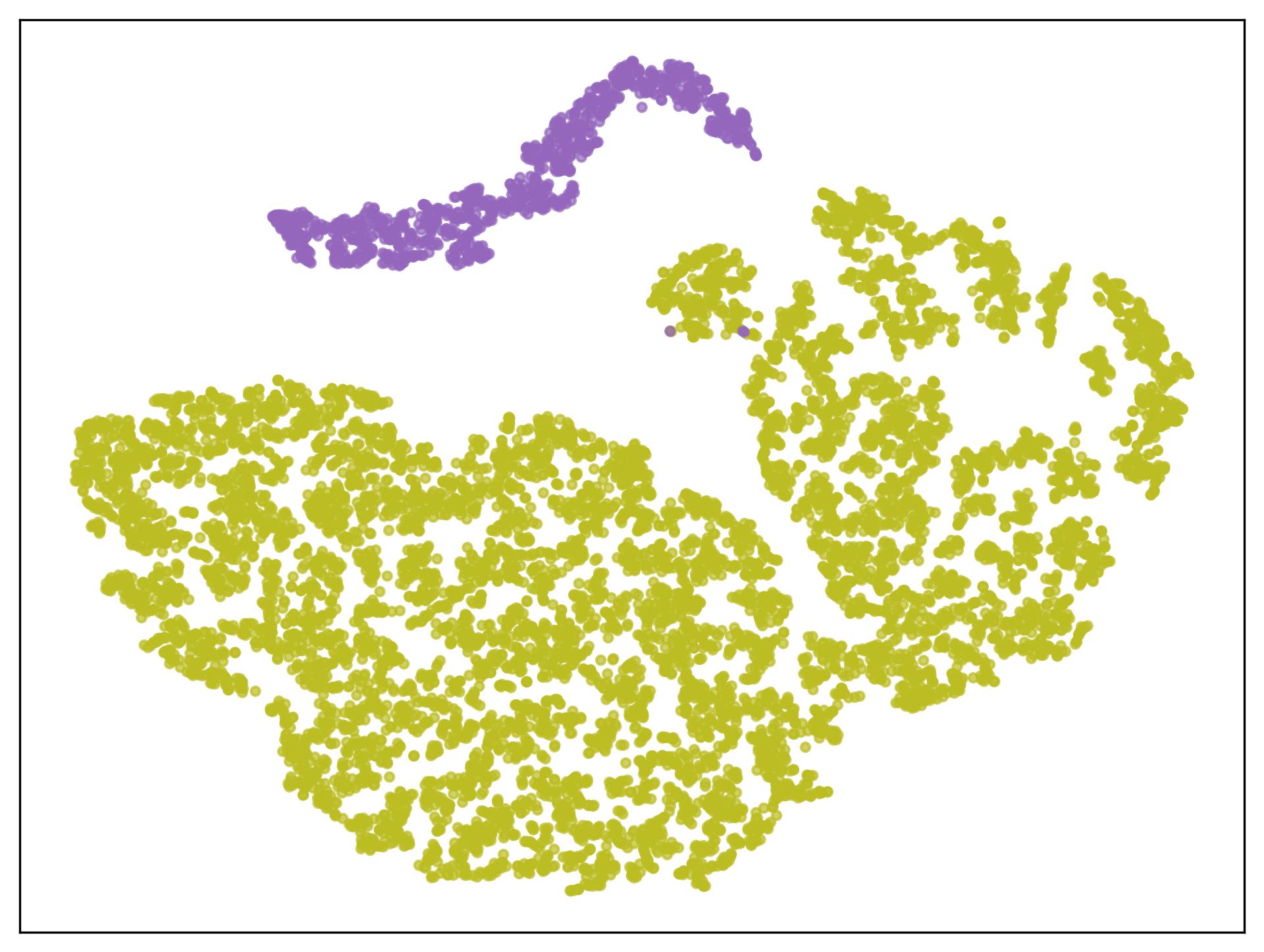}
        \caption{9:1}
    \end{subfigure}

    \caption{T-SNE visualization of user embeddings on TL;DR (AVG) across group ratios from 1:9 to 9:1. Points are colored by preference group.}
    \label{fig:embedding_space_ratios}
\end{figure*}

\begin{figure*}[!t]
    \centering
    \begin{subfigure}{0.3\textwidth}
        \centering
        \includegraphics[width=\linewidth]{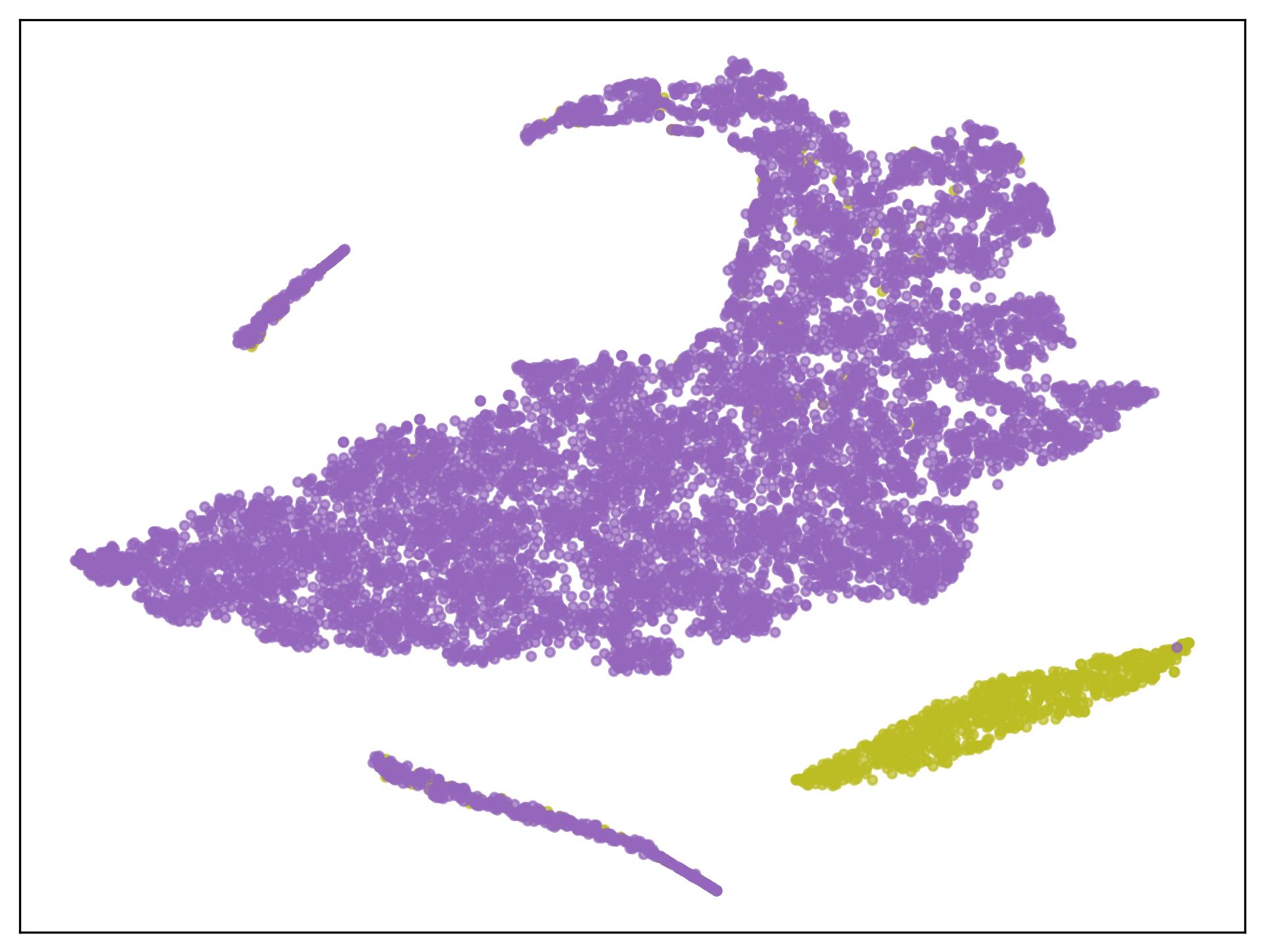}
        \caption{1:9}
    \end{subfigure}
    \hfill
    \begin{subfigure}{0.3\textwidth}
        \centering
        \includegraphics[width=\linewidth]{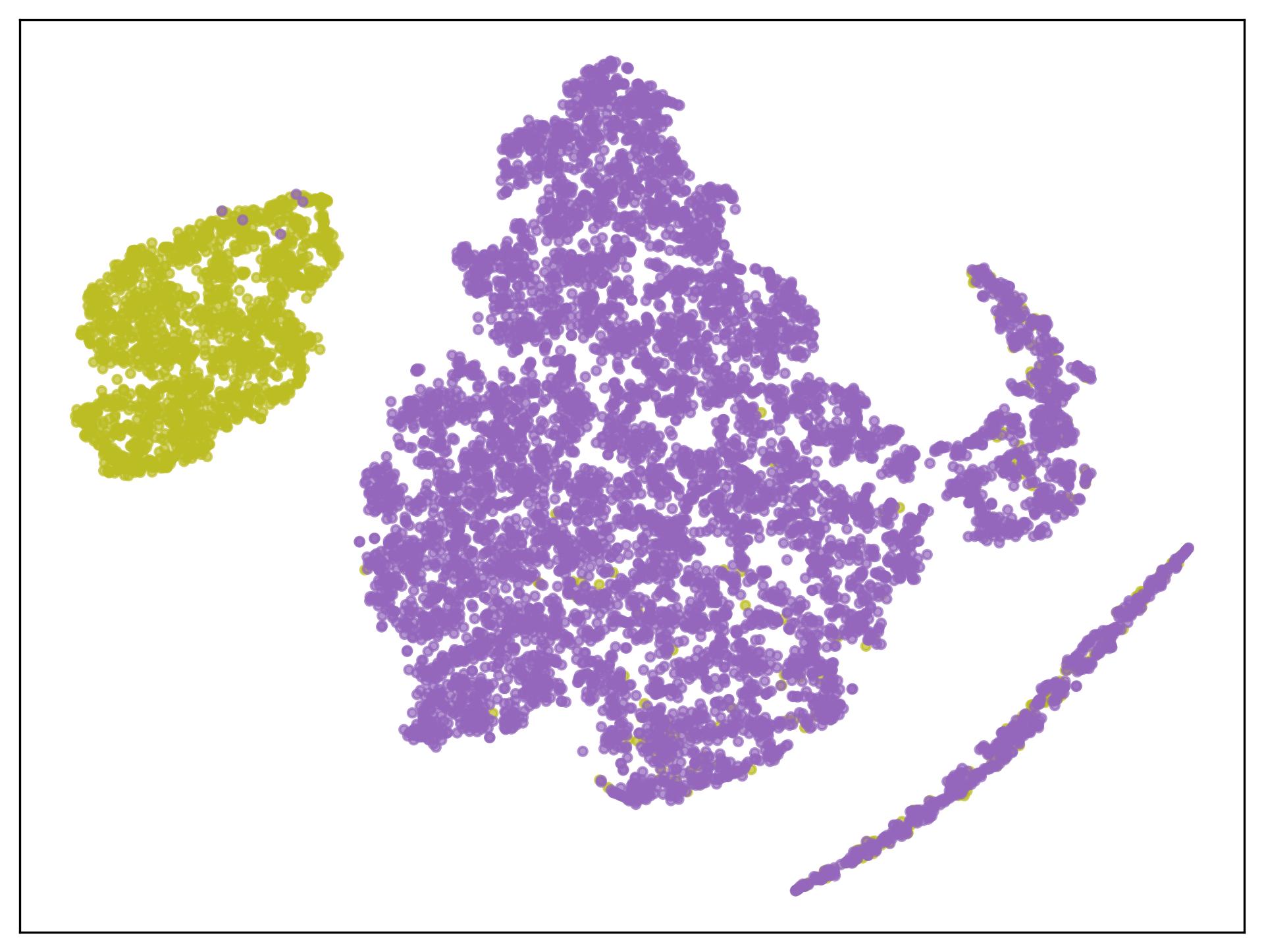}
        \caption{2:8}
    \end{subfigure}
    \hfill
    \begin{subfigure}{0.3\textwidth}
        \centering
        \includegraphics[width=\linewidth]{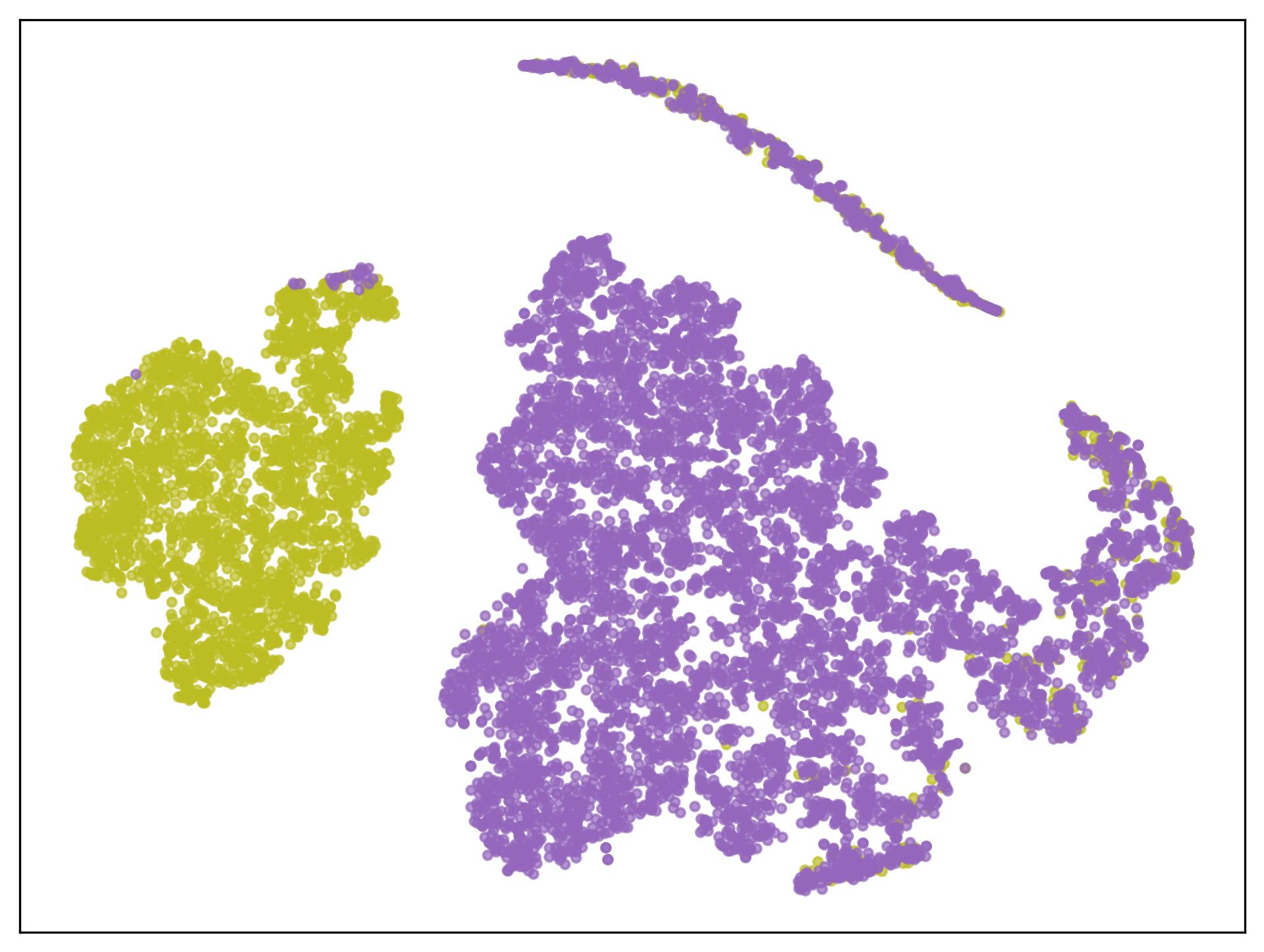}
        \caption{3:7}
    \end{subfigure}

    \vskip\baselineskip
    \begin{subfigure}{0.3\textwidth}
        \centering
        \includegraphics[width=\linewidth]{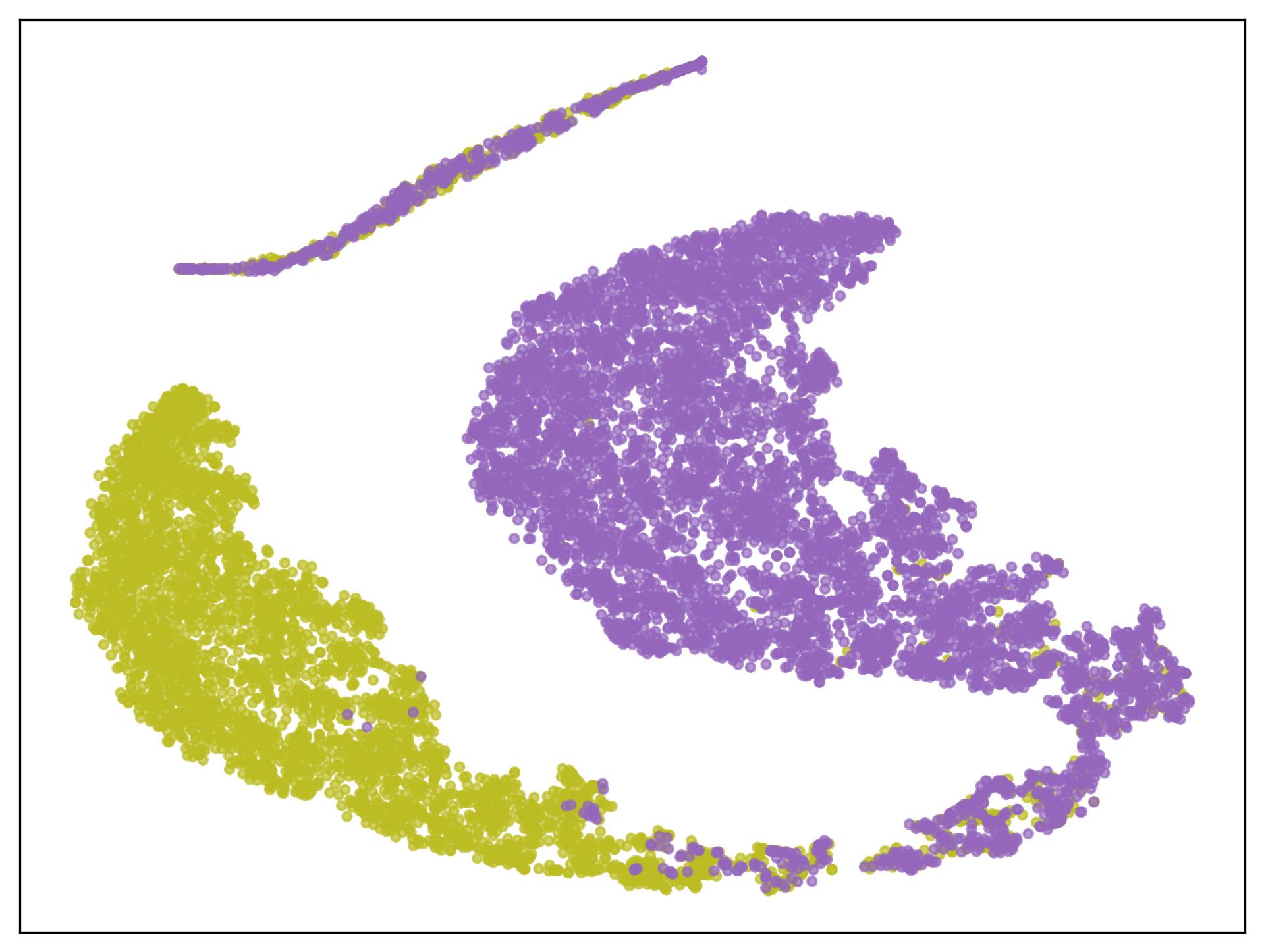}
        \caption{4:6}
    \end{subfigure}
    \hfill
    \begin{subfigure}{0.3\textwidth}
        \centering
        \includegraphics[width=\linewidth]{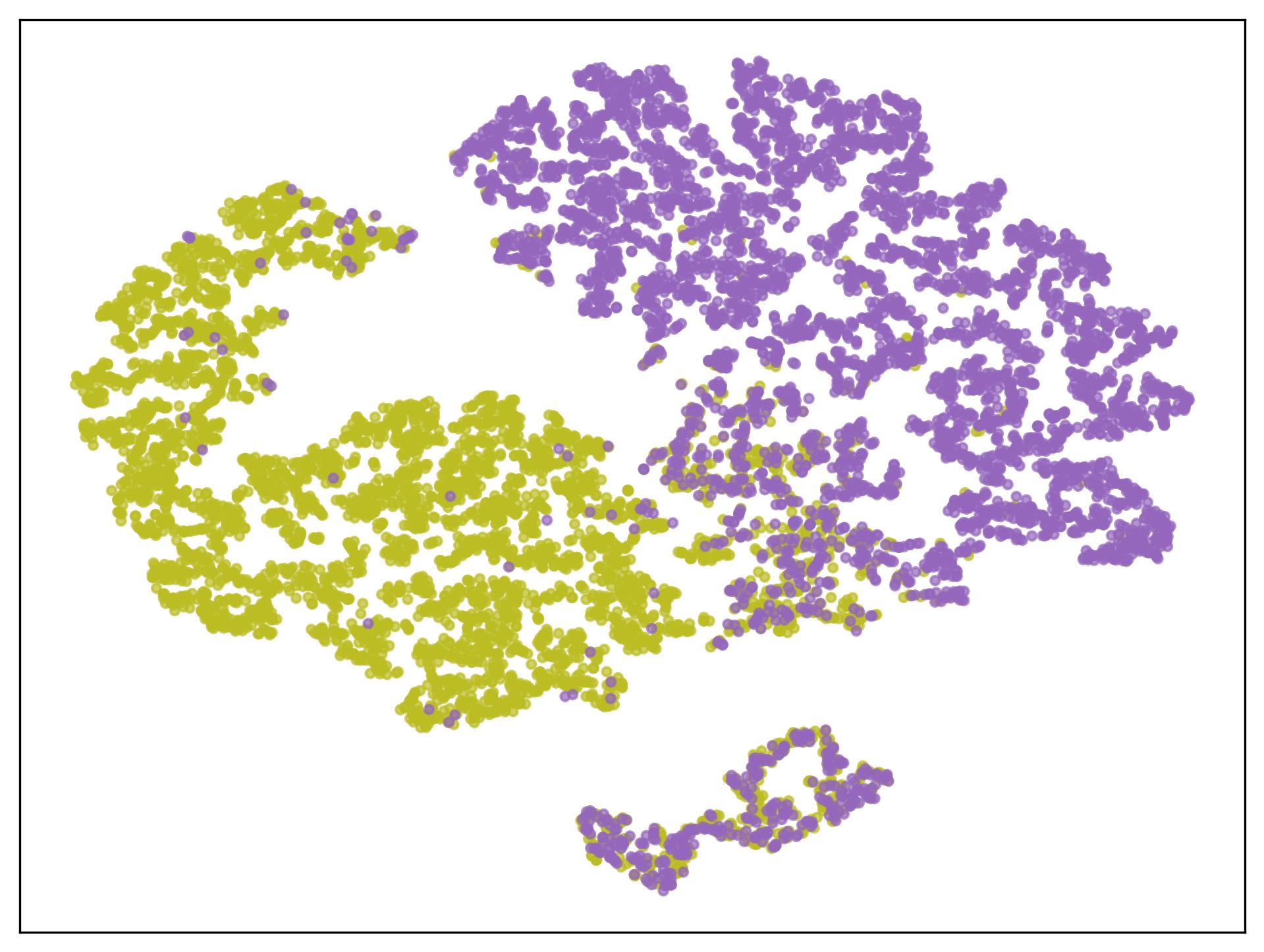}
        \caption{5:5}
    \end{subfigure}
    \hfill
    \begin{subfigure}{0.3\textwidth}
        \centering
        \includegraphics[width=\linewidth]{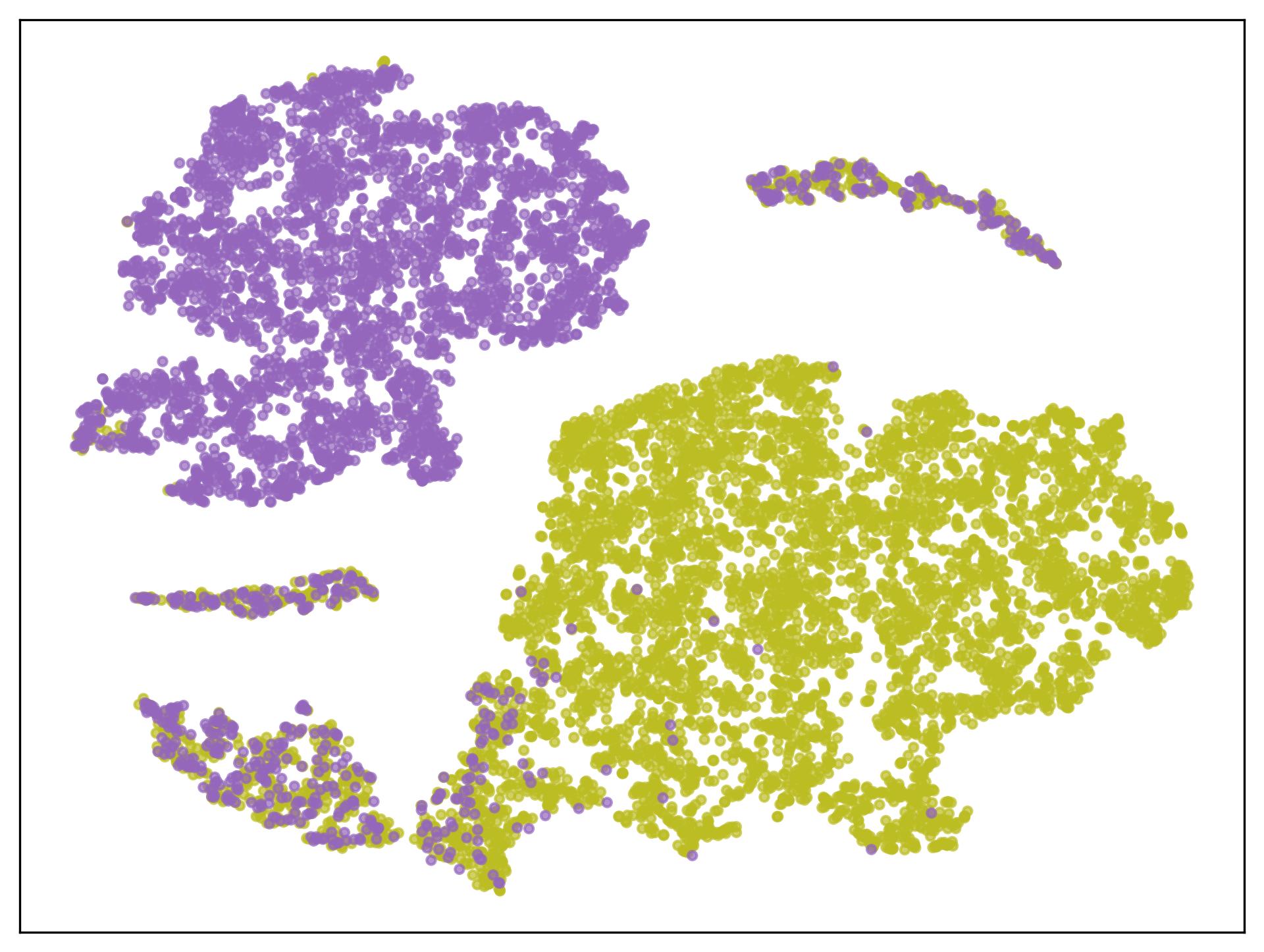}
        \caption{6:4}
    \end{subfigure}

    \vskip\baselineskip
    \begin{subfigure}{0.3\textwidth}
        \centering
        \includegraphics[width=\linewidth]{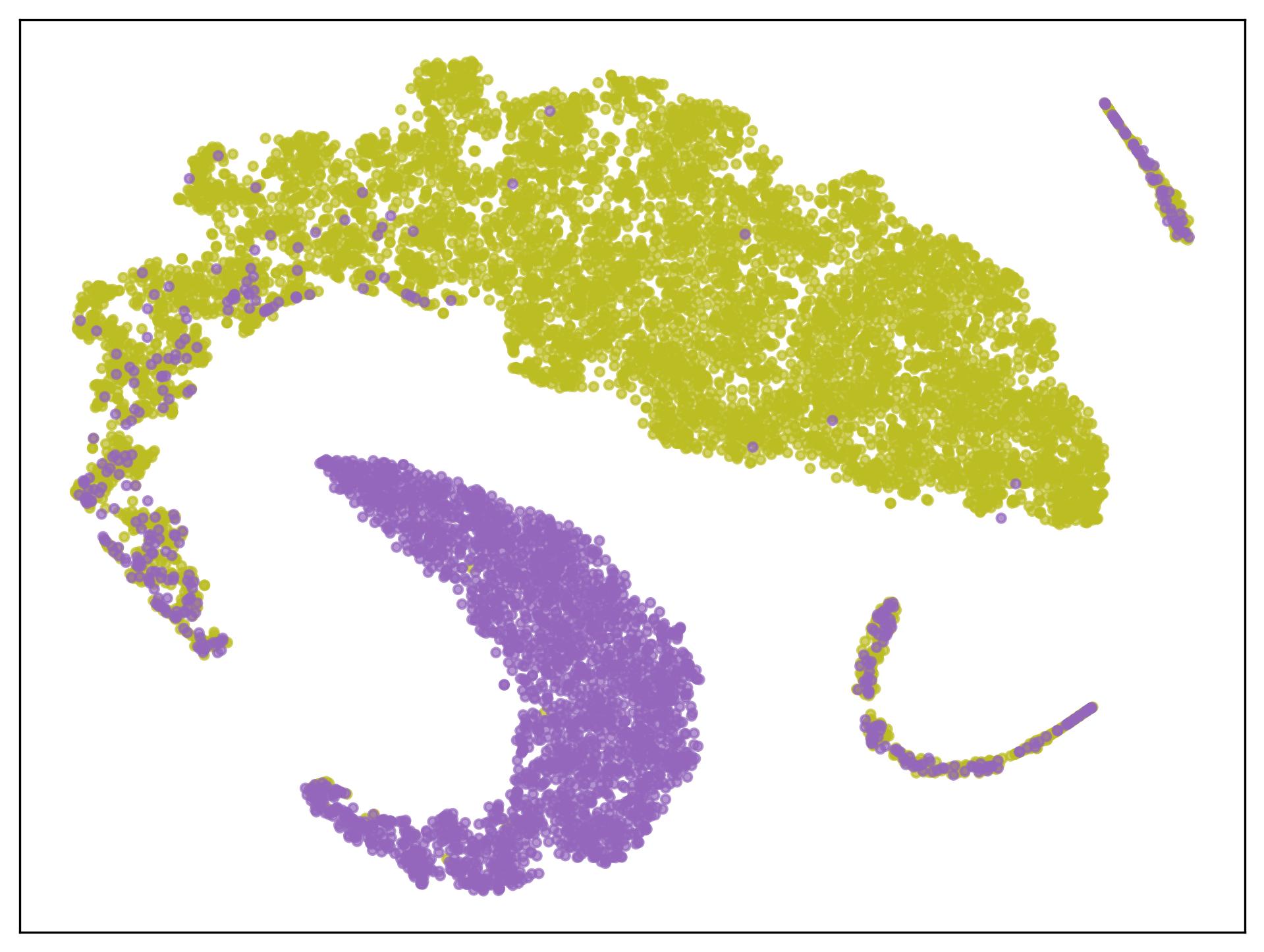}
        \caption{7:3}
    \end{subfigure}
    \hfill
    \begin{subfigure}{0.3\textwidth}
        \centering
        \includegraphics[width=\linewidth]{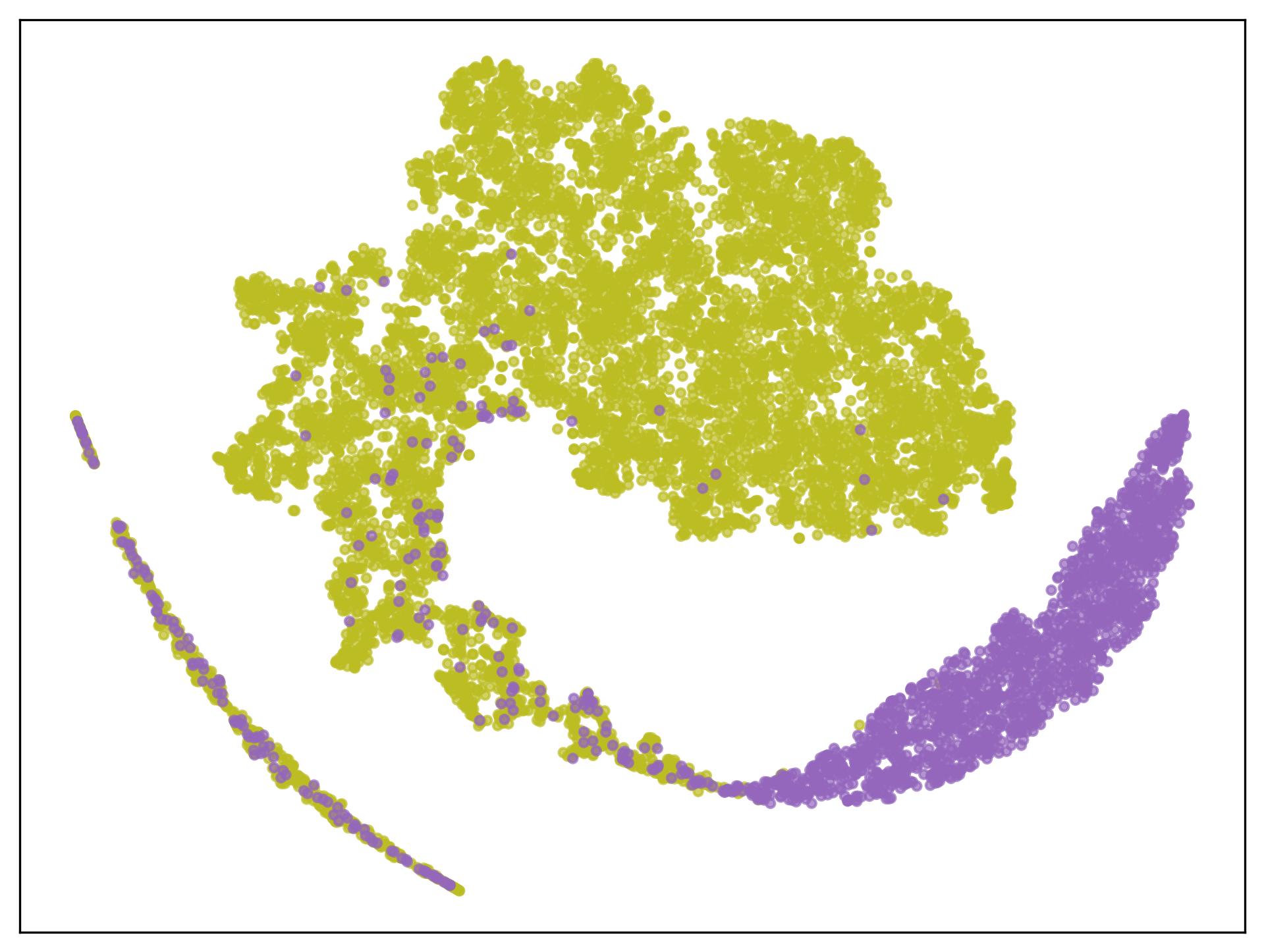}
        \caption{8:2}
    \end{subfigure}
    \hfill
    \begin{subfigure}{0.3\textwidth}
        \centering
        \includegraphics[width=\linewidth]{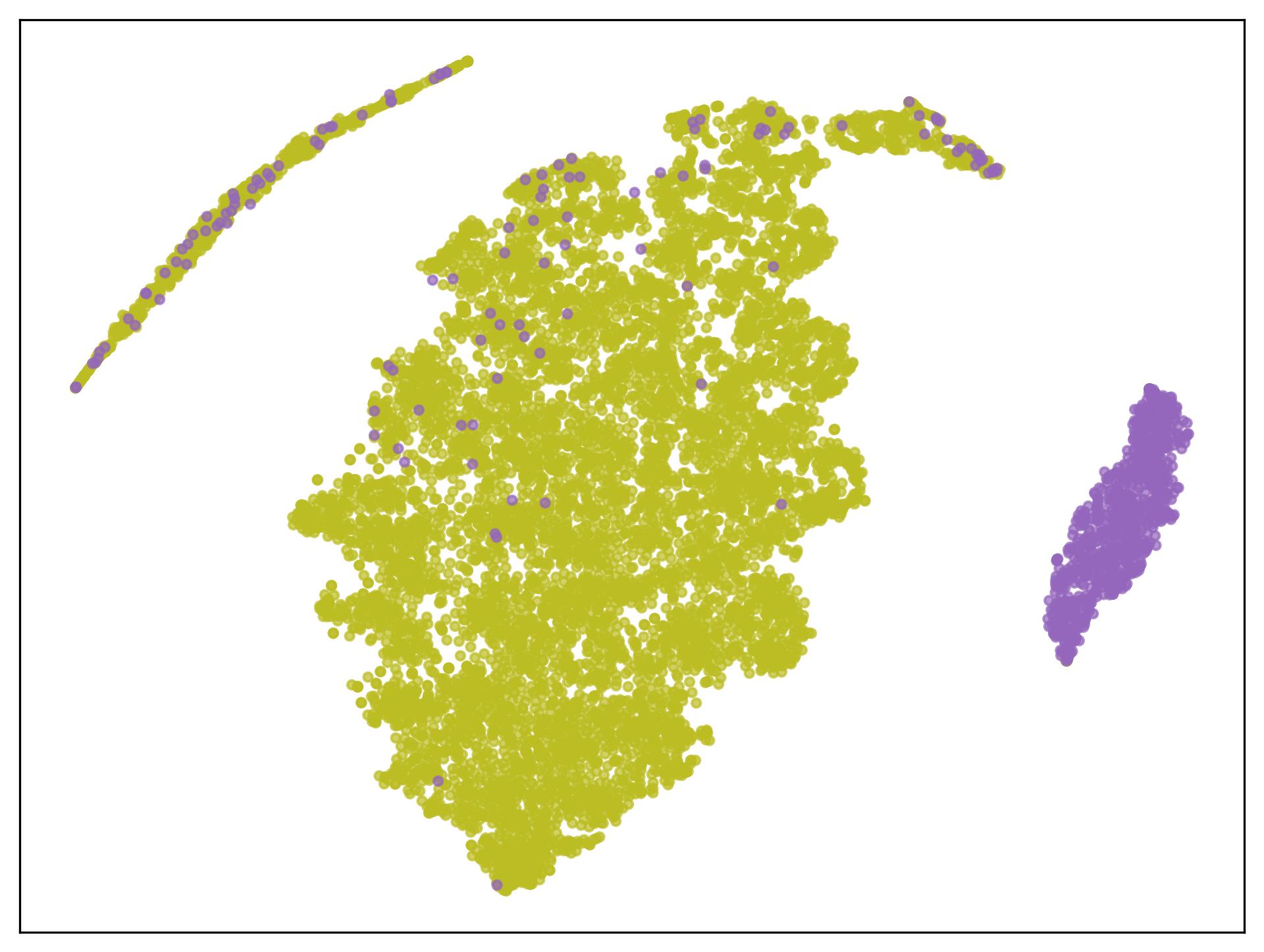}
        \caption{9:1}
    \end{subfigure}

    \caption{T-SNE visualization of user embeddings on UF-P-2 (AVG) across group ratios from 1:9 to 9:1. Points are colored by preference group.}
    \label{fig:group-ratio-ufp2}
\end{figure*}

\end{document}